%% file: _main.tex
\documentclass[twoside]{article}

\usepackage[accepted]{aistats2026}
\usepackage[round]{natbib}
\usepackage{appendix}
\usepackage{svg}

\usepackage[labelformat=parens]{caption}   
\usepackage{subcaption}      
\usepackage{centernot}
\usepackage{url}
\usepackage{amsmath}
\DeclareMathOperator*{\argmax}{arg\,max}
\DeclareMathOperator*{\argmin}{arg\,min}
\usepackage[backref=page]{hyperref}
\usepackage{microtype}
\usepackage{graphicx}
\usepackage{booktabs} 
\usepackage{arydshln}   
\usepackage{wrapfig}
\usepackage{amssymb}
\usepackage{mathtools}
\usepackage{amsthm}
\usepackage{graphicx}    
\usepackage{multirow}    
\usepackage{booktabs}
\usepackage[table]{xcolor}   
\usepackage{thmtools}
\usepackage{thm-restate}
\theoremstyle{plain}

\usepackage{fontawesome5}

\usepackage[most]{tcolorbox}
\tcbset{
  boxrule=0.7pt,
  arc=1.5pt,
  left=0.5pt,
  right=0.5pt,
  top=0.5pt,
  bottom=0.5pt,
  boxsep=4pt,
}

\newenvironment{Hypothesis}{
  \begin{tcolorbox}[
    colback=blue!5!white,
    colframe=blue!70!black,
  ]
    \textbf{Hypothesis: }
  \normalsize
}{
  \end{tcolorbox}
}

\newenvironment{takeaway}{
  \begin{tcolorbox}[
    colback=green!5!white,
    colframe=green!50!black,
  ]
    \textbf{Key Takeaway: }
  \normalsize
}{
  \end{tcolorbox}
}

\usepackage{lipsum}
\usepackage{bm}
\usepackage{multirow}
\usepackage{array}
\usepackage{multicol}
\usepackage{tabularx}
\usepackage{tabularray}
\UseTblrLibrary{booktabs}
\usepackage[inline]{enumitem}
\usepackage{lineno}
\usepackage{cleveref}
\definecolor{origBG}  {RGB}{243,246,255}  
\definecolor{calibBG} {RGB}{255,245,235}  
\definecolor{altBG}   {RGB}{250,240,255}  
\definecolor{highlightBG}{RGB}{255,240,245} 
\definecolor{myblue}{RGB}{32,119,180}
\definecolor{mygreen}{RGB}{44,160,44}
\definecolor{myorange}{RGB}{255,127,14}
\definecolor{mypurple}  {RGB}{204,153,255}  
\definecolor{mypink}    {RGB}{255,153,179}  
\newcommand{\blue}[1]{\textcolor{myblue}{#1}}
\newcommand{\green}[1]{\textcolor{mygreen}{#1}}
\newcommand{\orange}[1]{\textcolor{myorange}{#1}}

\newcommand{\ace}{\ensuremath{\widehat{\text{ACE }}}}
\newcommand{\ece}{\ensuremath{\widehat{\text{ECE }}}}
\definecolor{darkblue}{rgb}{0, 0, 0.5}
\hypersetup{colorlinks=true, citecolor=darkblue, linkcolor=darkblue, urlcolor=darkblue}
\newcommand{\bx}{\bm{x}}
\newcommand{\by}{\bm{y}}
\newcommand{\bh}{\bm{h}}
\newcommand{\bw}{\bm{w}}
\newcommand{\bb}{\bm{b}}
\newcommand{\bz}{\bm{z}}

\newcolumntype{?}{!{\vrule width 1.25pt}}

\definecolor{baseBG}   {RGB}{235,245,255}   
\definecolor{seBG}     {RGB}{255,240,220}   
\definecolor{tsBG}     {RGB}{230,250,230}   
\definecolor{plattBG}  {RGB}{255,230,230}   
\definecolor{atsBG}    {RGB}{245,235,255}   
\definecolor{escBG}    {RGB}{245,245,245}   

\begin{document}

%
\runningtitle{Improving Semantic Uncertainty Quantification in LM QA via Token-Level Temperature Scaling}

%

\twocolumn[

\aistatstitle{Improving Semantic Uncertainty Quantification in Language Model\\Question-Answering via Token-Level Temperature Scaling}



\begin{center}
\vspace*{-10pt}
\setlength{\tabcolsep}{4pt}%
\begin{tabular*}{0.93\textwidth}{@{\extracolsep{\fill}}cccc@{}}
\textbf{Tom A.~Lamb} &
\textbf{Desi R.~Ivanova} &
\textbf{Philip H.~S.~Torr} &
\textbf{Tim G.~J.~Rudner} \\\vspace*{-8pt}
University of Oxford &
University of Oxford &
University of Oxford &
University of Toronto \& Vijil \\
\end{tabular*}
\end{center}
\vskip 0.3in plus 2fil minus 0.1in
\vspace*{-10pt}
\begin{center}
\href{https://tomalamb.github.io/semantic-calibration-via-temp-scaling/}{\faGlobe\enspace Website}
\quad
\href{https://github.com/tomalamb/semantic-calibration}{\faGithub\enspace Code}
\end{center}
\vspace*{15pt}
]

\begin{abstract}
\noindent Calibration is central to reliable semantic uncertainty quantification, yet prior work has largely focused on discrimination, neglecting calibration. As calibration and discrimination capture distinct aspects of uncertainty, focusing on discrimination alone yields an incomplete picture. We address this gap by systematically evaluating both aspects across a broad set of confidence measures. We show that current approaches, particularly fixed-temperature heuristics, produce systematically miscalibrated and poorly discriminative semantic confidence distributions. We demonstrate that optimising a single scalar temperature, which, we argue, provides a suitable inductive bias, is a surprisingly simple yet effective solution. Our exhaustive evaluation confirms that temperature scaling consistently improves semantic calibration, discrimination, and downstream entropy, outperforming both heuristic baselines and more expressive token-level recalibration methods on question-answering tasks.
\end{abstract}
\begin{figure}[ht!]
    \centering
    \includegraphics[width=0.98\linewidth]{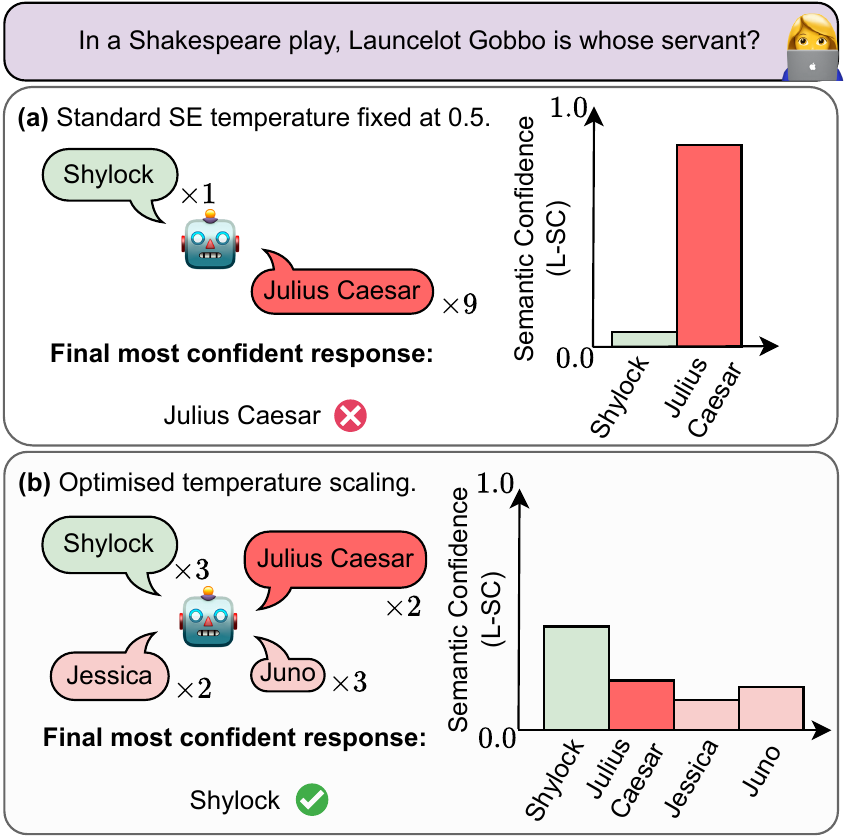}
    \vspace{-5pt}
    \caption{\textbf{Temperature Scaling Improves Semantic Uncertainty Quantification}. Same model using different temperatures generates 10 responses for the same input, which are clustered into semantic groups. We compute the L-SC semantic confidence measure of \citet{kuhn2023semantic}
    defined in \Cref{eq:L-SC}. Panel \textbf{(a)} uses the standard temperature of $0.5$; panel
    \textbf{(b)} uses a temperature optimised on a calibration set. We find that optimised temperature scaling
    improves both semantic calibration and discrimination.}
    \label{fig:pullfigure}
    \vspace*{-8pt}
\end{figure}

\section{\MakeUppercase{Introduction}} \label{sec:introduction}

Calibration, how well predicted confidences match observed frequencies, is fundamental to reliable uncertainty quantification (UQ).
However, prior work on semantic UQ for language models (LMs) has focused largely on \emph{discrimination}, evaluating measures such as semantic entropy \citep{kuhn2023semantic, farquhar2024detecting, santilli2025revisiting} without assessing calibration. This is a critical omission: perfect discrimination does not imply calibration, nor does perfect calibration necessarily imply accurate discrimination \citep{huang2020tutorial}.
Because discrimination and calibration capture distinct aspects of uncertainty, evaluating semantic UQ methods by discrimination alone yields an incomplete, and potentially misleading, assessment of reliability.

For instance, given the question \textit{``In a Shakespeare play, Launcelot Gobbo is whose servant?''} in \Cref{fig:pullfigure}, a semantically well-calibrated model should assign similarly high confidence to the semantically equivalent answers \textit{``Shylock''} and \textit{``Shylock in Merchant of Venice''} while assigning low confidence to semantically incorrect answers such as \textit{``Jessica''}.
Accurate alignment between confidence and semantic correctness is essential for trustworthy natural language generation; yet, as \Cref{fig:pullfigure} shows, standard approaches can produce systematically overconfident semantic distributions.

As semantic calibration in LMs remains underexplored, how best to recalibrate models semantically is an open question. While token-level recalibration is well-established, its translation to semantic prediction space remains undetermined \citep{kuhn2023semantic, xie2024calibrating}. This transition is fundamental: as semantic confidence measures are derived from token-level probabilities, any token-level miscalibration may translate into semantic miscalibration \citep{farquhar2024detecting, murray2018correcting}. Bridging this gap is essential for grounding semantic uncertainty quantification and may reveal surprisingly simple approaches to semantic recalibration.

\emph{Temperature scaling} (TS) is a well-established method for calibrating token-level probabilities \citep{guo2017calibration} and controlling diversity in generative LMs. While typically treated as a fixed heuristic in semantic UQ \citep{kuhn2023semantic}, we argue that optimising TS as a single global scalar provides a superior inductive bias compared to more expressive, and expensive, per-token recalibration methods like ATS. This global constraint acts as a natural regulariser, preventing the model from overfitting to filler tokens. Instead, by capturing the uncertainty of the sequence as a whole, TS more effectively reflects the overall meaning being conveyed. TS is thus a surprisingly simple and efficient tool for improving LM reliability, one that practitioners can adopt without altering existing workflows.
Since there is no generally agreed-upon way to define distributions in semantic prediction space,  a rigorous study of semantic calibration and discrimination requires evaluating a broad range of semantic confidence measures, rather than relying on a small, fixed set as done in prior work \citep{kuhn2023semantic, farquhar2024detecting}.
To address this, we consider several different approaches to extracting semantic uncertainty from LMs.

Our contributions are as follows:
\begin{enumerate}[topsep=0pt, leftmargin=15pt, itemsep=0.5pt]
    \item We provide the \textbf{first systematic evaluation of both semantic calibration and discrimination} across a wide range of semantic confidence measures, revealing that \textbf{base models and fixed-temperature heuristics} are \textbf{poorly calibrated and weakly discriminative}, thereby exposing fundamental limitations of prior work.
    \item We show that \textbf{optimised token-level temperature scaling}, a single scalar, is a surprisingly simple yet highly effective approach for semantic UQ, \textbf{outperforming} fixed-temperature baselines and \textbf{complex post-hoc calibration methods} such as ATS \citep{xie2024calibrating} and Platt scaling \citep{platt1999probabilistic}.
    \item We demonstrate that optimisation-based temperature scaling \textbf{improves downstream semantic entropy}, exposing the suboptimality of heuristics prevalent in the literature. Moreover, we show that a more \textbf{principled selection of the final response}, based directly on semantic confidence distributions rather than through an ad-hoc separate greedy-decoding procedure as done in prior work, yields \textbf{superior discrimination}.
    \item We validate the \textbf{robustness} of these findings through an \textbf{exhaustive evaluation} across multiple LMs, QA datasets, calibration metrics, and generation sample sizes.
    
\end{enumerate}

\begin{tcolorbox}[
    colback=blue!5!white,
    colframe=blue!70!black,
  ]
    Code to reproduce our experiments is available at:\\[-15pt]
    \begin{center}
        \href{https://github.com/tomalamb/semantic-calibration}{\texttt{github.com/tomalamb/semantic-calibration}}
    \end{center}
\end{tcolorbox}

\vspace{-3pt}
\section{\MakeUppercase{Related Work}}
\label{sec:background}
\vspace{-3pt}
\textbf{Confidence and Calibration for LMs.}
Confidence estimation in language models (LMs) typically relies on token-level likelihoods \citep{kadavath2022language}, post-processing techniques \citep{malininuncertainty}, or verbalised confidence scores \citep{kadavath2022language}. Calibration, the alignment between confidence and predictive correctness \citep{flach2016classifier}, is fundamental to reliability, yet often degrades following post-training procedures like RLHF \citep{achiam2023gpt, kadavath2022language}. This miscalibration necessitates post-hoc recalibration techniques such as temperature scaling (TS) \citep{guo2017calibration} or Platt scaling \citep{platt1999probabilistic}. More recently, \citet{xie2024calibrating} proposed Adaptive Temperature Scaling (ATS) to generate token-specific temperatures; however, this requires a computationally expensive transformer-based head compared to standard scalar methods such as TS.

\textbf{Semantic Uncertainty Quantification.}
Traditional UQ techniques, including Bayesian \citep{blundell2015weight, yang2023bayesian}, latent-space \citep{mukhoti2023deep, liu2020simple}, and ensemble-based \citep{lakshminarayanan2017simple} approaches, were primarily designed for classification tasks. More specific methods for LMs often focus on token-level instantiations of generations \citep{malininuncertainty}, neglecting the underlying semantics of generations \citep{kuhn2023semantic}. With the rise of open-ended generative tasks, research has shifted towards quantifying uncertainty over conveyed \emph{meanings} rather than literal token sequences. This has motivated multi-sampling methods that cluster responses by semantic equivalence, typically via Natural Language Inference (NLI) \citep{williams-etal-2018-broad}, to define \emph{semantic confidence measures} as distributions over meanings \citep{kuhn2023semantic, lin2023generating, nikitin2024kernel}. The entropy of these semantic measures is subsequently used to discriminate between \mbox{correct and incorrect responses \citep{kuhn2023semantic}.}

\textbf{The Transition to Semantic Calibration.}
While semantic UQ has improved discrimination, the calibration of semantic confidence measures remains underexplored. Consequently, it remains unknown how well existing methods align semantic confidence with actual correctness, or whether established token-level techniques can effectively translate to the more complex setting of meanings. We argue that temperature scaling (TS) is uniquely well-suited for this transition. Unlike expressive methods such as ATS which optimise for local, per-token calibration goals, TS imposes a single global constraint over the entire sequence. We posit that this constraint provides a superior inductive bias for semantics: it regularises against overfitting to semantically hollow filler tokens, forcing the calibration process to reflect the uncertainty of the sequence as a whole. Thus, TS emerges not merely as a heuristic, but as a principled, computationally cheap, and robust tool for reliable semantic UQ.

\vspace{-3pt}
\section{\MakeUppercase{Semantic Confidence}}
\label{sec:confidence measures}
\vspace{-3pt}

We consider an autoregressive LM, $p$, over a vocabulary $\mathcal{V}$ and denote the set of possible token sequences as $\mathcal{V}^*$. As established in \Cref{sec:introduction}, the absence of a unique way to define distributions over meanings motivates a broader and more holistic investigation into semantic UQ than the limited scope considered in prior work \citep{kuhn2023semantic}. Consequently, we define and evaluate seven semantic confidence measures; while two originate from existing work (E-SC and L-SC), we introduce five novel measures (ML-SC, IC-SC, B-SC, T-SC, and G-SC). Evaluating across this diverse set of semantic measures strengthens our conclusions, as we show in \Cref{fig:main results} that improvements from TS persist consistently across all measures.

For an input prompt and ground truth label sampled from a data distribution $(\bx, \by) \sim p_{\mathcal{D}}$, with $\bx,\by \in \mathcal{V}^*$, we generate $m$ responses from a LM: $\by^{(1)}, \dots, \by^{(m)} \sim p(\cdot \mid \bx)$, where each $\by^{(i)} \in \mathcal{V}^*$. We then follow \cite{kuhn2023semantic} and cluster responses based on which responses are semantically equivalent using natural language inference (NLI). 
This produces $k \in \mathbb{N}$ semantic clusters, $C_1, \dots, C_k$, where $k$ depends on the input $\bx$ and generations $\{ \by^{(i)}\}_{i=1}^m
$. For each of our proposed measures, we define a score for each cluster $C_i$. The confidence measures are then found by normalising these scores over all generated clusters.

\paragraph{Empirical Semantic Confidence (E-SC).} 
Following \citet{farquhar2024detecting}, we define the \emph{empirical semantic confidence} (E-SC) as the empirical counting measure over clusters:
\begin{equation}
    p^\text{E-SC}(C_i \mid x) \propto | C_i |, \quad \forall i\in[k],
    \label{eq:E-SC}
\end{equation}
where $[k] \coloneqq \{1,2, \dots, k \}$. We note that this is the same distribution used by \citet{farquhar2024detecting} to compute the semantic entropy of black-box models.
\paragraph{Likelihood-Based Semantic Confidence (L-SC).}
Within this work, we use length-normalised sequence likelihoods following the common practice of correcting for the exponential decay of raw likelihoods with sequence length \citep{murray2018correcting, malininuncertainty}.

For each semantic cluster $C_i$, we compute its score, denoted $s(C_i \mid \bx)$, by summing the length-normalised likelihoods of the samples within it:
\begin{equation} \label{eq:LSC_score}
    s(C_i \mid \bx) \coloneqq \sum_{\by \in C_i} p(\by \mid \bx)^{\frac{1}{|\by|}}, \quad \forall i\in[k].
\end{equation}
Normalising these scores 
yields the \emph{Likelihood-based semantic confidence} (L-SC) measure:
\begin{equation}
    p^\text{L-SC}(C_i \mid \bx) \propto s(C_i \mid \bx) , \quad \forall i\in[k]. \label{eq:L-SC}
\end{equation}
We note that this measure was originally alluded to in \citet{kuhn2023semantic} and explicitly formulated in this form in \citet{farquhar2024detecting}.

\begin{figure*}[t]
    \centering    
    \includegraphics[width=0.99\linewidth]{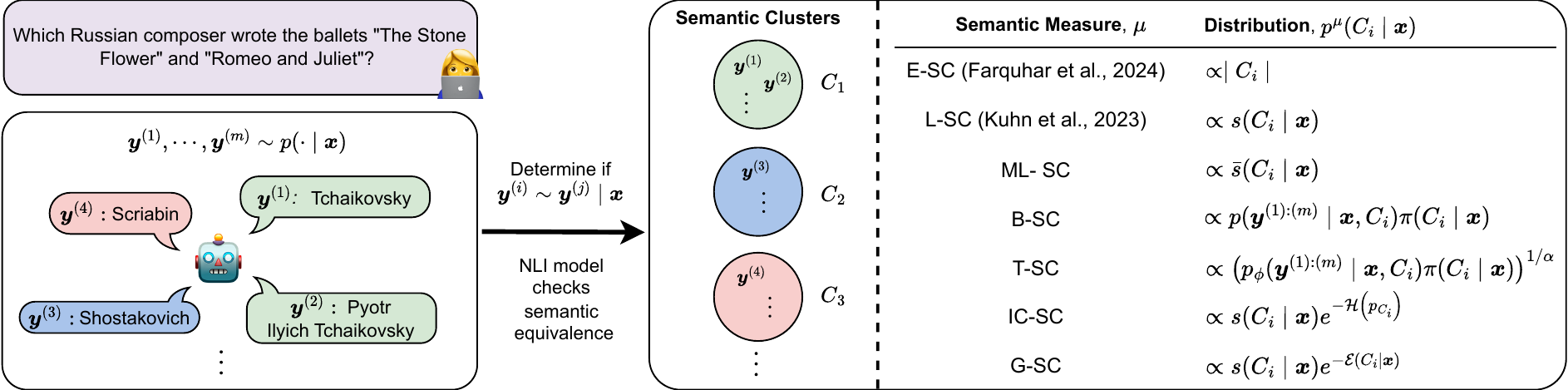}
    \captionsetup{width=\linewidth}
    \caption{\textbf{Semantic Confidence Measures.} Figure showing how existing \citep[E-SC;][]{farquhar2024detecting} and \citep[L-SC;][]{kuhn2023semantic}) as well as a broad set of novel baseline semantic measures (ML-SC, B-SC, T-SC, IC-SC, and G-SC) are computed.
    For an input $\bx$, we sample multiple responses from the model $p(\cdot \mid \bm{x})$, and use an NLI model to assess bidirectional entailment, determining whether responses $\by^{i}$ and $\by^{j}$ are semantically equivalent ($\by^{i} \sim \by^{j} \mid \bx$). Here, $s(C \mid \bm{x})$ denotes the sum, and $\bar{s}(C \mid \bm{x})$ the average, $\mathcal{H}(p_{C_i})$ the entropy, and $\mathcal{E}(C_i \mid \bx)$ the entropy of the length-normalised log-likelihoods of generations within cluster $C$. See \Cref{sec:confidence measures} for details.}
    \label{fig:motivation}
    \vspace{-5pt}
\end{figure*}

\paragraph{Mean Likelihood-Based Semantic Confidence (ML-SC).}
Summing length-normalised likelihoods may bias scores toward larger clusters, so we compute the mean score, $\bar{s}(C_i \mid \bx)$, for each cluster as:
\begin{equation*}
    \bar{s}(C_i \mid \bx) \coloneqq \frac{s(C_i \mid \bx)}{|C_i|}, \quad \forall i\in[k].
\end{equation*}
Normalising these scores yields the \emph{mean likelihood-based semantic confidence} (ML-SC) measure:
\begin{equation*}
    p^\text{ML-SC}(C_i \mid \bx) \propto \bar{s}(C_i \mid \bx), \quad \forall i\in[k]. \label{eq:ML-SC}
\end{equation*}

\paragraph{Bayesian Semantic Confidence (B-SC).}
We introduce a Bayesian-inspired semantic confidence measure that combines the E-SC and L-SC approaches. Specifically, we adopt the empirical distribution from \Cref{eq:E-SC} as a prior over clusters:
\begin{equation*}
    \pi(C_i \mid \bx) \coloneqq p^\text{E-SC}(C_i \mid \bx), \quad \forall i\in[k].
    \label{eq:BSC_prior}
\end{equation*}
We then define the cluster-level likelihood as the product of the length-normalised likelihoods for all  responses $\by^{(1):(m)} \coloneqq (\by^{(1)}, \dots, \by^{(m)})$ assigned to $C_i$: 
\begin{align*}
    \bar{p}\left(\by^{(1):(m)} \mid C_i, \bx\right) &= \prod_{\by \in C_i} p(\by \mid \bx)^{\frac{1}{|\by|}} \quad \forall i\in[k]. 
    \label{eq:BSC_likelihood}
\end{align*}
Combining the likelihood and the prior yields the posterior distribution over clusters:
\begin{equation}
    p^\text{B-SC}(C_i \mid \bx) \propto \bar{p}\left(\by^{(1):(m)} \mid C_i, \bx\right) \pi(C_i \mid \bx) ,
    \label{eq:BSC_posterior}
\end{equation}
for all $i\in[k]$. We refer to this as the \textit{Bayesian semantic confidence} (B-SC) measure.

\textbf{Tempered-Bayesian Posterior (T-SC).} 
We consider a tempered version of the Bayesian posterior in~\cref{eq:BSC_posterior} by introducing a scaling parameter $\alpha \in \mathbb{R}_{>0}$:
\begin{equation*}
\displaystyle{
    p_{\alpha}^\text{T-SC}(C_i \mid \bx) \propto \left(\bar{p}\left(\by^{(1):(m)} \mid C_i, \bx\right) \pi(C_i \mid \bx) \right)^{1/\alpha}\!,\!
    }
    \label{eq:cold_bayesian_posterior}
\end{equation*}
for all $i\in[k]$.
When $\alpha < 1$, the posterior becomes sharper, concentrating more heavily on the clusters with the highest likelihood, whilst $\alpha >1$ produces flatter distributions. 
The case  $\alpha <1$ is referred to as producing a cold posterior \citep{wenzel2020good}.

\textbf{Internal Consistency Semantic Confidence (IC-SC).} 
We introduce an unnormalised entropy-penalised semantic confidence measure that accounts for the internal consistency of clusters. 
Given an input $\bx$ and a cluster $C_i$, we define the internal agreement between the likelihoods of the responses within this cluster via:
\begin{align*}
p_{C_i}(\by_j) &\coloneq \frac{ p(\by_j \mid \bx)^{\frac{1}{|\by|}}}{\sum_{\by_k \in C_i} p(\by_k \mid \bx)^{\frac{1}{|\by|}}}
\end{align*}
and compute its internal entropy:
\begin{align*}
\mathcal{H}(p_{C_i}) &= -\sum_{\by_k \in C_i}p_{C_i} (\by_k) \log p_{C_i}(\by_k).
\end{align*}
We then penalise the scores in \cref{eq:LSC_score} with this entropy:
\begin{equation*}
    p^\text{IC-SC}(C_i \mid \bx) \propto s(C_i \mid \bx) e^{-\mathcal{H}(p_{C_i})}, \label{eq:IC-SC}
\end{equation*}
for all $i\in[k]$. This downweights  clusters containing responses with different likelihoods.

\paragraph{Gibbs Semantic Confidence (G-SC).}  
We consider the E-SC measure  again as a prior, $\pi$, over clusters as we did for \cref{eq:BSC_posterior}. 
The energy function over clusters, $\mathcal{E}(\cdot \mid \bx)$, is the negative (length-normalised) log-likelihood of the samples within this cluster:
\begin{equation*}
    \mathcal{E}(C_i \mid \bx) = -\log \bar p\left(\by^{(1):(m)}\mid C_i,\bx \right), \quad \forall i \in [k].
\end{equation*}

Using this energy, we then form the Gibbs distribution: \citep{cantoni2004statistical} as
\begin{equation*}
    p^\text{G-SC}(C_i \mid \bx ; \alpha) \propto \pi(C_i \mid \bx) e^{-\alpha \mathcal{E}(C_i \mid \bx)} \quad \forall i \in [k].\!
    \label{eq:gibbs posterior}
\end{equation*}
Here,  $\alpha \in \mathbb{R}_{>0}$ is a scaling parameter that controls the importance of the likelihood in forming the Gibbs posterior. This is  similar to the T-SC measure defined, but differs in that we only scale the likelihood term by $\alpha$, whilst the prior remains unchanged. 

\paragraph{Diversity of Behaviour.} In \Cref{sec:experiments}, we show that SC measures yield distinct uncertainty profiles, confirming there is no single uniformly best way to define semantic confidence distributions. This motivates the introduction of new measures to complement existing ones, ensuring a more robust and comprehensive evaluation of semantic UQ.

\subsection{Token-Level Recalibration }
We present several calibration techniques, often applied for token-level calibration, that we will compare in the context of semantic calibration.

\paragraph{Temperature Scaling (TS)} Given an input prompt $\bx \in \mathcal{V}^*$ and logits $\bm{z}_t \in \mathbb{R}^{|\mathcal{V}|}$ at decoding step $t$ from an LM $p(\cdot\mid\bx)$, the output probabilities are computed as
$
p(y_t\mid \bx, \by_{<t}; \tau) = \sigma\left(\bm{z}_t / \tau\right),
$
where $\sigma: \mathbb{R}^{|\mathcal{V}|} \to \Delta^{|\mathcal{V}|-1}$ is the softmax function and $\tau>0$ is a scalar temperature.

\paragraph{Adaptive Temperature Scaling (ATS)} \label{sec:ats} ATS \citep{xie2024calibrating} replaces the global scalar $\tau \in \mathbb{R}_{>0}  $ with token position-specific temperatures via a learned prediction head. Given input $\bx \in \mathcal{V}^*$ and final hidden representations $\bh \in \mathbb{R}^{d_{\text{model}} \times n}$, ATS applies a transformation $\psi_{\boldsymbol{\theta}}: \mathbb{R}^{d_{\text{model}} \times n} \to \mathbb{R}^{n}$, implemented as a single-layer transformer block \citep{vaswani2017attention}, to produce a scalar temperature for each token position: 
\begin{equation*}
    p(y_t \mid \bx, \by_{<t}; \boldsymbol{\theta}) = \sigma\left(\bz_t / \boldsymbol{\tau}_t\right).
\end{equation*}
for $\boldsymbol{\tau}^{-1} = \exp(\psi_{\boldsymbol{\theta}}(\bh))$. All operations using $\boldsymbol{\tau} \!\in\! \mathbb{R}^{n}$ are
performed element-wise.

\paragraph{Platt Scaling} \label{sec:ps}
Platt scaling is a classic technique used to recalibrate deep learning models \citep{platt1999probabilistic, niculescu2005predicting, guo2017calibration}. However, it is computationally expensive to directly apply it to LMs given the size of their vocabularies. Therefore, following \citet{xie2024calibrating}, we restrict the affine Platt transformation on the logits of a model to be diagonal:
\begin{equation*}
    p(y_t \mid \bx, \by_{<t}; \boldsymbol{\theta}) = \sigma(\text{diag}(\bw) \bz_t + \bb),
\end{equation*}
where $\boldsymbol{\theta} = (\bw, \bb) \in \mathbb{R}^{2\mid \mathcal{V} \mid }$ are the learnable parameters.

\paragraph{Temperature Scaling’s Promise for Semantic Recalibration.} Despite its lower expressivity, we argue that TS provides a superior inductive bias for semantic calibration. Unlike Platt scaling, which can distort token-level likelihood rankings, TS strictly preserves the likelihood rankings of tokens, ensuring that the relative ordering of semantically important tokens remains intact. Moreover, compared to ATS which optimises for local per-token goals, TS enforces a single global constraint across the entire generation. This global focus acts as a regulariser against overfitting to meaningless filler tokens, preventing the method from minimising loss by merely fitting frequent, non-semantic words. Instead, TS forces the calibration to reflect the uncertainty of the sequence as a whole, thereby better capturing the overall meaning conveyed. We provide an extended analysis of these mechanisms, including empirical evidence of ATS overfitting to filler tokens, in \Cref{sec:inductive_bias_discussion}. In this context, TS represents an Occam's razor-style methodology \citep{mackay2003information}.

\begin{Hypothesis}
Temperature scaling yields better semantic calibration and overall semantic UQ than more expressive methods like Platt scaling and ATS.
\end{Hypothesis}

\vspace{-3pt}
\subsection{Calibration Loss Functions} \label{sec:calibration-loss-functions}
\vspace{-3pt}
We consider two alternative losses for learning the calibration parameters  $\boldsymbol{\theta}$ (the temperature $\tau$ for TS, ATS head, or Platt scaling parameters).

\paragraph{Negative Log-Likelihood (NLL).}
As a strictly proper scoring rule \citep{gneiting2007strictly}, NLL, equivalent to standard cross-entropy with one-hot targets, is often a natural choice for achieving calibration: 
\begin{equation}\label{eq:ce loss}
    \ell_{\text{NLL}}(p(\cdot \mid \bx, \by_{<t}; \boldsymbol{\theta}), y_t) = - \log p(y_t \mid \bx, \by_{<t}; \boldsymbol{\theta}).
\end{equation}

\paragraph{Selective Smoothing (SS).}
Introduced by \cite{xie2024calibrating}, the selective smoothing loss minimises the NLL for correct token predictions while maximising the entropy for incorrect token predictions:
\begin{align}
\label{eq:ss loss}
\begin{split}
    \ell_{\text{SS}}(p(\cdot & \mid \bx, \by_{<t}; \boldsymbol{\theta}), y_t)
    \\ 
             =
             &-(1 - \alpha) \log p(y_t \mid \bx, \by_{<t}; \boldsymbol{\theta}) \bm{1}(\hat{y}_t = y_t)\\
             &-\frac{\alpha}{|\mathcal{V}|} \sum_{y \in \mathcal{V}} \log p(y \mid \bx, \by_{<t}; \boldsymbol{\theta}) \bm{1}(\hat{y}_t \neq y_t),
\end{split}
\end{align}
where $\hat{y}_t = \argmax_{y \in \mathcal{V}} p(y \mid \bx, \by_{<t}; \boldsymbol{\theta})$ is the model's top token prediction, $\bm{1}(\cdot)$ is the indicator function, and $\alpha \in [0, 1]$ controls the balance between the two terms.

\paragraph{Optimisation Objective.} The parameters $\boldsymbol{\theta}$ are optimised by minimising the expected loss  over the distribution, $p_{\mathcal{D}}$: \begin{equation} \label{eq:erm_objective} \boldsymbol{\theta}^* = \argmin_{\boldsymbol{\theta}} \mathbb{E}_{(\bx, \by) \sim p_{\mathcal{D}}}\left[ \ell\left(p(\cdot \mid \bx; \boldsymbol{\theta}), \by\right) \right], \end{equation} where $\ell$ 
denotes the loss function and can be the sequence-level aggregation of the token-level NLL of \Cref{eq:ce loss} or SS of \Cref{eq:ss loss}. We optimise this objective using Stochastic Gradient Descent (SGD) over a calibration set of samples.

\begin{figure*}[t]
    \centering
    \includegraphics[width=0.99\linewidth]{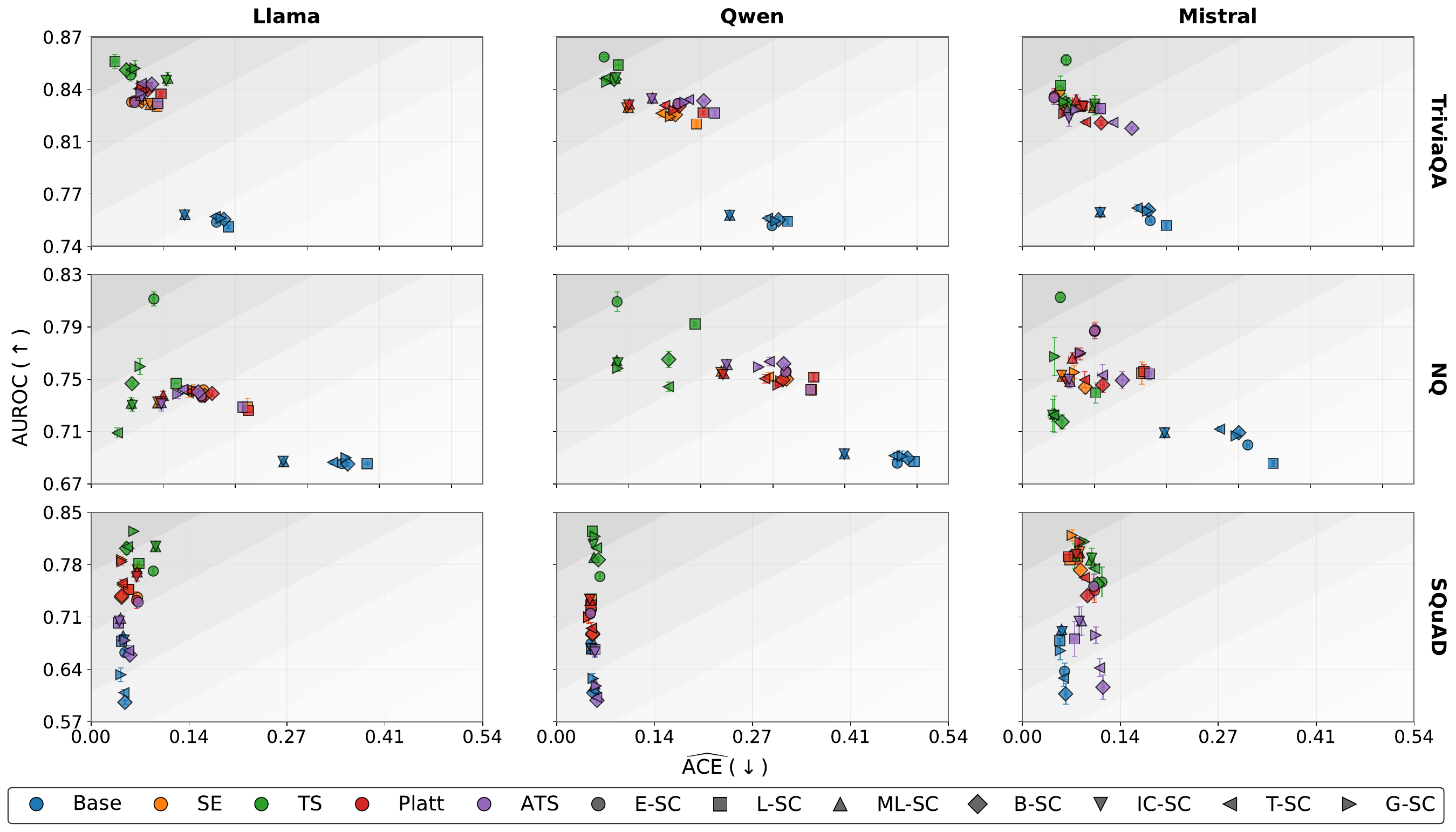}
    \vspace{-5pt}
        \caption{\textbf{Uncertainty Metrics of SC Measures Across Methods.} Mean and standard error of $\widehat{\text{ACE}}$ ($\downarrow$) and AUROC ($\uparrow$) scores for SC measures across baseline, calibration methods, and datasets. {\textit{Closer to top left of plots indicates better discrimination and calibration, and hence better overall semantic uncertainty quantification.}}}
        \vspace{-5pt}
    \label{fig:main results}
\end{figure*}

\vspace{-3pt}
\section{\MakeUppercase{Empirical Evaluation}} \label{sec:experiments}
\vspace{-3pt}
\paragraph{Models and Datasets.} 
We evaluate Llama-3.1-8B-Instruct \citep{dubey2024llama}, Ministral-8B-Instruct-2410 \citep{mistral2024ministraux}, and Qwen-2.5-7B-Instruct \citep{qwen2024qwen} on generative short-form question answering (QA). We focus on this task to enable direct comparison with prior work on semantic uncertainty quantification \citep{kuhn2023semantic, farquhar2024detecting} and because semantic correctness is well-defined, unlike for more nuanced tasks such as summarisation or creative writing (see \Cref{sec:discussion}). We use TriviaQA \citep{joshi2017triviaqa} and Natural Questions (NQ; \citealt{kwiatkowski2019natural}) for closed-book QA, and SQuAD \citep{rajpurkar2016squad} for open-book QA. Each dataset is split into calibration and test sets (\Cref{sec:dataset splits}), with calibration further divided into training and validation splits for hyperparameter selection (\Cref{sec:hyperparameters}). We use few-shot prompting throughout \citep{kuhn2023semantic, aichberger2025improving}, with 10 in-context examples for TriviaQA and NQ, and 4 for SQuAD due to longer contexts.

\vspace{-3pt}
\paragraph{Calibration Methods and Baselines.} We  compare a range of calibration methods and baselines:
\vspace*{-8pt}
\begin{itemize}[leftmargin=1em]
\setlength{\itemsep}{1pt}
\setlength{\parskip}{0pt}
\setlength{\topsep}{0pt}
    \item \textit{Uncalibrated base model} ($\text{Base}$): the original instruction-tuned LM with temperature fixed at $\tau = 1.0$ (used by \citet{farquhar2024detecting}).
    \item \textit{Base model with SE temperature setting} ($\text{SE}$): the original instruction-tuned LM, with temperature fixed at $\tau = 0.5$, 
    the setting used in semantic entropy \citep{kuhn2023semantic}.
    \item \textit{Temperature scaling} ($\text{TS}$): Optimised single scalar temperature parameter.
    \item \textit{Adaptive temperature scaling} (${\text{ATS}})$: Adaptive temperature scaling methods, optimising a temperature prediction head as detailed in \Cref{sec:ats}.
   \item \textit{Platt scaling} ($\text{Platt}$): Platt scaling optimising a diagonal affine transformation as in \Cref{sec:ps}.
\end{itemize}

\begin{table*}[ht!]
  \centering
\vspace{-5pt}
  \setlength{\tabcolsep}{5pt}
  \label{tab:se-combined}
    \caption{\textbf{Discrimination Comparison of Entropy for Qwen.} 
Mean and standard error of AUROC ($\uparrow$) values. 
(a) reports $\text{SE}_{\text{conf}}$, where correctness is determined by the most confident semantic cluster under a given SC measure.
(b) reports $\text{SE}_{\text{vanilla}}$ from \citet{kuhn2023semantic}, where correctness is determined via greedy decoding. 
Bold entries denote the best result within each SC measure per dataset, and \underline{underlined} entries indicate the best overall per dataset.}
  \begin{subtable}{0.49\textwidth}
    \centering
    \footnotesize
    \resizebox{\textwidth}{!}{%
\begin{tabular}{cccccc}
\toprule
 &  & E-SC & L-SC & IC-SC & G-SC \\
\midrule
\multirow{5}{*}{\rotatebox{90}{\textbf{TriviaQA}}} 
 & $\text{Base}$ & $0.766_{{\pm 0.002}}$ & $0.761_{{\pm 0.002}}$ & $0.765_{{\pm 0.002}}$ & $0.767_{{\pm 0.002}}$ \\
 & $\text{SE}$   & $0.834_{{\pm 0.002}}$ & $0.833_{{\pm 0.001}}$ & $0.830_{{\pm 0.002}}$ & $0.834_{{\pm 0.002}}$ \\
 & $\text{Platt}$& $0.839_{{\pm 0.002}}$ & $0.840_{{\pm 0.001}}$ & $0.836_{{\pm 0.001}}$ & $0.839_{{\pm 0.002}}$ \\
 & $\text{ATS}$  & $0.843_{{\pm 0.001}}$ & $0.840_{{\pm 0.001}}$ & $0.838_{{\pm 0.002}}$ & $0.843_{{\pm 0.001}}$ \\
  &  $\text{TS}$   & $\mathbf{0.853_{{\pm 0.002}}}$ & $\mathbf{\underline{0.865}_{{\pm 0.002}}}$ & $\mathbf{\underline{0.864}_{{\pm 0.004}}}$ & $\mathbf{0.851_{{\pm 0.002}}}$ \\
\midrule
\multirow{5}{*}{\rotatebox{90}{\textbf{NQ \; }}} 
 & $\text{Base}$ & $0.693_{{\pm 0.003}}$ & $0.691_{{\pm 0.002}}$ & $0.692_{{\pm 0.003}}$ & $0.693_{{\pm 0.003}}$ \\
 & $\text{SE}$   & $0.749_{{\pm 0.003}}$ & $0.752_{{\pm 0.002}}$ & $0.746_{{\pm 0.003}}$ & $0.750_{{\pm 0.003}}$ \\
 & $\text{Platt}$& $0.746_{{\pm 0.003}}$ & $0.755_{{\pm 0.005}}$ & $0.752_{{\pm 0.002}}$ & $0.746_{{\pm 0.003}}$ \\
 & $\text{ATS}$  & $0.759_{{\pm 0.003}}$ & $0.754_{{\pm 0.005}}$ & $0.743_{{\pm 0.001}}$ & $\mathbf{0.759_{{\pm 0.002}}}$ \\
  &  $\text{TS}$   & $\mathbf{0.769_{{\pm 0.004}}}$ & $\mathbf{\underline{0.795}_{{\pm 0.007}}}$ & $\mathbf{\underline{0.789}_{{\pm 0.003}}}$ & $0.747_{{\pm 0.002}}$ \\
\midrule
\multirow{5}{*}{\rotatebox{90}{\textbf{SQuAD }}} 
 & $\text{Base}$ & $0.569_{{\pm 0.007}}$ & $0.605_{{\pm 0.004}}$ & $0.598_{{\pm 0.003}}$ & $0.571_{{\pm 0.008}}$ \\
 & $\text{SE}$   & $0.666_{{\pm 0.007}}$ & $0.655_{{\pm 0.006}}$ & $0.669_{{\pm 0.005}}$ & $0.665_{{\pm 0.007}}$ \\
 & $\text{Platt}$& $0.665_{{\pm 0.007}}$ & $0.649_{{\pm 0.002}}$ & $0.660_{{\pm 0.002}}$ & $0.666_{{\pm 0.007}}$ \\
 & $\text{ATS}$  & $0.579_{{\pm 0.009}}$ & $0.649_{{\pm 0.002}}$ & $0.649_{{\pm 0.006}}$ & $0.591_{{\pm 0.009}}$ \\
  & $\text{TS}$   & $\mathbf{\underline{0.780}_{{\pm 0.003}}}$ & $\mathbf{0.704_{{\pm 0.004}}}$ & $\mathbf{0.772_{{\pm 0.002}}}$ & $\mathbf{\underline{0.780}_{{\pm 0.003}}}$ \\
\bottomrule
\end{tabular}}%
    \label{tab:se-ours-comparison}
    \vspace{0.15em}
    \caption{ $\text{SE}_{\text{conf}}$ (correctness via most confident cluster).}
  \end{subtable}
  \hfill
  \begin{subtable}{0.49\textwidth}
    \centering
    \footnotesize
    \resizebox{\textwidth}{!}{
\begin{tabular}{cccccc}
\toprule
 &  & E-SC & L-SC & IC-SC & G-SC \\
\midrule
\multirow{5}{*}{\rotatebox{90}{\textbf{TriviaQA}}} 
 & $\text{Base}$ & $0.755_{{\pm 0.002}}$ & $0.755_{{\pm 0.002}}$ & $0.754_{{\pm 0.002}}$ & $0.755_{{\pm 0.002}}$ \\
 & $\text{SE}$   & $0.833_{{\pm 0.001}}$ & $0.833_{{\pm 0.001}}$ & $0.833_{{\pm 0.001}}$ & $0.833_{{\pm 0.001}}$ \\
 & $\text{Platt}$& $0.832_{{\pm 0.002}}$ & $0.832_{{\pm 0.002}}$ & $0.832_{{\pm 0.002}}$ & $0.832_{{\pm 0.002}}$ \\
 & $\text{ATS}$  & $0.834_{{\pm 0.001}}$ & $0.832_{{\pm 0.002}}$ & $0.832_{{\pm 0.002}}$ & $0.834_{{\pm 0.001}}$ \\
  & $\text{TS}$   & $\mathbf{0.849_{{\pm 0.001}}}$ & $\mathbf{0.844_{{\pm 0.001}}}$ & $\mathbf{\underline{0.857_{{\pm 0.002}}}}$ & $\mathbf{0.848_{{\pm 0.002}}}$ \\
\midrule
\multirow{5}{*}{\rotatebox{90}{\textbf{NQ \; }}} 
 & $\text{Base}$ & $0.690_{{\pm 0.002}}$ & $0.689_{{\pm 0.002}}$ & $0.689_{{\pm 0.002}}$ & $0.691_{{\pm 0.002}}$ \\
 & $\text{SE}$   & $0.749_{{\pm 0.001}}$ & $\mathbf{0.745_{{\pm 0.001}}}$ & $0.745_{{\pm 0.001}}$ & $\mathbf{0.750_{{\pm 0.001}}}$ \\
 & $\text{Platt}$& $0.745_{{\pm 0.001}}$ & $0.743_{{\pm 0.003}}$ & $0.744_{{\pm 0.003}}$ & $0.744_{{\pm 0.002}}$ \\
 & $\text{ATS}$  & $0.740_{{\pm 0.002}}$ & $0.743_{{\pm 0.003}}$ & $0.741_{{\pm 0.002}}$ & $0.740_{{\pm 0.002}}$ \\
  &  $\text{TS}$   & $\mathbf{\underline{0.758_{{\pm 0.001}}}}$ & $0.737_{{\pm 0.002}}$ & $\mathbf{0.751_{{\pm 0.004}}}$ & $0.745_{{\pm 0.001}}$ \\
\midrule
\multirow{5}{*}{\rotatebox{90}{\textbf{SQuAD }}} 
 & $\text{Base}$ & $0.590_{{\pm 0.001}}$ & $0.595_{{\pm 0.001}}$ & $0.595_{{\pm 0.002}}$ & $0.595_{{\pm 0.001}}$ \\
 & $\text{SE}$   & $0.653_{{\pm 0.002}}$ & $0.653_{{\pm 0.001}}$ & $0.653_{{\pm 0.002}}$ & $0.653_{{\pm 0.002}}$ \\
 & $\text{Platt}$& $0.653_{{\pm 0.002}}$ & $0.652_{{\pm 0.004}}$ & $0.649_{{\pm 0.004}}$ & $0.653_{{\pm 0.002}}$ \\
 & $\text{ATS}$  & $0.575_{{\pm 0.007}}$ & $0.652_{{\pm 0.004}}$ & $0.624_{{\pm 0.002}}$ & $0.578_{{\pm 0.009}}$ \\
  &  $\text{TS}$   & $\mathbf{0.707_{{\pm 0.007}}}$ & $\mathbf{0.717_{{\pm 0.005}}}$ & $\mathbf{\underline{0.748_{{\pm 0.003}}}}$ & $\mathbf{0.707_{{\pm 0.007}}}$ \\
\bottomrule
\end{tabular}}
    \label{tab:vanilla-se-comparison}
    \caption{ $\text{SE}_{\text{vanilla}}$ (correctness via greedy-decoding).}
  \end{subtable}
\vspace{-15pt}
\end{table*}
\renewcommand{\arraystretch}{1.0}
\vspace{-5pt}

\vspace{-5pt}
\paragraph{Producing semantic clusters.} We form semantic clusters using the DeBERTa-V2-XXLarge NLI model \citep{he2021deberta}, following the same methodology as prior work \citep{kuhn2023semantic}. This approach has been shown to produce consistent semantic clusterings for short-form generative question answering tasks, which are the focus of both previous studies and our evaluation for direct comparison \citep{kuhn2023semantic, farquhar2024detecting}. We further corroborate these findings in an experiment detailed in \cref{sec: clustering quality}.

\vspace{-5pt}
\paragraph{Selecting a Final Response.}
For each semantic measure  in \Cref{sec:confidence measures}, we identify the most confident cluster. Since  responses in clusters are semantically equivalent, any  member could serve as the model’s final answer. For robustness against minor variations, we randomly sample a subset of up to four responses from this top cluster and compare each  against the ground truth. If at least one sampled response is correct, we mark the model’s response as correct. 
This makes correctness evaluation less brittle to superficial phrasing differences and better reflects whether the model has identified the correct underlying meaning.

\vspace{-5pt}
\paragraph{Measuring Correctness.} In our QA experiments, we view correctness as binary, defined as 
$
c = \bm{1}(\hat{\by} \sim \by \mid \bx),
$
where $\sim$ denotes semantic equivalence given $\bx$, and $\hat{\by} \sim p(\;\cdot \mid \bx; \boldsymbol{\theta})$ is a model's final response. To precisely determine the equivalence of a response to a ground truth reference, we use a combination of criteria that includes soft matching, SQuAD-F1 and Rouge-L metrics \citep{kuhn2023semantic, farquhar2024detecting, aichberger2025improving}. We discuss and justify this evaluation setup in more detail in \Cref{sec: evaluation of accuracy}. In addition, in \Cref{sec:practical-issues}, we note some common edge cases within this evaluation setup that we carefully mitigate in this work and which are not explicitly addressed in prior work using similar setups \citep{kuhn2023semantic, farquhar2024detecting, aichberger2025improving}.

\vspace{-5pt}
\paragraph{Calibration and Discrimination Metrics.}
We measure semantic calibration via Adaptive Calibration Error ($\widehat{\text{ACE}}$) \citep{nixon2019measuring}; \Cref{sec: robustness of choice of calibration metric} shows that our calibration results are robust to the choice of calibration metric including ECE \citep{naeini2015obtaining} and CORP-MCB \citep{dimitriadis2021stable} metrics). We assess the discrimination of semantic measures across methods by reporting AUROC scores \citep{bradley1997use}. See \Cref{sec:metrics} for more details. 

\vspace{-5pt}
\paragraph{Reporting Results.} We perform four independent runs, which include both calibration and evaluation stages. We compute the mean and standard error of each evaluation metric over the four inference runs.

\subsection{Optimised Temperatures Improve Semantic Uncertainty Quantification} 
\Cref{fig:main results} shows \ace and AUROC evaluated across test sets, semantic confidence (SC) measures, and methods. Overall, optimised Temperature Scaling (\green{TS}) proves to be a simple and robust method for improving semantic uncertainty quantification, echoing the key findings of \citep{guo2017calibration} non-trivially in a more complex domain. In particular, TS outperforms complex methods such as Platt scaling and ATS, whose performance is less robust in a semantic setting. It consistently produces results toward the top-left (lower \ace, higher AUROC) on both closed-book (TriviaQA, NQ) and open-book (SQuAD) datasets. On the open-book SQuAD dataset, where base models are already well-calibrated, improvements are primarily in discrimination, and G-SC consistently yields the strongest AUROC. Conversely, on the closed-book TriviaQA and NQ datasets, base models exhibit poor calibration. In this setting, E-SC and G-SC provide the best balance, delivering consistent improvements in both \ace and AUROC. Across all experiments, E-SC, L-SC, and G-SC emerge as the most effective semantic measures. Finally, and crucially, optimised TS outperforms the fixed, ad-hoc temperature settings from prior work, including $\tau=1.0$ from \citet{farquhar2024detecting} (\blue{Base}) and $\tau=0.5$ from \citet{kuhn2023semantic} (\orange{SE}).

\begin{takeaway}
Optimised temperature scaling improves semantic calibration and discrimination.
\end{takeaway}

\vspace{-3pt}
\subsection{Optimised Token-Level Temperature Improves Semantic Entropy} \label{sec:entropy-comparison}
\vspace{-3pt}
We evaluate our methods on the downstream task of discriminating correct from incorrect instances using semantic entropy (SE) \citep{kuhn2023semantic}. We compare our principled variant, $\text{SE}_{\text{conf}}$, which selects the final answer from the most confident semantic class (see \Cref{sec:experiments}), against the ad-hoc baseline from prior work, $\text{SE}_{\text{vanilla}}$. The latter determines correctness using a greedily decoded final answer, while estimating uncertainty by sampling from a temperature-smoothed distribution \citep{kuhn2023semantic}. As a result, the reported performance of the \blue{Base} method can differ between panels (a) and (b) of \Cref{tab:se-combined}, despite using the same model and temperature ($\tau = 1.0$), due to the differences in how the final response is selected for semantic correctness evaluation.

\Cref{tab:se-combined} reports the results for the Qwen model across datasets. TS consistently improves the discriminative power of each semantic measure's entropy under both formulations, crucially outperforming the fixed-temperature heuristics (e.g., $\tau \in \{0.5, 1.0\}$) from prior work once again \citep{kuhn2023semantic, farquhar2024detecting}. Gains are particularly notable on well-calibrated datasets like SQuAD, highlighting the favourable trade-off TS provides between calibration and discrimination. Moreover, a direct comparison reveals that our $\text{SE}_{\text{conf}}$ shown in panel (a) approach generally achieves higher discrimination (AUROC) than $\text{SE}_{\text{vanilla}}$ shown in panel (b). This validates deriving the final answer according to the semantic measures used for uncertainty quantification provides a more principled and effective method for downstream semantic UQ.

\begin{takeaway}Temperature scaling enhances semantic entropy discrimination. Selecting final responses via semantic confidence yields superior discrimination to adhoc prior approaches.
\end{takeaway}

\begin{figure}[t]
    \centering
    \vspace{-3pt}
    \includegraphics[width=1.0\linewidth]{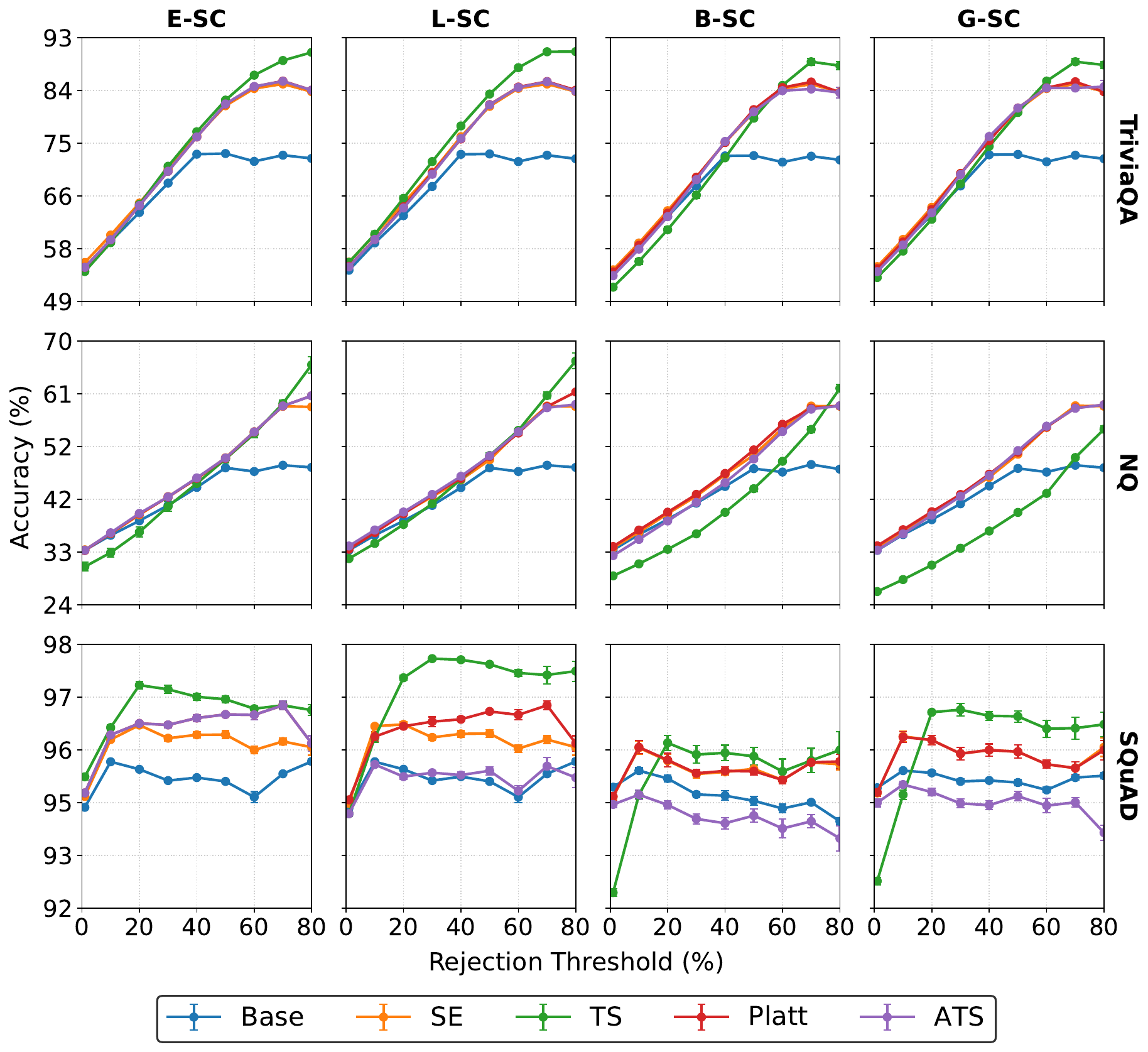}
    \vspace{-10pt}
    \caption{\textbf{Selective Accuracy for Qwen Across Rejection Rates and Datasets.} Mean selective accuracy ($\uparrow$) curves, comparing E-SC, L-SC, B-SC and G-SC measures across baseline and calibration methods. Error bars show one standard error across runs.}
    \label{fig:select-acc-qwen}
\end{figure}

\subsection{Selective Prediction}
\label{sec:selective-prediction}
\Cref{fig:select-acc-qwen} presents selective accuracy results. We observe distinct behaviours across datasets: on SQuAD, TS yields pronounced gains even at low rejection rates, whereas on TriviaQA, benefits appear primarily at high rejection rates by mitigating the most overconfident errors. On NQ, we note a drop in base performance for B-SC and G-SC; this occurs due to lower base accuracy when using B-SC and G-SC for selecting the most probable meaning under the model. Crucially, however, TS consistently drives a sharper rate of improvement (steeper slope) across all settings once rejection begins. This demonstrates that, regardless of the starting point, TS aligns confidence scores more effectively with correctness, ensuring that rejecting low-confidence predictions rapidly and reliably filters incorrect responses.

\begin{takeaway}
Temperature scaling improves selective accuracy by strictly aligning confidence with correctness, driving a generally steeper rise in accuracy as low confidence examples are filtered out.
\end{takeaway}

\subsection{Number of Generations Ablation} \label{sec: number-of-example-ablation}
\begin{figure}[ht!]
    \centering
    \includegraphics[width=1.0\linewidth]{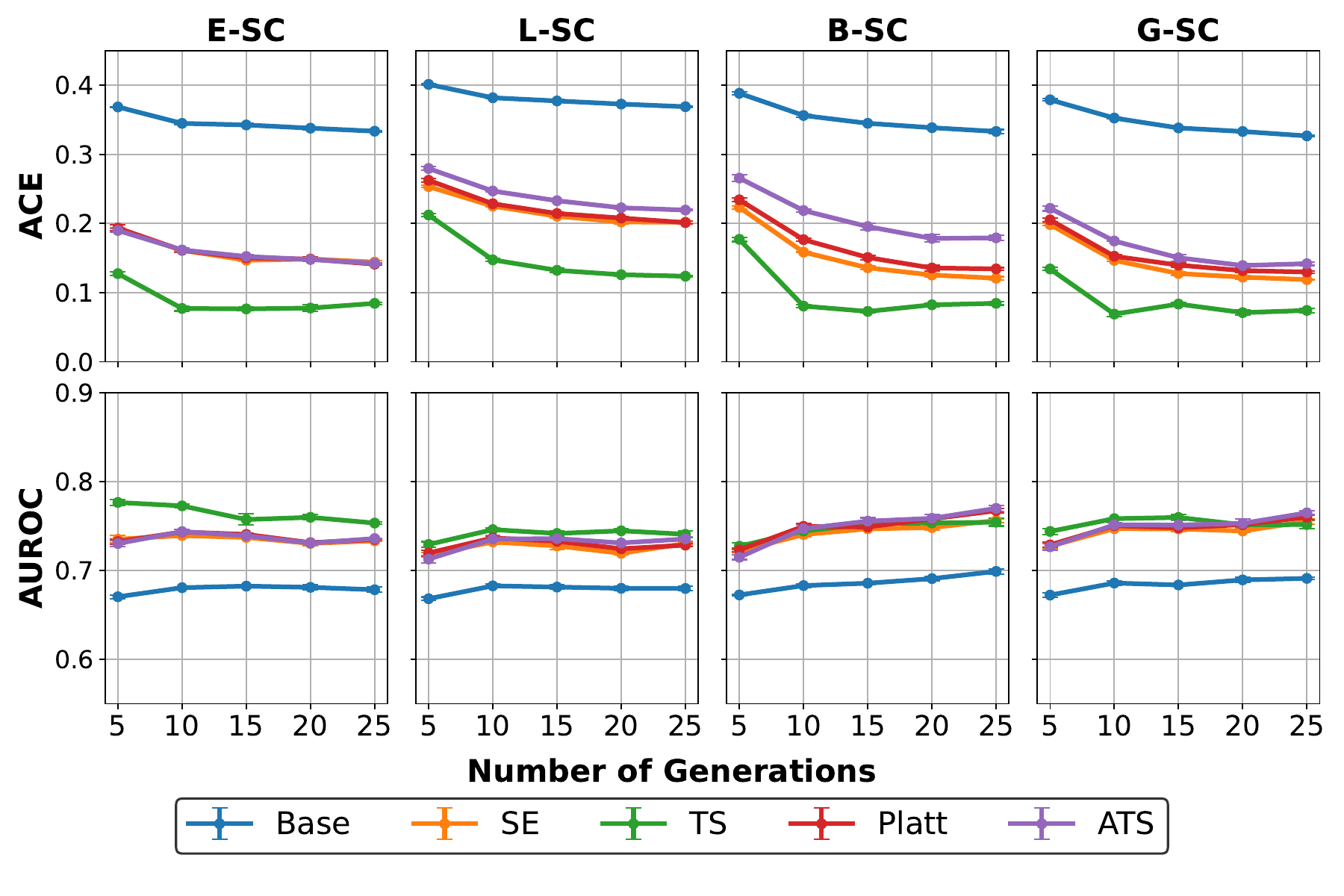}
    \vspace{-14pt}
    \caption{\textbf{SC Measures over Varying Numbers of Samples for Llama.} Mean $\ace$ ($\downarrow$) and AUROC ($\uparrow$) on the NQ dataset over varying number of sample generations per example. Error bars show one standard error across runs.}
    \label{fig:gen-ablation-plots}
\end{figure}

\Cref{fig:gen-ablation-plots} shows AUROC and \ace\ results for the Llama model on NQ as the number of sampled generations per example varies from 5 to 25. AUROC (bottom row) remains largely stable across sample sizes, indicating that varying the number of sampled generations leads to minor variation in discrimination across measures. Calibration (\ace, top row) improves as the number of samples increases, with gains beginning to saturate beyond roughly 10 samples. Across all sample sizes and measures, we find that TS enhances semantic UQ, primarily by improving calibration. These results justify our use of 10 generations throughout the paper, consistent with prior work \citep{kuhn2023semantic}.

\begin{takeaway}
Semantic calibration improves with sample size. Temperature scaling consistently boosts performance across all sample counts.
\end{takeaway}

\section{\MakeUppercase{Discussion}} \label{sec:discussion}
We demonstrated that existing semantic confidence measures are systematically miscalibrated. Optimising a single scalar temperature parameter (TS), whilst being surprisingly simple, computationally cheap and easy to implement, consistently improved both semantic calibration and discrimination. Crucially, this enhancement extends to downstream semantic entropy. Simple token-level temperature scaling outperformed both heuristic fixed temperatures and expressive token-level methods like ATS and Platt scaling. Our findings represent a meaningful update of the seminal work by \citet{guo2017calibration}, linking token-level classification to semantic prediction space in modern autoregressive transformer models in the generative QA domain. 

The effectiveness of TS is driven by its global inductive bias. Unlike expressive, computationally expensive methods such as ATS \citep{xie2024calibrating} which optimise for localised, token-specific calibration goals, TS is constrained to a single global scalar for entire sequences. This global constraint regularises against overly specific token-level adjustments that can overfit semantically irrelevant or filler words to reduce calibration loss. We find that this sequence-level bias transfers more reliably to semantic calibration than token-local methods. Overall, the choice of calibration method has a larger impact on performance than the specific form of the semantic measure itself. 

Our analysis focused on short-form generative QA where correctness admits a clear binary definition. Extending semantic calibration to tasks with partially correct outputs, such as summarisation, remains challenging as the notion of calibration is less well-defined in such settings \citep{wei2024long}. Developing principled evaluation frameworks for semantic calibration beyond binary correctness is a direction for future research

\vspace{-3pt}
\section{\MakeUppercase{Conclusion}}
\vspace{-3pt}
We systematically evaluated semantic calibration and discrimination, showing that base models and fixed-temperature heuristics produce miscalibrated and poorly discriminative semantic confidence estimates. We demonstrated that optimising a single scalar temperature parameter consistently improves calibration, discrimination, and downstream semantic entropy, outperforming both heuristic baselines and more sophisticated token-level methods. Consequently, temperature scaling offers a surprisingly simple, cheap, plug-and-play solution for practitioners to enhance semantic reliability without altering existing workflows.


\section*{Acknowledgements}
TGJR acknowledges support from the Foundational Research Grants program at Georgetown University's Center for Security and Emerging Technology.


\bibliography{refs}
\bibliographystyle{apalike}

\section*{Checklist}

\begin{enumerate}

  \item For all models and algorithms presented, check if you include:
  \begin{enumerate}
    \item A clear description of the mathematical setting, assumptions, algorithm, and/or model. [Yes]
    \item An analysis of the properties and complexity (time, space, sample size) of any algorithm. [Yes]
    \item (Optional) Anonymized source code, with specification of all dependencies, including external libraries. [Yes]
  \end{enumerate}

  \item For any theoretical claim, check if you include:
  \begin{enumerate}
    \item Statements of the full set of assumptions of all theoretical results. [Not Applicable]
    \item Complete proofs of all theoretical results. [Not Applicable]
    \item Clear explanations of any assumptions. [Not Applicable]     
  \end{enumerate}

  \item For all figures and tables that present empirical results, check if you include:
  \begin{enumerate}
    \item The code, data, and instructions needed to reproduce the main experimental results (either in the supplemental material or as a URL). [Yes]
    \item All the training details (e.g., data splits, hyperparameters, how they were chosen). [Yes]
    \item A clear definition of the specific measure or statistics and error bars (e.g., with respect to the random seed after running experiments multiple times). [Yes]
    \item A description of the computing infrastructure used. (e.g., type of GPUs, internal cluster, or cloud provider). [Yes]
  \end{enumerate}

  \item If you are using existing assets (e.g., code, data, models) or curating/releasing new assets, check if you include:
  \begin{enumerate}
    \item Citations of the creator If your work uses existing assets. [Yes]
    \item The license information of the assets, if applicable. [Yes]
    \item New assets either in the supplemental material or as a URL, if applicable. [Yes]
    \item Information about consent from data providers/curators. [Not Applicable]
    \item Discussion of sensible content if applicable, e.g., personally identifiable information or offensive content. [Not Applicable]
  \end{enumerate}

  \item If you used crowdsourcing or conducted research with human subjects, check if you include:
  \begin{enumerate}
    \item The full text of instructions given to participants and screenshots. [Not Applicable]
    \item Descriptions of potential participant risks, with links to Institutional Review Board (IRB) approvals if applicable. [Not Applicable]
    \item The estimated hourly wage paid to participants and the total amount spent on participant compensation. [Not Applicable]
  \end{enumerate}

\end{enumerate}

\clearpage
\onecolumn
\input{supplement}

\end{document}

%% file: supplement.tex
\begin{appendices}
\section*{\LARGE \centering Appendix
}
\section{{Dataset Splits, Hyperparameters, and Computational Resources}}
\label{sec:training details}

\subsection{Dataset Splits} \label{sec:dataset splits}  
Our calibration pipeline is organised into three stages: calibration training, calibration validation, and final test evaluation. Each stage relies on dedicated dataset splits to ensure that training, model selection, and evaluation remain strictly separated.  

\begin{itemize}  
    \item \textbf{Calibration Training.} The data is first divided into calibration-training and calibration-validation subsets. Using the pre-trained base LMs, we fit calibration parameters on the calibration-training split. Specifically, we optimise either a scalar temperature parameter or Platt scaling or temperature head parameters for 2 epochs.  

    \item \textbf{Calibration Validation.} The held-out calibration-validation split is used to select the best-performing hyperparameters from the calibration training stage. 

    \item \textbf{Final Test Evaluation.} Finally, we evaluate the selected models on a held-out test set. Results are reported on this dataset across methods, SC measures, and evaluation metrics (e.g., \ace and AUROC). For each setting, we include error bars to reflect performance variability and enable fair comparisons across both methods and measures.  
\end{itemize}  

To ensure fair cross-dataset comparison, we constrain the size of each split to be approximately matched across datasets. The exact split sizes for each dataset are provided in \Cref{tab:dataset split sizes}.  

\begin{table}[ht!]
\centering
\caption{\textbf{Dataset Split Sizes.} For each dataset, we report the number of examples used at different stages of the calibration pipeline: calibration training, calibration validation, and final test evaluation. Split sizes are matched across datasets to ensure fair comparison.}
\label{tab:dataset split sizes}
\begin{tabular}{clccc}
\toprule
\textbf{Stage} & \textbf{Split} & \textbf{TriviaQA} & \textbf{Natural Questions} & \textbf{SQuAD} \\
\midrule
\multirow{2}{*}{\textbf{Calibration}} & Training & 59374 &61600  & 59577 \\
 & Validation & 2000 &  2000   & 2000 \\
\midrule
\textbf{Final Evaluation} & Test & 2000 & 2000  & 2000 \\
\bottomrule
\end{tabular}
\end{table}

\subsection{Hyperparameter settings} \label{sec:hyperparameters}
Below we list the hyperparameter settings swept over for calibration optimisation. All models are trained using the AdamW optimiser \citep{loshchilov2017decoupled} with a cosine-annealing learning rate scheduler and a linear warm-up over the first $10\%$ of training examples within the first epoch of training \citep{radford2018improving}. Unless otherwise specified, all methods use the following shared hyperparameters: number of epochs $\{2\}$, calibration loss $\{\ell_\text{NLL}, \ell_\text{SS}\}$, and sweeps over the SS-loss weight $\alpha \in \{0.1, 0.25, 0.5, 0.75\}$ as well as G-SC and T-SC parameter $\alpha \in \{0.5, 0.75, 1.25\}$.

\textbf{Optimised Temperature Scaling (TS).}  
For scalar temperature scaling, we sweep over:
\begin{itemize}
\setlength{\itemsep}{0.5pt}
    \item Learning rate: $\{10^{-4}\}$.
    \item Initial temperature $\tau$: $\{1.0\}$.
\end{itemize}

\textbf{Adaptive Temperature Scaling (ATS).}  
For adaptive temperature head optimisation, we sweep over:
\begin{itemize}
\setlength{\itemsep}{0.5pt}
    \item Learning rate: $\{10^{-5}\}$ (smaller for stability).
    \item Weight decay: $\{0.0, 0.01\}$.
    \item Gradient clipping (max norm): $\{1.0\}$ (again for stability).
    \item Temperature head architecture: a single transformer block from the LLaMA-2 family, following \citet{xie2024calibrating}.
\end{itemize}

\textbf{Platt Scaling (PS).}  
For Platt scaling calibration, we sweep over:
\begin{itemize}
\setlength{\itemsep}{0.5pt}
    \item Learning rate: $\{10^{-5}\}$.
    \item Weight decay: $\{0.0, 0.01\}$.
    \item Gradient clipping (max norm): $\{1.0\}$.
    \item Transformation: affine mapping constrained to be diagonal, as in \citet{xie2024calibrating}, to reduce vocabulary-scale cost.
\end{itemize}

\paragraph{Hyperparameter Selection.}  
Hyperparameters are selected independently for each SC measure based on Brier score on the calibration-validation set. We selected based on Brier score to balance calibration and discrimination whilst avoiding degenerate selection that can arise from optimising solely for calibration.

\subsection{Computational Resources} \label{sec:computational-resources}

All models were trained and evaluated on our internal cluster equipped with NVIDIA A40 GPUs. Calibration training for each method was performed on a single GPU. During evaluation, we executed model inference and likelihood computations on one GPU, while semantic clustering using an NLI model was performed concurrently on a separate GPU. This setup ensures efficient parallelisation of the evaluation pipeline while maintaining manageable memory and runtime requirements.

\newpage

\section{{Evaluation}}
\subsection{Evaluation Metrics} \label{sec:metrics}
Below we detail the evaluation metrics that we report for our results presented in both the main paper and within the supplementary material. We frame these metrics in the context of representing an NLG task, specifically QA within this paper, as a binary task, with model's predictions being correct $c=1$ or incorrect $c=0$. 

Let $(\bx, \by) \sim p_\mathcal{D}$ be a dataset sample, where $\bx \in \mathcal{V}^*$ is the input and $\by \in \mathcal{V}^*$ is the ground-truth label. Let $p$ be a language model, and let $\hat{\by} \sim p(\cdot \mid \bx)$ be a model's prediction conditioned on the input $\bx$. The correctness of a model prediction $\hat{\by}$ is a function of the input, the ground-truth label, and the prediction itself, denoted as $c(\bx, \by, \hat{\by}) \in \{0,1\}$.

\paragraph{Expected Calibration Error (ECE)}
Calibration errors aim to quantify the misalignment between a model's predicted confidence and its actual correctness. The (L1) expected calibration error is defined as:
\begin{equation} \label{eqn:ece}
 \text{ECE} = \mathbb{E}_{\bx \sim p_{\mathcal{D}}} \left[ \left| \mathbb{P} (c = 1\mid p(\hat{\by} \mid \bx) = p) - p \right| \right].
\end{equation}
Following \citet{naeini2015obtaining}, ECE is empirically estimated by binning predictions from a dataset sample into $M$ intervals, denoted $\{B_m\}_{i=1}^M$. The weighted average of the absolute accuracy-confidence difference is then computed over the bins to estimate the expectation in \Cref{eqn:ece}:
\begin{equation*}
 \text{ECE} \approx \sum_{m=1}^M \frac{|B_m|} {n} |\text{acc}(B_m) - \text{conf}(B_m)|,
\end{equation*}
where:
\begin{equation*}
\mathrm{conf}(B_m)
= \frac{1}{\lvert B_m\rvert}\sum_{i \in B_m} p(\hat{\by}_i \mid \bx_i), \quad \mathrm{acc}(B_m)
= \frac{1}{\lvert B_m\rvert}\sum_{i \in B_m} c\bigl(\bx_i,\by_i,\hat{\by}_i\bigr).
\end{equation*}
We list two separate estimates of this metric, which arise from different binning schemes for confidence values within the $[0,1]$ interval:
\begin{itemize}
\item \textbf{$\widehat{\text{ECE}}$:} An even partition where each bin has an equal range provides the standard empirical approximation of the ECE.
 \item \textbf{$\widehat{\text{ACE}}$:} The empirical \emph{adaptive calibration error} \citep{nixon2019measuring} is obtained using a binning scheme where each of the $M$ bins contains an equal number of examples, with the bin ranges varying to accommodate this constraint.
\end{itemize}
\ace is generally considered a more robust metric than  
$\widehat{\text{ECE}}$ because it mitigates issues with uneven sample sizes across bins, which can lead to unreliable and noisy bin estimates in sparse regions of the confidence spectrum \citep{nixon2019measuring}. Therefore, throughout this work, we report the \ace.

\paragraph{AUROC}  
The Area Under the Receiver Operating Characteristic Curve (AUROC) measures how well confidence scores distinguish between correct and incorrect responses. The ROC curve is constructed by varying a confidence threshold $\lambda$ and plotting the true positive rate (TPR) against the false positive rate (FPR) at each threshold \citep{bradley1997use}. At its core, the AUROC value represents the probability that a randomly chosen correct response will receive a higher confidence score than a randomly chosen incorrect response \citep{bradley1997use}.

Given a dataset of $N$ examples with inputs $\bx_i \in \mathcal{V}^*$, model-generated responses $\hat{\by}_i \in \mathcal{V}^*$, correctness indicators $c_i = \bm{1}(\hat{\by}_i \sim \by_i \mid \bx)$, and model confidence scores $p(\hat{\by}_i \mid \bx_i)$ in response $\hat{\by}_i$, we define:
\begin{align*}
    \text{TPR}(\lambda) = \frac{\sum_{i: c_i = 1} \bm{1}(p(\hat{\by}_i \mid \bx_i) \geq \lambda)}{\sum_{i: c_i = 1} 1}, \quad
    \text{FPR}(\lambda) = \frac{\sum_{i: c_i = 0} \bm{1}(p(\hat{\by}_i \mid \bx_i) \geq \lambda)}{\sum_{i: c_i = 0} 1}.
\end{align*}

AUROC is then computed as:
\begin{equation*}
    \text{AUROC} = \int_0^1 \text{TPR}(\lambda) \ d\text{FPR}(\lambda),
\end{equation*}

A higher AUROC indicates better uncertainty quantification through the lens of ranking, where a model's confidence scores can better discriminate between correct and incorrect predictions \citep{bradley1997use}. An AUROC of 0.5 corresponds to a confidence metric that is no better than random guessing. An AUROC below 0.5 indicates that the metric is inverted, in the sense that it systematically assigns higher scores to incorrect predictions than to correct ones.

\paragraph{Brier Score (BS)}  
The Brier Score (BS) is a proper scoring rule that evaluates both calibration and correctness \citep{murphy1973new}. It is defined and decomposes into interpretable terms as follows:
\begin{align*}
\text{BS} &= \mathbb{E}_{(\mathbf{x}, \by) \sim p_{\mathcal{D}}} \big[ (p(\hat{\mathbf{y}} \mid \mathbf{x}) - c)^2 \big] \\
&= \underbrace{\mathbb{E}_{\mathbf{x} \sim p_{\mathcal{D}}} \big[ (p(\hat{\mathbf{y}} \mid \mathbf{x}) - \mathbb{P}(c = 1 \mid \mathbf{x}))^2 \big]}_{\text{Calibration}}
+ \underbrace{\mathbb{E}_{\mathbf{x} \sim p_{\mathcal{D}}} \mathbb{V}[c \mid \mathbf{x}]}_{\text{Refinement}} \\
&= \underbrace{\mathbb{E}_{\mathbf{x} \sim p_{\mathcal{D}}} \big[ (p(\hat{\mathbf{y}} \mid \mathbf{x}) - \mathbb{P}(c = 1 \mid \mathbf{x}))^2 \big]}_{\text{Calibration}}
+ \underbrace{\mathbb{V}[c]}_{\text{Uncertainty}} - \underbrace{\mathbb{V}_{\mathbf{x} \sim p_{\mathcal{D}}}[\mathbb{E}[c \mid \mathbf{x}]]}_{\text{Resolution}}.
\end{align*}
The last line uses the law of total variance to split the refinement term into uncertainty and resolution components. Here, $c$ is a Bernoulli random variable due to being either 0 or 1, which allows simple interchange of probability and expectation.

We can interpret the three terms as follows: \emph{calibration} measures the agreement between the model's predicted probabilities and the true frequencies of outcomes; \emph{resolution} quantifies how much the true outcome probabilities vary across inputs, rewarding models that assign distinct probabilities to different cases (and is closely related to discrimination); and finally, \emph{uncertainty} represents the inherent, irreducible variance in the data, which cannot be reduced by the model.

\paragraph{CORP-MCB.} The CORP methodology \citep{dimitriadis2021stable} provides an alternative framework that addresses limitations of standard binning algorithms like $\widehat{\text{ECE}}$ and $\ace$ through a non-parametric, isotonic regression approach.

Given a calibration set $\mathcal{D} = \{(p_1, c_1), \ldots, (p_N, c_N)\}$, where $p_i$ is the model's predicted probability for input $x_i$ and $c_i \in \{0,1\}$ denotes the correctness of the response $\hat{y}_i$, we estimate the conditional event probability (CEP), $\mathbb{P}(c=1 \mid p)$, using isotonic least-squares regression:
\begin{equation} \label{eqn:isotonic}
    f_* = \argmin_{f \in \mathcal{M}} \sum_{i=1}^N(f(p_i) - c_i )^2,
\end{equation}
where $\mathcal{M} \coloneqq \{ f: [0,1] \to [0,1] \mid f(p_1) \le f(p_2) \le \dots \le f(p_N) \}$ denotes the set of monotonic functions mapping forecasts in $[0,1]$ to recalibrated forecasts in $[0,1]$. For each input forecast $p_i$, the recalibrated forecast is $\tilde{p}_i = f_*(p_i)$. The optimisation problem in \cref{eqn:isotonic} is solved non-parametrically using the pool-adjacent-violators (PAV) algorithm \citep{de2010isotone}, which runs in $\mathcal{O}(N)$.

Given a score function $S: [0,1] \times \{0,1\} \to \mathbb{R}_{\ge 0}$, we define the following empirical scores over the dataset $\mathcal{D}$:
\begin{equation*}
    S_p = \frac{1}{N} \sum_{i=1}^N S(p_i, c_i ), \quad 
    S_{\tilde{p}} = \frac{1}{N} \sum_{i=1}^N S(\tilde{p}_i, c_i ), \quad 
    S_r = \frac{1}{N} \sum_{i=1}^N S(r, c_i ),
\end{equation*}
where $S_p, S_{\tilde{p}}, S_r$ denote the empirical scores of the original forecasts, recalibrated forecasts, and a constant reference forecast $r$, respectively. The decomposition of $S_p$ is given by:
\begin{equation*}
    S_p  = \underbrace{(S_p  - S_{\tilde{p}})}_{\text{MCB}} - \underbrace{(S_r - S_{\tilde{p}})}_{\text{DSC}} + \underbrace{S_r}_{\text{UNC}}.
\end{equation*}

Here, the terms correspond to: \emph{MCB}, a miscalibration term that is non-negative and zero when predictions are perfectly calibrated; \emph{DSC}, a discrimination term measuring the ability of the model to separate correct from incorrect outcomes; and \emph{UNC}, the inherent uncertainty in the data. The MCB term provides an additional quantitative measure of calibration.

We use the Brier score as our score function $S$, as it is a proper scoring rule. Following \citet{dimitriadis2021stable}, the constant reference forecast is taken as the empirical correctness $r = \frac{1}{N} \sum_{i=1}^N c_i$. This choice, together with using PAV-transformed recalibrated predictions, satisfies the calibration condition specified in their work (c.f. Equation 4).

\paragraph{Selective Prediction via Selective Accuracy}
\label{sec:selective-prediction-app}
Fix a threshold $\eta \in [0,1]$ and define the \emph{coverage set}:
\[
S_{\eta} = \bigl\{ \mathbf{x} \in \mathcal{V}^* \;\big|\; p(\hat{\mathbf{y}} \mid \mathbf{x}) \ge \eta , \; \text{for } \hat{\by} \sim p(\cdot \mid \bx)\bigr\}.
\]
This is the set of examples for which the model's confidence exceeds the threshold $\eta$. The \emph{selective accuracy} is then the average correctness over this set:
\[
\mathcal{A}_{\mathrm{sel}}(\eta) = \frac{1}{|S_{\eta}|} \sum_{\mathbf{x} \in S_{\eta}} c(\mathbf{x}, \mathbf{y}, \hat{\mathbf{y}}).
\]

Varying $\eta$ traces out a selective accuracy curve, showing how model accuracy changes as we focus on increasingly confident predictions. For a well-calibrated confidence score, one expects $\mathcal{A}_{\mathrm{sel}}(\eta)$ to generally increase as lower-confidence examples are filtered out.

\subsection{Evaluation of Accuracy} \label{sec: evaluation of accuracy}

To evaluate accuracy in generative QA tasks, we apply a multi-step procedure that balances correctness assessment with computational efficiency and overall cost:

\begin{itemize}
    \item \textbf{Initial Text Cleaning:} The model's response is preprocessed by discarding any extraneous text beyond the first direct answer to the question.
    \item \textbf{Direct Answer Matching:} If any reference answer is present verbatim in the response, the response is considered correct.
    \item \textbf{Fuzzy Matching:} When a direct match is absent, we apply fuzzy string matching \citep{bachmann2021rapidfuzz} using string-distance metrics. Responses exceeding a similarity threshold of 90 are classified as correct.
    \item \textbf{SQuAD F1 Evaluation:} For remaining unmatched responses, we compute the SQuAD-F1 score. Responses with F1 above 50.0 are considered correct as done by \citet{farquhar2024detecting}.
\end{itemize}

An alternative evaluation approach could involve using a model such as Llama-3.1 \citep{dubey2024llama} or GPT-4 \citep{achiam2023gpt} as a judge to assess equivalence with reference answers. However, this introduces additional cost and latency. Our current methodology follows prior work \citep{kuhn2023semantic, farquhar2024detecting}, which relies on token-level matching and has been shown to be effective in practice for short-form response QA tasks such as those that we work with in this paper. Unlike \citep{farquhar2024detecting}, we find that combining multiple accuracy checks beyond SQuAD-F1 is necessary to reduce both false positives and negatives in correctness assessment.

\subsection{Practical Considerations for Semantic Clustering and Evaluation Accuracy in a Semantic Uncertainty Pipeline}
\label{sec:practical-issues}

We note that our clustering procedure is identical to that of prior work \citep{kuhn2023semantic, farquhar2024detecting}, and our complete evaluation of correctness pipeline is detailed in \Cref{sec: evaluation of accuracy}. As discussed in the aforementioned section, while prior work often adopts a single correctness criterion, we find that relying on only one measure is insufficient to reliably judge answer correctness. In addition, we now  identify several practical pitfalls that are not adequately addressed as far as we can tell in the existing literature that affect both accuracy metrics and semantic clustering via NLI models. Below, we highlight these issues and our remedies.

\begin{itemize}
  \item \textbf{Clustering Numeric Responses.}  
    Off‐the‐shelf NLI models frequently fail to recognise equivalence between numeric and verbal forms of answers (e.g.\ “20” vs.\ “twenty”). Without intervention, this leads to over‐clustering.  
    \emph{Remedy:} we normalise all numeric responses to their digit form before clustering (e.g.\ “twenty” $\to$ “20”).

  \item \textbf{Clustering and Evaluating Dates.}  
    Standard NLI‐based criteria struggle with varied date formats and often produce false positives for nearby dates (e.g.\ “20th December 1988” vs.\ “19/12/1988”). Moreover, the standard components of the evaluation criteria discussed in \Cref{sec: evaluation of accuracy} also penalise model outputs that consist of full dates when the ground truth answers provide only a year.  
    \emph{Remedy:} we first parse each date into an ISO‐8601 string (YYYY-MM-DD). Then, for cases where only a year is required, we accept any normalised date within that year. Finally, we apply a hierarchical correctness check: exact match on year, month, and day only if the ground truth specifies those levels of granularity.

\end{itemize}

If unaddressed, these issues can confound both semantic clustering and evaluation scores. By applying normalisation and hierarchical checking for numbers and dates, we ensure that our reported metrics reflect true semantic correctness rather than artifacts of formatting.

\subsection{Evaluation Computational Complexity} \label{sec:computational-complexity}

Let \(M \in \mathbb{N}\) denote the number of generations sampled per input, and let \(L \in \mathbb{Z}\) denote the average sequence length. The overall time complexity of our evaluation pipeline decomposes into the following steps:

\begin{itemize}
  \item \textbf{Sample generation.}  
    Generate \(M\) samples via multinomial sampling from the autoregressive LM. Using KV caching, each forward step requires retrieving prior keys and values, yielding a total cost of \(\mathcal{O}(ML)\). This step is the dominant cost as it is inherently sequential, requiring \(L\) separate model invocations.

  \item \textbf{Text normalisation.}  
    Normalise each sample to extract the model’s final answer, discarding extraneous generated content. This step is negligible in cost.

  \item \textbf{Likelihood computation.}  
    Compute the length-normalised log-likelihood of each normalised sample under the model, requiring an additional forward pass. This also scales as \(\mathcal{O}(ML)\), but is substantially cheaper than generation because the entire sequence is processed in a single parallel pass. 

  \item \textbf{Clustering.}  
    Perform semantic clustering by running \(\binom{M}{2} = \mathcal{O}(M^2)\) pairwise NLI comparisons to determine entailment relations between responses. Although quadratic in \(M\), this step is far cheaper than LM generation or likelihood scoring, since it uses a much smaller NLI model. 

  \item \textbf{Semantic confidence calculation.}  
    Compute semantic confidence measures using the cluster structure and log-likelihoods. This is negligible compared to the other steps.
\end{itemize}

In practice, the computational cost is dominated by autoregressive sample generation, which scales as \(\mathcal{O}(ML)\) with KV caching. Semantic clustering is formally quadratic in the number of samples (\(\mathcal{O}(M^2)\)), but is comparatively inexpensive because it relies on a much smaller NLI model and can be efficiently batched. Likelihood computation contributes modestly at \(\mathcal{O}(ML)\), while text normalisation and semantic-confidence calculations are negligible. Overall, a modest sample budget (\(M \approx 10\)) provides representative coverage while keeping runtime tractable; performance gains saturate around 10–15 samples (\Cref{sec: number-of-example-ablation}), consistent with prior observations \citep{kuhn2023semantic}.

\section{{On the Necessity of Jointly Evaluating Calibration and Discrimination in UQ}} \label{sec: Need to evaluate both calibration and discrimination}
As discussed in the introduction, a central contribution of this work is the joint evaluation of
\emph{discrimination} and \emph{calibration} for semantic uncertainty quantification (UQ). Prior work on semantic UQ has largely focused on discrimination, typically evaluating
confidence-derived quantities such as semantic entropy using ranking-based metrics (e.g.,
AUROC), while neglecting calibration. This is a fundamental limitation: calibration and
discrimination capture distinct aspects of predictive uncertainty, and strong performance on one
does not imply strong performance on the other. Consequently, discrimination, only evaluation
provides an incomplete—and potentially misleading, assessment of UQ reliability.

In this section, we present a simple illustrative example demonstrating the logical independence
of calibration and discrimination, motivating the need to evaluate both jointly. To the best of
our knowledge, this work is the first, in the context of semantic uncertainty quantification for
LLMs, to systematically assess both properties within a unified framework.

We consider a binary prediction setting consistent with the evaluation framework used throughout
this work. Let $\mathcal{D} = \{(x_i, y_i, c_i)\}_{i=1}^n$ denote a dataset of triples, where
$x_i \in \mathcal{V}^*$ is an input, $y_i \in \mathcal{V}^*$ is a model-generated output, and
$c_i \in \{0,1\}$ indicates whether $y_i$ is correct given $x_i$. The model assigns a confidence score to each prediction via $p(y_i \mid x_i)$.

For simplicity, consider two examples:
\[
\mathcal{D} = \{(x_1, y_1, 0), (x_2, y_2, 1)\},
\]
corresponding to one incorrect and one correct generated response.

\paragraph{Perfect Calibration $\centernot\implies$ Discrimination.}
Suppose the model assigns identical confidence to both predictions:
\begin{equation*}
    p(y_1 \mid x_1) = p(y_2 \mid x_2) = 0.5.
\end{equation*}
This model is perfectly calibrated, as the average predicted confidence matches the empirical
correctness rate. However, the confidence scores provide no discriminative signal, preventing the
model from ranking correct predictions above incorrect ones (e.g., AUROC $= 0.5$), despite perfect
calibration.

\paragraph{Perfect Discrimination $\centernot\implies$ Calibration.}
Now suppose the model assigns higher confidence to the correct prediction:
\begin{equation*}
    p(y_1 \mid x_1) = 0.9 \quad \text{and} \quad p(y_2 \mid x_2) = 0.99.
\end{equation*}
This yields perfect discrimination, as the correct prediction is always ranked above the incorrect
one. However, the model is poorly calibrated: the predicted confidence values do not reflect the
empirical correctness frequencies and are substantially overconfident.

These examples illustrate that calibration and discrimination are complementary but logically
independent. Consequently, evaluating uncertainty quantification methods solely via discrimination
is insufficient; both properties must be assessed jointly to obtain a reliable evaluation.

 \newpage
\section{{Prompts}}
\begin{figure}[ht!]
    \centering
    \includegraphics[width=1.0\linewidth]{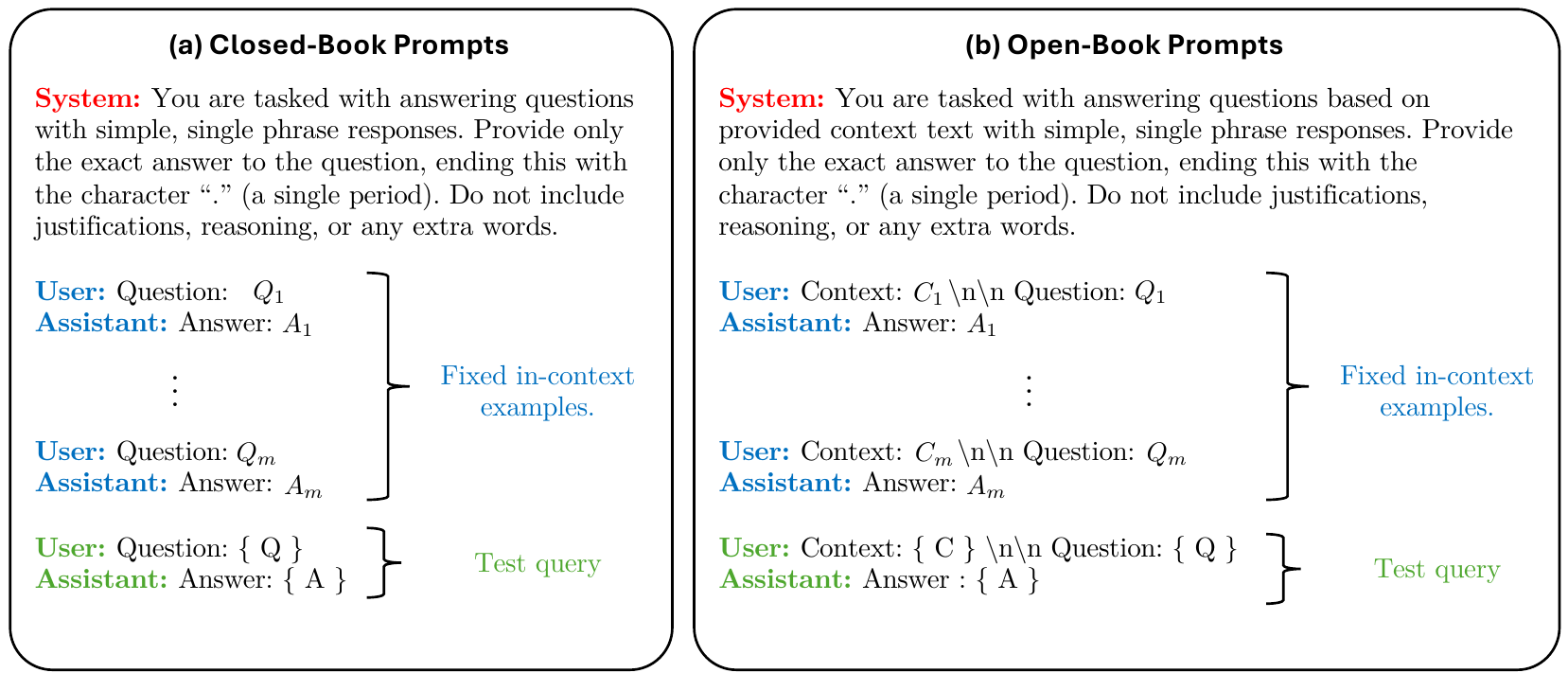}
    \caption{\textbf{Training and Evaluation Prompts.} Prompts applied for both calibration and evaluation across all methods, shown separately for (a) closed-book datasets (TriviaQA and Natural Questions) and (b) open-book datasets (SQuAD).}
    \label{fig:prompts}
\end{figure}

We present the exact prompts used for both calibration training and evaluation in \Cref{fig:prompts}. For closed-book datasets, we use $m=10$ in-context examples, while for open-book datasets, we use $m=4$. For each dataset, examples from the training set (which is not used within any part of our pipeline) are uniformly sampled to serve as in-context examples, and these examples are fixed across both training and evaluation for all methods. The use of in-context examples encourages the model to produce concise outputs, containing only the final response.

\section{{On the Standard Practice of Task-Specific Calibration}}
\label{app:discussion_calibration_practice}

In our evaluation, we adopt a protocol where calibration parameters such as scalar temperature parameters or transformer calibration heads are optimised separately for each model on every new dataset or domain. Regarding the practical application of this task-based, held-out calibration, we argue that this workflow aligns seamlessly with the standard paradigm of transfer learning, where pre-trained models are adapted to specific domains via learnable parameters. A prominent example of this paradigm is Low-Rank Adaptation (LoRA) \citep{hu2021lora}, where small datasets are used to adapt general models to specific domains. 

Task-specific temperature optimisation can be viewed as a lightweight, stripped-down version of this adaptation: instead of injecting new knowledge or capabilities, it adjusts the model's beliefs to align with the uncertainty of the new domain. Therefore, our approach fits naturally within modern deployment frameworks. Furthermore, prior work by \citet{desai2020calibration} demonstrates that pre-trained transformers require task-specific recalibration to remain reliable when shifting to new domains. This evidence suggests that post-hoc recalibration is not merely an optional step, but a necessity for reliably adapting pre-trained language models to downstream tasks. Consequently, we argue that optimising a scalar temperature on a small held-out set is a highly practical, computationally efficient, and methodologically sound procedure for current community practices.

\newpage

\section{{Supplementary Results}}
\subsection{Model Accuracies}
We report the accuracies of the base models on the held-out test sets for reference. These are the beam search results giving a indication of general model capabilities on the datasets before further calibrating model semantic confidence.  

\begin{table}[ht!]
\centering
\caption{\textbf{Base Model Accuracies (\%).} Test-set accuracies ($\uparrow$) of the base models through beam search.}
\label{tab:model-accuracies}
\begin{tabular}{cccc}
\toprule
\textbf{Model}  & \textbf{TriviaQA} & \textbf{Natural Questions} & \textbf{SQuAD} \\
\midrule
\multirow{1}{*}{Llama} & 71.9 & 42.5 & 95.5 \\

\multirow{1}{*}{Qwen}  & 53.0  & 33.2  & 94.4 \\

\multirow{1}{*}{Mistral}  & 64.8 & 33.2  & 94.2  \\

\bottomrule
\end{tabular}
\end{table}

\subsection{Comparison of Recalibration Loss Functions}
\begin{figure}[ht!]
  \centering
  \begin{subfigure}{0.495\textwidth}
    \centering
    \includegraphics[width=\textwidth]{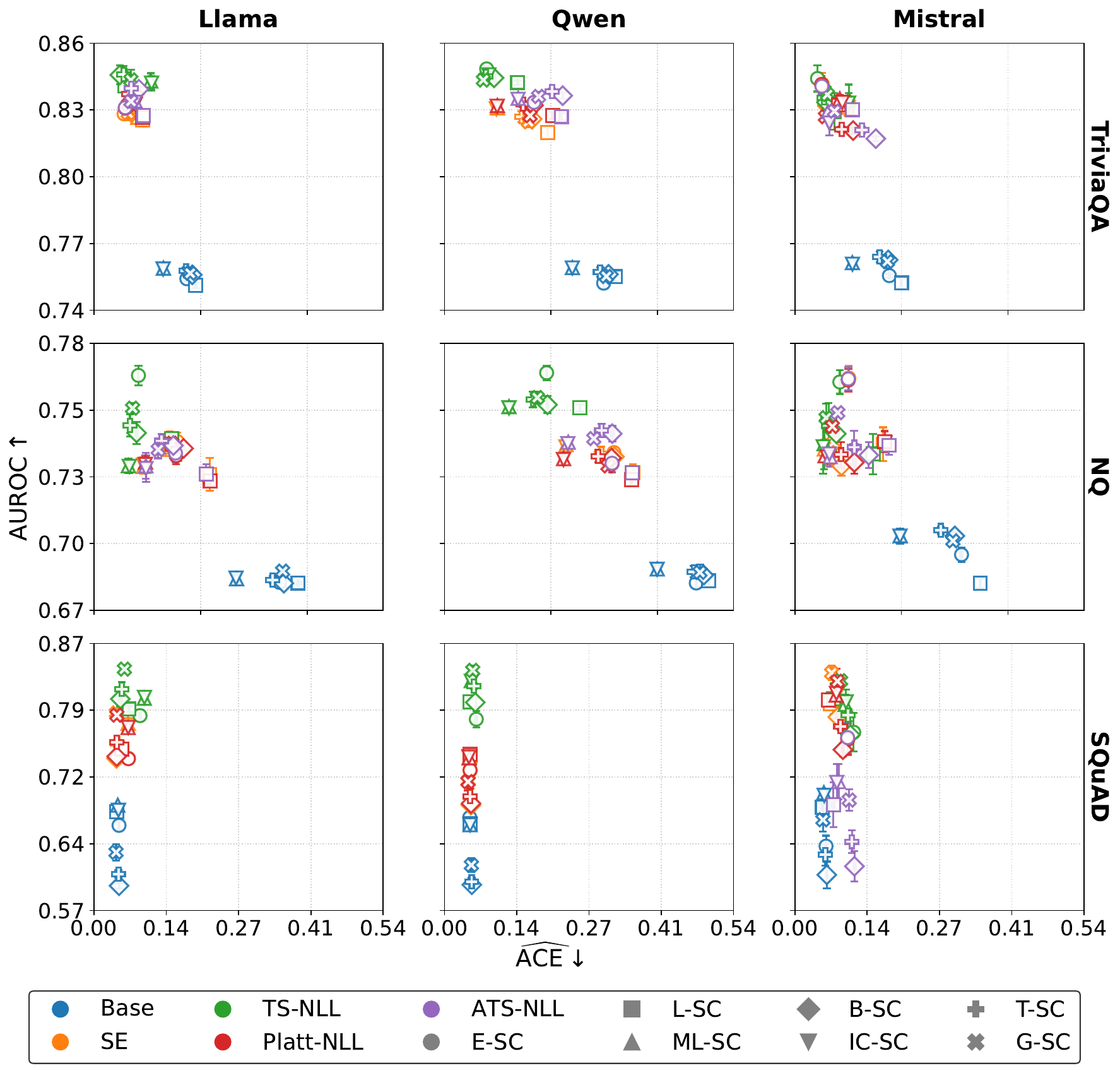}
    \caption{Calibration methods with NLL loss.}
    \label{fig:nll-uncertainty-results}
  \end{subfigure}
  \hfill
  \begin{subfigure}{0.495\textwidth}
    \centering
    \includegraphics[width=\textwidth]{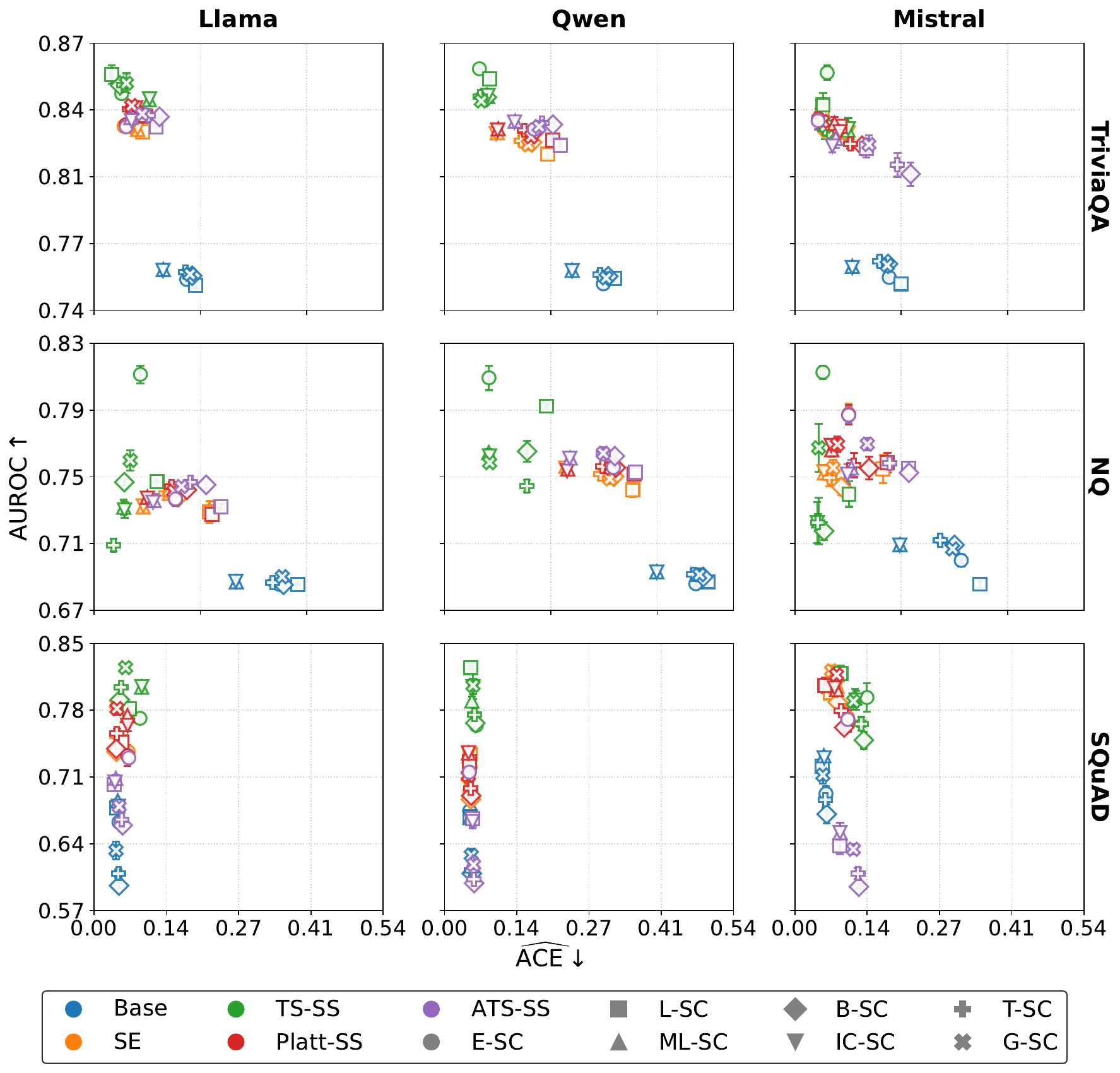}
    \caption{Calibration methods with SS loss.}
    \label{fig:ss-uncertainty-results}
  \end{subfigure}

  \caption{\textbf{Comparison of Uncertainty Metrics of Semantic Measures Across Calibration Methods Using SS and NLL Losses.} Mean and standard deviation across four inference runs of $\widehat{\text{ACE}}$ ($\downarrow$) and AUROC ($\uparrow$) scores for semantic confidence measures for each model across baselines and calibration methods. We compare using a (a) NLL and (b) SS calibration loss.}
  \label{fig:comparison-of-recalibration-losses}
\end{figure}

We compare the calibration and discrimination of models across semantic confidence measures, recalibration methods, and baselines, focusing on the calibration loss function used for recalibration. The results are shown in \Cref{fig:comparison-of-recalibration-losses}.

\subsection{Main Figure Results in Tabular Form}
To supplement the results shown in \Cref{fig:main results}, we include the same results in tabular form that allow for more fine grained and precise comparison between methods. We present these in \Cref{tab:main-llama-results}, \Cref{tab:main-qwen-results}, and \Cref{tab:main-mistral-results}.

\begin{table}[h!]
\footnotesize
\centering
\setlength{\tabcolsep}{0.5pt}
\caption{\textbf{Uncertainty Metrics for SC Measures Across Methods for Llama.} Mean and standard error of $\widehat{\text{ACE}}$ ($\downarrow$) and AUROC ($\uparrow$) scores for SC measures across  measures baseline and recalibration methods for Llama.}
\resizebox{\textwidth}{!}{%
\begin{tabular}{lcccccccc|ccccccc}
\toprule
 & & \multicolumn{7}{c}{$\widehat{\textbf{ACE}}$} & \multicolumn{7}{c}{\textbf{AUROC}} \\
 &  & E-SC & L-SC & ML-SC & B-SC & IC-SC & T-SC & G-SC & E-SC & L-SC & ML-SC & B-SC & IC-SC & T-SC & G-SC \\
\midrule
\multirow{5}{*}{\rotatebox{90}{\textbf{TriviaQA}}} & $\text{Base}$ & $0.184_{{\pm 0.002}}$ & $0.174_{{\pm 0.002}}$ & $0.180_{{\pm 0.001}}$ & $0.130_{{\pm 0.001}}$ & $0.191_{{\pm 0.002}}$ & $0.130_{{\pm 0.002}}$ & $0.172_{{\pm 0.002}}$ & $0.757_{{\pm 0.001}}$ & $0.755_{{\pm 0.001}}$ & $0.757_{{\pm 0.001}}$ & $0.759_{{\pm 0.001}}$ & $0.752_{{\pm 0.001}}$ & $0.760_{{\pm 0.001}}$ & $0.758_{{\pm 0.002}}$ \\
 & $\text{SE}$ & $0.070_{{\pm 0.003}}$ & $0.056_{{\pm 0.001}}$ & $0.064_{{\pm 0.003}}$ & $0.082_{{\pm 0.002}}$ & $0.092_{{\pm 0.002}}$ & $0.082_{{\pm 0.002}}$ & $0.066_{{\pm 0.003}}$ & $0.832_{{\pm 0.003}}$ & $0.832_{{\pm 0.002}}$ & $0.832_{{\pm 0.003}}$ & $0.831_{{\pm 0.002}}$ & $0.829_{{\pm 0.003}}$ & $0.830_{{\pm 0.002}}$ & $0.833_{{\pm 0.003}}$ \\
 & $\text{TS}$ & $\mathbf{0.049_{{\pm 0.001}}}$ & $\mathbf{0.055_{{\pm 0.001}}}$ & $\mathbf{0.061_{{\pm 0.001}}}$ & $0.104_{{\pm 0.004}}$ & $\mathbf{0.033_{{\pm 0.003}}}$ & $0.106_{{\pm 0.002}}$ & $\mathbf{0.056_{{\pm 0.004}}}$ & $\mathbf{0.852_{{\pm 0.003}}}$ & $\mathbf{0.849_{{\pm 0.004}}}$ & $\mathbf{0.853_{{\pm 0.005}}}$ & $\mathbf{0.846_{{\pm 0.003}}}$ & $\mathbf{0.858_{{\pm 0.005}}}$ & $\mathbf{0.847_{{\pm 0.004}}}$ & $\mathbf{0.853_{{\pm 0.003}}}$ \\
 & $\text{Platt}$ & $0.078_{{\pm 0.001}}$ & $0.060_{{\pm 0.002}}$ & $0.071_{{\pm 0.002}}$ & $0.078_{{\pm 0.002}}$ & $0.097_{{\pm 0.002}}$ & $0.077_{{\pm 0.001}}$ & $0.066_{{\pm 0.003}}$ & $0.840_{{\pm 0.002}}$ & $0.833_{{\pm 0.002}}$ & $0.842_{{\pm 0.001}}$ & $0.841_{{\pm 0.002}}$ & $0.837_{{\pm 0.002}}$ & $0.842_{{\pm 0.002}}$ & $0.840_{{\pm 0.002}}$ \\
 & $\text{ATS}$ & $0.084_{{\pm 0.001}}$ & $0.061_{{\pm 0.002}}$ & $0.069_{{\pm 0.002}}$ & $\mathbf{0.070_{{\pm 0.004}}}$ & $0.093_{{\pm 0.003}}$ & $\mathbf{0.069_{{\pm 0.005}}}$ & $0.070_{{\pm 0.002}}$ & $0.843_{{\pm 0.001}}$ & $0.832_{{\pm 0.002}}$ & $0.838_{{\pm 0.002}}$ & $0.835_{{\pm 0.001}}$ & $0.831_{{\pm 0.003}}$ & $0.836_{{\pm 0.001}}$ & $0.844_{{\pm 0.001}}$ \\
\midrule
\multirow{5}{*}{\rotatebox{90}{\textbf{NQ}}} & $\text{Base}$ & $0.356_{{\pm 0.002}}$ & $0.348_{{\pm 0.002}}$ & $0.354_{{\pm 0.002}}$ & $0.267_{{\pm 0.002}}$ & $0.383_{{\pm 0.001}}$ & $0.268_{{\pm 0.002}}$ & $0.335_{{\pm 0.002}}$ & $0.685_{{\pm 0.002}}$ & $0.685_{{\pm 0.002}}$ & $0.690_{{\pm 0.001}}$ & $0.687_{{\pm 0.001}}$ & $0.685_{{\pm 0.001}}$ & $0.687_{{\pm 0.002}}$ & $0.686_{{\pm 0.002}}$ \\
 & $\text{SE}$ & $0.158_{{\pm 0.002}}$ & $0.156_{{\pm 0.001}}$ & $0.142_{{\pm 0.003}}$ & $0.093_{{\pm 0.003}}$ & $0.217_{{\pm 0.002}}$ & $0.093_{{\pm 0.003}}$ & $0.135_{{\pm 0.002}}$ & $0.737_{{\pm 0.004}}$ & $0.741_{{\pm 0.004}}$ & $0.741_{{\pm 0.004}}$ & $0.732_{{\pm 0.003}}$ & $0.728_{{\pm 0.006}}$ & $0.731_{{\pm 0.003}}$ & $0.739_{{\pm 0.004}}$ \\
 & $\text{TS}$ & $\mathbf{0.057_{{\pm 0.002}}}$ & $\mathbf{0.087_{{\pm 0.001}}}$ & $\mathbf{0.068_{{\pm 0.002}}}$ & $\mathbf{0.057_{{\pm 0.002}}}$ & $\mathbf{0.118_{{\pm 0.002}}}$ & $\mathbf{0.056_{{\pm 0.002}}}$ & $\mathbf{0.036_{{\pm 0.003}}}$ & $\mathbf{0.746_{{\pm 0.002}}}$ & $\mathbf{0.810_{{\pm 0.005}}}$ & $\mathbf{0.759_{{\pm 0.006}}}$ & $0.730_{{\pm 0.005}}$ & $\mathbf{0.746_{{\pm 0.000}}}$ & $0.731_{{\pm 0.004}}$ & $0.709_{{\pm 0.004}}$ \\
 & $\text{Platt}$ & $0.168_{{\pm 0.001}}$ & $0.154_{{\pm 0.002}}$ & $0.146_{{\pm 0.001}}$ & $0.096_{{\pm 0.005}}$ & $0.218_{{\pm 0.000}}$ & $0.100_{{\pm 0.002}}$ & $0.138_{{\pm 0.001}}$ & $0.739_{{\pm 0.002}}$ & $0.736_{{\pm 0.004}}$ & $0.741_{{\pm 0.002}}$ & $\mathbf{0.732_{{\pm 0.002}}}$ & $0.726_{{\pm 0.002}}$ & $\mathbf{0.737_{{\pm 0.003}}}$ & $0.740_{{\pm 0.002}}$ \\
 & $\text{ATS}$ & $0.149_{{\pm 0.003}}$ & $0.153_{{\pm 0.002}}$ & $0.120_{{\pm 0.003}}$ & $0.097_{{\pm 0.004}}$ & $0.211_{{\pm 0.000}}$ & $0.097_{{\pm 0.002}}$ & $0.127_{{\pm 0.003}}$ & $0.740_{{\pm 0.003}}$ & $0.737_{{\pm 0.004}}$ & $0.738_{{\pm 0.003}}$ & $0.731_{{\pm 0.005}}$ & $0.728_{{\pm 0.004}}$ & $0.732_{{\pm 0.005}}$ & $\mathbf{0.741_{{\pm 0.003}}}$ \\
\midrule
\multirow{5}{*}{\rotatebox{90}{\textbf{SQuAD}}} & $\text{Base}$ & $0.047_{{\pm 0.001}}$ & $\mathbf{0.047_{{\pm 0.001}}}$ & $\mathbf{0.041_{{\pm 0.001}}}$ & $0.046_{{\pm 0.001}}$ & $0.043_{{\pm 0.001}}$ & $0.045_{{\pm 0.001}}$ & $0.046_{{\pm 0.001}}$ & $0.594_{{\pm 0.003}}$ & $0.662_{{\pm 0.005}}$ & $0.632_{{\pm 0.009}}$ & $0.679_{{\pm 0.002}}$ & $0.677_{{\pm 0.003}}$ & $0.685_{{\pm 0.002}}$ & $0.607_{{\pm 0.004}}$ \\
 & $\text{SE}$ & $\mathbf{0.042_{{\pm 0.002}}}$ & $0.065_{{\pm 0.001}}$ & $0.042_{{\pm 0.002}}$ & $0.064_{{\pm 0.001}}$ & $0.052_{{\pm 0.001}}$ & $0.064_{{\pm 0.001}}$ & $0.043_{{\pm 0.003}}$ & $0.737_{{\pm 0.004}}$ & $0.737_{{\pm 0.003}}$ & $0.788_{{\pm 0.004}}$ & $0.775_{{\pm 0.005}}$ & $0.748_{{\pm 0.006}}$ & $0.776_{{\pm 0.002}}$ & $0.754_{{\pm 0.004}}$ \\
 & $\text{TS}$ & $0.049_{{\pm 0.003}}$ & $0.086_{{\pm 0.002}}$ & $0.059_{{\pm 0.001}}$ & $0.090_{{\pm 0.002}}$ & $0.066_{{\pm 0.001}}$ & $0.090_{{\pm 0.002}}$ & $0.051_{{\pm 0.001}}$ & $\mathbf{0.804_{{\pm 0.008}}}$ & $\mathbf{0.773_{{\pm 0.005}}}$ & $\mathbf{0.827_{{\pm 0.002}}}$ & $\mathbf{0.807_{{\pm 0.005}}}$ & $\mathbf{0.783_{{\pm 0.003}}}$ & $\mathbf{0.806_{{\pm 0.005}}}$ & $\mathbf{0.806_{{\pm 0.002}}}$ \\
 & $\text{Platt}$ & $0.043_{{\pm 0.002}}$ & $0.063_{{\pm 0.003}}$ & $0.043_{{\pm 0.001}}$ & $0.063_{{\pm 0.001}}$ & $0.052_{{\pm 0.001}}$ & $0.064_{{\pm 0.002}}$ & $\mathbf{0.043_{{\pm 0.002}}}$ & $0.739_{{\pm 0.003}}$ & $0.73_{{\pm 0.01}}$ & $0.786_{{\pm 0.005}}$ & $0.765_{{\pm 0.006}}$ & $0.748_{{\pm 0.007}}$ & $0.772_{{\pm 0.003}}$ & $0.757_{{\pm 0.005}}$ \\
 & $\text{ATS}$ & $0.054_{{\pm 0.001}}$ & $0.065_{{\pm 0.001}}$ & $0.047_{{\pm 0.001}}$ & $\mathbf{0.039_{{\pm 0.001}}}$ & $\mathbf{0.038_{{\pm 0.001}}}$ & $\mathbf{0.041_{{\pm 0.001}}}$ & $0.052_{{\pm 0.001}}$ & $0.659_{{\pm 0.007}}$ & $0.731_{{\pm 0.004}}$ & $0.679_{{\pm 0.007}}$ & $0.705_{{\pm 0.005}}$ & $0.702_{{\pm 0.005}}$ & $0.709_{{\pm 0.004}}$ & $0.665_{{\pm 0.007}}$ \\
\bottomrule
\end{tabular}}%
\label{tab:main-llama-results}
\end{table}

\begin{table}[h!]
\footnotesize
\centering
\setlength{\tabcolsep}{0.5pt}
\caption{\textbf{Uncertainty Metrics for SC Measures Across Methods for Qwen.} Mean and standard error of $\widehat{\text{ACE}}$ ($\downarrow$) and AUROC ($\uparrow$) scores for SC measures across  measures baseline and recalibration methods for Qwen.}
\resizebox{\textwidth}{!}{%
\begin{tabular}{lcccccccc|ccccccc}
\toprule
 &  & \multicolumn{7}{c}{$\widehat{\textbf{ACE}}$} & \multicolumn{7}{c}{\textbf{AUROC}} \\
 &  & E-SC & L-SC & ML-SC & B-SC & IC-SC & T-SC & G-SC & E-SC & L-SC & ML-SC & B-SC & IC-SC & T-SC & G-SC \\
\midrule
\multirow{5}{*}{\rotatebox{90}{\textbf{TriviaQA}}} & $\text{Base}$ & $0.308_{{\pm 0.001}}$ & $0.299_{{\pm 0.001}}$ & $0.304_{{\pm 0.000}}$ & $0.240_{{\pm 0.000}}$ & $0.320_{{\pm 0.001}}$ & $0.240_{{\pm 0.000}}$ & $0.293_{{\pm 0.001}}$ & $0.763_{{\pm 0.002}}$ & $0.759_{{\pm 0.002}}$ & $0.762_{{\pm 0.002}}$ & $0.765_{{\pm 0.001}}$ & $0.761_{{\pm 0.002}}$ & $0.766_{{\pm 0.001}}$ & $0.764_{{\pm 0.002}}$ \\
 & $\text{SE}$ & $0.165_{{\pm 0.002}}$ & $0.159_{{\pm 0.001}}$ & $0.158_{{\pm 0.001}}$ & $0.097_{{\pm 0.002}}$ & $0.194_{{\pm 0.001}}$ & $0.099_{{\pm 0.002}}$ & $0.145_{{\pm 0.002}}$ & $0.832_{{\pm 0.002}}$ & $0.831_{{\pm 0.001}}$ & $0.830_{{\pm 0.002}}$ & $0.836_{{\pm 0.002}}$ & $0.825_{{\pm 0.002}}$ & $0.837_{{\pm 0.002}}$ & $0.833_{{\pm 0.002}}$ \\
 & $\text{TS}$ & $\mathbf{0.080_{{\pm 0.001}}}$ & $\mathbf{0.066_{{\pm 0.005}}}$ & $\mathbf{0.069_{{\pm 0.002}}}$ & $\mathbf{0.082_{{\pm 0.003}}}$ & $\mathbf{0.085_{{\pm 0.004}}}$ & $\mathbf{0.081_{{\pm 0.003}}}$ & $\mathbf{0.067_{{\pm 0.002}}}$ & $\mathbf{0.855_{{\pm 0.002}}}$ & $\mathbf{0.870_{{\pm 0.002}}}$ & $\mathbf{0.853_{{\pm 0.002}}}$ & $\mathbf{0.856_{{\pm 0.000}}}$ & $\mathbf{0.864_{{\pm 0.004}}}$ & $\mathbf{0.855_{{\pm 0.001}}}$ & $\mathbf{0.855_{{\pm 0.002}}}$ \\
 & $\text{Platt}$ & $0.170_{{\pm 0.001}}$ & $0.168_{{\pm 0.002}}$ & $0.163_{{\pm 0.002}}$ & $0.100_{{\pm 0.002}}$ & $0.203_{{\pm 0.001}}$ & $0.101_{{\pm 0.003}}$ & $0.150_{{\pm 0.001}}$ & $0.837_{{\pm 0.002}}$ & $0.839_{{\pm 0.001}}$ & $0.834_{{\pm 0.001}}$ & $0.838_{{\pm 0.001}}$ & $0.833_{{\pm 0.001}}$ & $0.839_{{\pm 0.001}}$ & $0.838_{{\pm 0.002}}$ \\
 & $\text{ATS}$ & $0.204_{{\pm 0.003}}$ & $0.168_{{\pm 0.002}}$ & $0.178_{{\pm 0.002}}$ & $0.132_{{\pm 0.003}}$ & $0.219_{{\pm 0.001}}$ & $0.133_{{\pm 0.002}}$ & $0.183_{{\pm 0.003}}$ & $0.841_{{\pm 0.001}}$ & $0.839_{{\pm 0.001}}$ & $0.840_{{\pm 0.002}}$ & $0.842_{{\pm 0.001}}$ & $0.833_{{\pm 0.002}}$ & $0.843_{{\pm 0.001}}$ & $0.842_{{\pm 0.001}}$ \\
\midrule
\multirow{5}{*}{\rotatebox{90}{\textbf{NQ}}} & $\text{Base}$ & $0.487_{{\pm 0.001}}$ & $0.473_{{\pm 0.001}}$ & $0.480_{{\pm 0.001}}$ & $0.399_{{\pm 0.001}}$ & $0.496_{{\pm 0.001}}$ & $0.399_{{\pm 0.001}}$ & $0.468_{{\pm 0.001}}$ & $0.690_{{\pm 0.003}}$ & $0.687_{{\pm 0.002}}$ & $0.692_{{\pm 0.004}}$ & $0.693_{{\pm 0.002}}$ & $0.688_{{\pm 0.003}}$ & $0.693_{{\pm 0.002}}$ & $0.692_{{\pm 0.003}}$ \\
 & $\text{SE}$ & $0.319_{{\pm 0.002}}$ & $0.318_{{\pm 0.002}}$ & $0.311_{{\pm 0.002}}$ & $0.228_{{\pm 0.002}}$ & $0.354_{{\pm 0.002}}$ & $0.228_{{\pm 0.002}}$ & $0.294_{{\pm 0.002}}$ & $0.745_{{\pm 0.003}}$ & $0.747_{{\pm 0.002}}$ & $0.744_{{\pm 0.003}}$ & $0.750_{{\pm 0.003}}$ & $0.738_{{\pm 0.004}}$ & $0.751_{{\pm 0.002}}$ & $0.747_{{\pm 0.003}}$ \\
 & $\text{TS}$ & $\mathbf{0.155_{{\pm 0.002}}}$ & $\mathbf{0.084_{{\pm 0.002}}}$ & $\mathbf{0.085_{{\pm 0.003}}}$ & $\mathbf{0.085_{{\pm 0.001}}}$ & $\mathbf{0.192_{{\pm 0.002}}}$ & $\mathbf{0.083_{{\pm 0.002}}}$ & $\mathbf{0.155_{{\pm 0.002}}}$ & $\mathbf{0.759_{{\pm 0.006}}}$ & $\mathbf{0.799_{{\pm 0.007}}}$ & $0.753_{{\pm 0.002}}$ & $\mathbf{0.756_{{\pm 0.002}}}$ & $\mathbf{0.784_{{\pm 0.002}}}$ & $\mathbf{0.758_{{\pm 0.002}}}$ & $0.740_{{\pm 0.003}}$ \\
 & $\text{Platt}$ & $0.314_{{\pm 0.001}}$ & $0.318_{{\pm 0.001}}$ & $0.307_{{\pm 0.001}}$ & $0.231_{{\pm 0.002}}$ & $0.357_{{\pm 0.000}}$ & $0.232_{{\pm 0.002}}$ & $0.289_{{\pm 0.001}}$ & $0.745_{{\pm 0.003}}$ & $0.751_{{\pm 0.005}}$ & $0.741_{{\pm 0.001}}$ & $0.748_{{\pm 0.001}}$ & $0.747_{{\pm 0.003}}$ & $0.749_{{\pm 0.001}}$ & $0.746_{{\pm 0.003}}$ \\
 & $\text{ATS}$ & $0.315_{{\pm 0.004}}$ & $0.318_{{\pm 0.001}}$ & $0.280_{{\pm 0.002}}$ & $0.236_{{\pm 0.004}}$ & $0.353_{{\pm 0.002}}$ & $0.237_{{\pm 0.003}}$ & $0.295_{{\pm 0.004}}$ & $0.756_{{\pm 0.004}}$ & $0.750_{{\pm 0.005}}$ & $\mathbf{0.754_{{\pm 0.001}}}$ & $0.755_{{\pm 0.002}}$ & $0.738_{{\pm 0.001}}$ & $0.755_{{\pm 0.001}}$ & $\mathbf{0.757_{{\pm 0.003}}}$ \\
\midrule
\multirow{5}{*}{\rotatebox{90}{\textbf{SQuAD}}} & $\text{Base}$ & $0.051_{{\pm 0.001}}$ & $0.047_{{\pm 0.000}}$ & $0.051_{{\pm 0.001}}$ & $0.048_{{\pm 0.000}}$ & $0.048_{{\pm 0.000}}$ & $0.048_{{\pm 0.000}}$ & $0.051_{{\pm 0.001}}$ & $0.534_{{\pm 0.004}}$ & $0.605_{{\pm 0.004}}$ & $0.554_{{\pm 0.007}}$ & $0.597_{{\pm 0.002}}$ & $0.598_{{\pm 0.003}}$ & $0.597_{{\pm 0.002}}$ & $0.537_{{\pm 0.004}}$ \\
 & $\text{SE}$ & $\mathbf{0.050_{{\pm 0.001}}}$ & $0.048_{{\pm 0.001}}$ & $0.044_{{\pm 0.001}}$ & $0.046_{{\pm 0.000}}$ & $0.048_{{\pm 0.002}}$ & $\mathbf{0.046_{{\pm 0.000}}}$ & $0.049_{{\pm 0.001}}$ & $0.618_{{\pm 0.007}}$ & $0.654_{{\pm 0.006}}$ & $0.643_{{\pm 0.009}}$ & $0.669_{{\pm 0.005}}$ & $0.669_{{\pm 0.005}}$ & $0.667_{{\pm 0.004}}$ & $0.627_{{\pm 0.008}}$ \\
 & $\text{TS}$ & $0.058_{{\pm 0.001}}$ & $0.060_{{\pm 0.001}}$ & $0.053_{{\pm 0.002}}$ & $0.051_{{\pm 0.001}}$ & $0.049_{{\pm 0.003}}$ & $0.052_{{\pm 0.002}}$ & $0.055_{{\pm 0.001}}$ & $\mathbf{0.726_{{\pm 0.005}}}$ & $\mathbf{0.702_{{\pm 0.004}}}$ & $\mathbf{0.760_{{\pm 0.005}}}$ & $\mathbf{0.748_{{\pm 0.005}}}$ & $\mathbf{0.767_{{\pm 0.002}}}$ & $\mathbf{0.729_{{\pm 0.005}}}$ & $\mathbf{0.743_{{\pm 0.003}}}$ \\
 & $\text{Platt}$ & $0.050_{{\pm 0.001}}$ & $\mathbf{0.047_{{\pm 0.000}}}$ & $\mathbf{0.044_{{\pm 0.001}}}$ & $\mathbf{0.046_{{\pm 0.000}}}$ & $\mathbf{0.047_{{\pm 0.002}}}$ & $0.046_{{\pm 0.000}}$ & $\mathbf{0.048_{{\pm 0.001}}}$ & $0.619_{{\pm 0.007}}$ & $0.649_{{\pm 0.002}}$ & $0.642_{{\pm 0.009}}$ & $0.668_{{\pm 0.005}}$ & $0.660_{{\pm 0.002}}$ & $0.667_{{\pm 0.004}}$ & $0.627_{{\pm 0.008}}$ \\
 & $\text{ATS}$ & $0.056_{{\pm 0.001}}$ & $\mathbf{0.047_{{\pm 0.000}}}$ & $0.055_{{\pm 0.001}}$ & $0.053_{{\pm 0.001}}$ & $0.053_{{\pm 0.001}}$ & $0.053_{{\pm 0.001}}$ & $0.056_{{\pm 0.001}}$ & $0.523_{{\pm 0.006}}$ & $0.649_{{\pm 0.002}}$ & $0.544_{{\pm 0.005}}$ & $0.593_{{\pm 0.008}}$ & $0.596_{{\pm 0.009}}$ & $0.547_{{\pm 0.005}}$ & $0.527_{{\pm 0.008}}$ \\
\bottomrule
\end{tabular}}%
\label{tab:main-qwen-results}
\end{table}

\begin{table}[h!]
\footnotesize
\centering
\setlength{\tabcolsep}{0.5pt}
\caption{\textbf{Uncertainty Metrics for SC Measures Across Methods for Mistral.} Mean and standard error of $\widehat{\text{ACE}}$ ($\downarrow$) and AUROC ($\uparrow$) scores for SC measures across  measures baseline and recalibration methods for Mistral.}
\resizebox{\textwidth}{!}{%
\begin{tabular}{lcccccccc|ccccccc}
\toprule
 &  & \multicolumn{7}{c}{$\widehat{\textbf{ACE}}$} & \multicolumn{7}{c}{\textbf{AUROC}} \\
 &  & E-SC & L-SC & ML-SC & B-SC & IC-SC & T-SC & G-SC & E-SC & L-SC & ML-SC & B-SC & IC-SC & T-SC & G-SC \\
\midrule
\multirow{5}{*}{\rotatebox{90}{\textbf{TriviaQA}}} & $\text{Base}$ & $0.175_{{\pm 0.001}}$ & $0.177_{{\pm 0.001}}$ & $0.174_{{\pm 0.001}}$ & $0.108_{{\pm 0.001}}$ & $0.200_{{\pm 0.001}}$ & $0.108_{{\pm 0.001}}$ & $0.159_{{\pm 0.001}}$ & $0.792_{{\pm 0.002}}$ & $0.787_{{\pm 0.001}}$ & $0.792_{{\pm 0.002}}$ & $0.791_{{\pm 0.001}}$ & $0.785_{{\pm 0.002}}$ & $0.791_{{\pm 0.001}}$ & $0.793_{{\pm 0.002}}$ \\
 & $\text{SE}$ & $\mathbf{0.062_{{\pm 0.001}}}$ & $0.051_{{\pm 0.006}}$ & $0.059_{{\pm 0.004}}$ & $0.100_{{\pm 0.007}}$ & $0.080_{{\pm 0.002}}$ & $0.099_{{\pm 0.006}}$ & $0.056_{{\pm 0.003}}$ & $0.842_{{\pm 0.002}}$ & $0.849_{{\pm 0.004}}$ & $0.843_{{\pm 0.003}}$ & $0.842_{{\pm 0.002}}$ & $0.842_{{\pm 0.001}}$ & $0.842_{{\pm 0.003}}$ & $0.843_{{\pm 0.002}}$ \\
 & $\text{TS}$ & $0.064_{{\pm 0.002}}$ & $0.061_{{\pm 0.007}}$ & $0.058_{{\pm 0.001}}$ & $0.100_{{\pm 0.004}}$ & $\mathbf{0.053_{{\pm 0.004}}}$ & $0.100_{{\pm 0.004}}$ & $\mathbf{0.055_{{\pm 0.002}}}$ & $\mathbf{0.843_{{\pm 0.003}}}$ & $\mathbf{0.865_{{\pm 0.003}}}$ & $\mathbf{0.844_{{\pm 0.004}}}$ & $\mathbf{0.843_{{\pm 0.004}}}$ & $\mathbf{0.852_{{\pm 0.004}}}$ & $0.843_{{\pm 0.005}}$ & $\mathbf{0.845_{{\pm 0.006}}}$ \\
 & $\text{Platt}$ & $0.109_{{\pm 0.002}}$ & $0.044_{{\pm 0.003}}$ & $\mathbf{0.057_{{\pm 0.004}}}$ & $0.085_{{\pm 0.005}}$ & $0.082_{{\pm 0.004}}$ & $0.075_{{\pm 0.005}}$ & $0.088_{{\pm 0.003}}$ & $0.834_{{\pm 0.001}}$ & $0.847_{{\pm 0.004}}$ & $0.839_{{\pm 0.001}}$ & $0.842_{{\pm 0.002}}$ & $0.842_{{\pm 0.002}}$ & $\mathbf{0.845_{{\pm 0.002}}}$ & $0.835_{{\pm 0.001}}$ \\
 & $\text{ATS}$ & $0.152_{{\pm 0.001}}$ & $\mathbf{0.044_{{\pm 0.002}}}$ & $0.074_{{\pm 0.003}}$ & $\mathbf{0.065_{{\pm 0.004}}}$ & $0.108_{{\pm 0.003}}$ & $\mathbf{0.062_{{\pm 0.002}}}$ & $0.126_{{\pm 0.001}}$ & $0.832_{{\pm 0.001}}$ & $0.846_{{\pm 0.003}}$ & $0.840_{{\pm 0.002}}$ & $0.837_{{\pm 0.004}}$ & $0.841_{{\pm 0.002}}$ & $0.840_{{\pm 0.002}}$ & $0.834_{{\pm 0.001}}$ \\
\midrule
\multirow{5}{*}{\rotatebox{90}{\textbf{NQ}}} & $\text{Base}$ & $0.300_{{\pm 0.003}}$ & $0.313_{{\pm 0.002}}$ & $0.297_{{\pm 0.002}}$ & $0.197_{{\pm 0.001}}$ & $0.348_{{\pm 0.002}}$ & $0.198_{{\pm 0.001}}$ & $0.274_{{\pm 0.003}}$ & $0.722_{{\pm 0.001}}$ & $0.715_{{\pm 0.003}}$ & $0.720_{{\pm 0.001}}$ & $0.722_{{\pm 0.003}}$ & $0.705_{{\pm 0.002}}$ & $0.722_{{\pm 0.002}}$ & $0.724_{{\pm 0.001}}$ \\
 & $\text{SE}$ & $0.087_{{\pm 0.005}}$ & $0.101_{{\pm 0.003}}$ & $0.072_{{\pm 0.003}}$ & $0.055_{{\pm 0.001}}$ & $0.166_{{\pm 0.003}}$ & $0.055_{{\pm 0.000}}$ & $0.065_{{\pm 0.005}}$ & $0.748_{{\pm 0.004}}$ & $0.780_{{\pm 0.004}}$ & $0.756_{{\pm 0.004}}$ & $\mathbf{0.754_{{\pm 0.003}}}$ & $0.755_{{\pm 0.006}}$ & $0.754_{{\pm 0.003}}$ & $0.752_{{\pm 0.004}}$ \\
 & $\text{TS}$ & $\mathbf{0.055_{{\pm 0.002}}}$ & $\mathbf{0.053_{{\pm 0.004}}}$ & $\mathbf{0.045_{{\pm 0.004}}}$ & $\mathbf{0.042_{{\pm 0.007}}}$ & $\mathbf{0.102_{{\pm 0.001}}}$ & $\mathbf{0.045_{{\pm 0.006}}}$ & $\mathbf{0.043_{{\pm 0.002}}}$ & $0.728_{{\pm 0.004}}$ & $\mathbf{0.798_{{\pm 0.003}}}$ & $0.76_{{\pm 0.01}}$ & $0.732_{{\pm 0.009}}$ & $0.745_{{\pm 0.006}}$ & $0.73_{{\pm 0.01}}$ & $0.732_{{\pm 0.004}}$ \\
 & $\text{Platt}$ & $0.112_{{\pm 0.003}}$ & $0.100_{{\pm 0.002}}$ & $0.080_{{\pm 0.003}}$ & $0.061_{{\pm 0.005}}$ & $0.169_{{\pm 0.002}}$ & $0.069_{{\pm 0.005}}$ & $0.086_{{\pm 0.002}}$ & $0.749_{{\pm 0.004}}$ & $0.779_{{\pm 0.004}}$ & $0.766_{{\pm 0.003}}$ & $0.751_{{\pm 0.001}}$ & $\mathbf{0.756_{{\pm 0.004}}}$ & $\mathbf{0.764_{{\pm 0.003}}}$ & $0.752_{{\pm 0.005}}$ \\
 & $\text{ATS}$ & $0.139_{{\pm 0.006}}$ & $0.100_{{\pm 0.004}}$ & $0.081_{{\pm 0.004}}$ & $0.065_{{\pm 0.004}}$ & $0.177_{{\pm 0.003}}$ & $0.066_{{\pm 0.002}}$ & $0.111_{{\pm 0.007}}$ & $\mathbf{0.752_{{\pm 0.005}}}$ & $0.779_{{\pm 0.004}}$ & $\mathbf{0.767_{{\pm 0.002}}}$ & $0.752_{{\pm 0.004}}$ & $0.755_{{\pm 0.003}}$ & $0.751_{{\pm 0.004}}$ & $\mathbf{0.754_{{\pm 0.006}}}$ \\
\midrule
\multirow{5}{*}{\rotatebox{90}{\textbf{SQuAD}}} & $\text{Base}$ & $\mathbf{0.060_{{\pm 0.002}}}$ & $\mathbf{0.059_{{\pm 0.001}}}$ & $\mathbf{0.053_{{\pm 0.000}}}$ & $\mathbf{0.055_{{\pm 0.001}}}$ & $\mathbf{0.052_{{\pm 0.001}}}$ & $\mathbf{0.054_{{\pm 0.001}}}$ & $\mathbf{0.057_{{\pm 0.002}}}$ & $0.676_{{\pm 0.010}}$ & $0.697_{{\pm 0.007}}$ & $0.716_{{\pm 0.008}}$ & $0.733_{{\pm 0.003}}$ & $0.725_{{\pm 0.002}}$ & $0.735_{{\pm 0.002}}$ & $0.690_{{\pm 0.008}}$ \\
 & $\text{SE}$ & $0.081_{{\pm 0.003}}$ & $0.099_{{\pm 0.002}}$ & $0.069_{{\pm 0.001}}$ & $0.080_{{\pm 0.003}}$ & $0.066_{{\pm 0.000}}$ & $0.079_{{\pm 0.004}}$ & $0.077_{{\pm 0.002}}$ & $\mathbf{0.789_{{\pm 0.004}}}$ & $0.77_{{\pm 0.01}}$ & $\mathbf{0.821_{{\pm 0.005}}}$ & $\mathbf{0.806_{{\pm 0.010}}}$ & $0.799_{{\pm 0.003}}$ & $\mathbf{0.81_{{\pm 0.01}}}$ & $\mathbf{0.801_{{\pm 0.003}}}$ \\
 & $\text{TS}$ & $0.104_{{\pm 0.001}}$ & $0.110_{{\pm 0.003}}$ & $0.086_{{\pm 0.001}}$ & $0.096_{{\pm 0.004}}$ & $0.073_{{\pm 0.003}}$ & $0.095_{{\pm 0.002}}$ & $0.100_{{\pm 0.001}}$ & $0.777_{{\pm 0.003}}$ & $\mathbf{0.78_{{\pm 0.01}}}$ & $0.815_{{\pm 0.002}}$ & $0.800_{{\pm 0.009}}$ & $\mathbf{0.80_{{\pm 0.01}}}$ & $0.798_{{\pm 0.007}}$ & $0.791_{{\pm 0.004}}$ \\
 & $\text{Platt}$ & $0.090_{{\pm 0.002}}$ & $0.100_{{\pm 0.002}}$ & $0.080_{{\pm 0.001}}$ & $0.075_{{\pm 0.004}}$ & $0.064_{{\pm 0.001}}$ & $0.077_{{\pm 0.003}}$ & $0.086_{{\pm 0.004}}$ & $0.766_{{\pm 0.005}}$ & $0.77_{{\pm 0.01}}$ & $0.815_{{\pm 0.004}}$ & $0.804_{{\pm 0.008}}$ & $0.801_{{\pm 0.004}}$ & $0.80_{{\pm 0.01}}$ & $0.782_{{\pm 0.002}}$ \\
 & $\text{ATS}$ & $0.112_{{\pm 0.002}}$ & $0.099_{{\pm 0.002}}$ & $0.102_{{\pm 0.001}}$ & $0.079_{{\pm 0.002}}$ & $0.072_{{\pm 0.003}}$ & $0.082_{{\pm 0.001}}$ & $0.107_{{\pm 0.002}}$ & $0.68_{{\pm 0.01}}$ & $0.77_{{\pm 0.01}}$ & $0.730_{{\pm 0.008}}$ & $0.74_{{\pm 0.01}}$ & $0.73_{{\pm 0.02}}$ & $0.74_{{\pm 0.01}}$ & $0.700_{{\pm 0.008}}$ \\
\bottomrule
\end{tabular}}%
\label{tab:main-mistral-results}
\end{table}

\subsection{Full Results for $\text{SE}_{\text{conf}}$ and $\text{SE}_{\text{vanilla}}$}
\label{sec:full-se-results}
In \Cref{sec:entropy-comparison}, we compared two formulations of semantic entropy: 
(i) $\text{SE}_{\text{conf}}$, which defines correctness relative to the most confident semantic class, providing a principled alignment between entropy computation and correctness evaluation; and 
(ii) $\text{SE}_{\text{vanilla}}$, which follows the original implementation using a separate greedy-decoding procedure to determine correctness \citep{kuhn2023semantic}. 

Importantly, $\text{SE}_{\text{conf}}$ uses the same distribution that generates the final response to estimate both confidence and uncertainty, providing a principled and self-consistent alignment between prediction and uncertainty. In contrast, $\text{SE}{\text{vanilla}}$ relies on separate distributions for correctness and uncertainty, effectively mixing distinct model beliefs in an ad hoc manner. We observed in \Cref{sec:entropy-comparison} that temperature scaling improves discriminability for both formulations, with higher overall AUROCs achieved under $\text{SE}_{\text{conf}}$.

Here, we present the full results for the Qwen model across TriviaQA, NQ, and SQuAD. Complete tables are provided in \Cref{tab:full-se-ours-comparison} and \Cref{tab:full-vanilla-se-comparison} for $\text{SE}_{\text{conf}}$ and $\text{SE}_{\text{vanilla}}$ respectively.
\renewcommand{\arraystretch}{0.7}
\begin{table*}[t!]
\footnotesize
\centering
\setlength{\tabcolsep}{3pt}
\caption{\textbf{AUROC for $\text{SE}_{\text{conf}}$ Across Baseline and Calibration Methods for Qwen.} 
Mean and standard error of AUROC ($\uparrow$) scores computed using $\text{SE}_{\text{conf}}$, where correctness is defined relative to the most confident semantic class. Results are reported for TriviaQA, NQ, and SQuAD across baseline and calibration methods. 
Bold values indicate the best result within each SC measure for a dataset, while \underline{underlined} bold values denote the overall best result for that dataset.}
\resizebox{\textwidth}{!}{%
\begin{tabular}{ccccccccc}
\toprule
 &  & E-SC & L-SC & ML-SC & B-SC & IC-SC & T-SC & G-SC \\
\midrule
\multirow{5}{*}{\rotatebox{90}{\textbf{TriviaQA}}} & $\text{Base}$ & $0.766_{{\pm 0.002}}$ & $0.761_{{\pm 0.002}}$ & $0.764_{{\pm 0.002}}$ & $0.765_{{\pm 0.002}}$ & $0.765_{{\pm 0.002}}$ & $0.766_{{\pm 0.002}}$ & $0.767_{{\pm 0.002}}$ \\
 & $\text{SE}$ & $0.834_{{\pm 0.002}}$ & $0.833_{{\pm 0.001}}$ & $0.832_{{\pm 0.002}}$ & $0.835_{{\pm 0.002}}$ & $0.830_{{\pm 0.002}}$ & $0.836_{{\pm 0.002}}$ & $0.834_{{\pm 0.002}}$ \\
 &  $\text{TS}$ & $\mathbf{0.853_{{\pm 0.002}}}$ & $\mathbf{\underline{0.865}_{{\pm 0.002}}}$ & $\mathbf{0.849_{{\pm 0.002}}}$ & $\mathbf{0.854_{{\pm 0.000}}}$ & $\mathbf{\underline{0.864}_{{\pm 0.004}}}$ & $\mathbf{0.853_{{\pm 0.001}}}$ & $\mathbf{0.851_{{\pm 0.002}}}$ \\
 & $\text{Platt}$ & $0.839_{{\pm 0.002}}$ & $0.840_{{\pm 0.001}}$ & $0.836_{{\pm 0.001}}$ & $0.837_{{\pm 0.001}}$ & $0.836_{{\pm 0.001}}$ & $0.838_{{\pm 0.001}}$ & $0.839_{{\pm 0.002}}$ \\
 & $\text{ATS}$ & $0.843_{{\pm 0.001}}$ & $0.840_{{\pm 0.001}}$ & $0.840_{{\pm 0.002}}$ & $0.843_{{\pm 0.001}}$ & $0.838_{{\pm 0.002}}$ & $0.843_{{\pm 0.001}}$ & $0.843_{{\pm 0.001}}$ \\
\midrule
\multirow{5}{*}{\rotatebox{90}{\textbf{NQ \; }}} & $\text{Base}$ & $0.693_{{\pm 0.003}}$ & $0.691_{{\pm 0.002}}$ & $0.695_{{\pm 0.003}}$ & $0.694_{{\pm 0.002}}$ & $0.692_{{\pm 0.003}}$ & $0.694_{{\pm 0.002}}$ & $0.693_{{\pm 0.003}}$ \\
 & $\text{SE}$ & $0.749_{{\pm 0.003}}$ & $0.752_{{\pm 0.002}}$ & $0.749_{{\pm 0.003}}$ & $0.751_{{\pm 0.002}}$ & $0.746_{{\pm 0.003}}$ & $0.752_{{\pm 0.002}}$ & $0.750_{{\pm 0.003}}$ \\
 & $\text{TS}$ & $\mathbf{0.769_{{\pm 0.004}}}$ & $\mathbf{\underline{0.795}_{{\pm 0.007}}}$ & $\mathbf{0.754_{{\pm 0.002}}}$ & $\mathbf{0.783_{{\pm 0.003}}}$ & $\mathbf{\underline{0.789}_{{\pm 0.003}}}$ & $\mathbf{\underline{0.784}_{{\pm 0.004}}}$ & $0.747_{{\pm 0.002}}$ \\
 & $\text{Platt}$ & $0.746_{{\pm 0.003}}$ & $0.755_{{\pm 0.005}}$ & $0.743_{{\pm 0.001}}$ & $0.749_{{\pm 0.001}}$ & $0.752_{{\pm 0.002}}$ & $0.749_{{\pm 0.001}}$ & $0.746_{{\pm 0.003}}$ \\
 & $\text{ATS}$ & $0.759_{{\pm 0.003}}$ & $0.754_{{\pm 0.005}}$ & $0.754_{{\pm 0.001}}$ & $0.755_{{\pm 0.001}}$ & $0.743_{{\pm 0.001}}$ & $0.756_{{\pm 0.001}}$ & $\mathbf{0.759_{{\pm 0.002}}}$ \\
\midrule
\multirow{5}{*}{\rotatebox{90}{\textbf{SQuAD }}} & $\text{Base}$ & $0.569_{{\pm 0.007}}$ & $0.605_{{\pm 0.004}}$ & $0.574_{{\pm 0.006}}$ & $0.597_{{\pm 0.002}}$ & $0.598_{{\pm 0.003}}$ & $0.597_{{\pm 0.002}}$ & $0.571_{{\pm 0.008}}$ \\
 & $\text{SE}$ & $0.666_{{\pm 0.007}}$ & $0.655_{{\pm 0.006}}$ & $0.666_{{\pm 0.006}}$ & $0.670_{{\pm 0.005}}$ & $0.669_{{\pm 0.005}}$ & $0.668_{{\pm 0.004}}$ & $0.665_{{\pm 0.007}}$ \\
 & $\text{TS}$ & $\mathbf{\underline{0.780}_{{\pm 0.003}}}$ & $\mathbf{0.704_{{\pm 0.004}}}$ & $\mathbf{\underline{0.774}_{{\pm 0.004}}}$ & $\mathbf{\underline{0.752}_{{\pm 0.005}}}$ & $\mathbf{0.772_{{\pm 0.002}}}$ & $\mathbf{\underline{0.732}_{{\pm 0.005}}}$ & $\mathbf{\underline{0.780}_{{\pm 0.003}}}$ \\
 & $\text{Platt}$ & $0.665_{{\pm 0.007}}$ & $0.649_{{\pm 0.002}}$ & $0.665_{{\pm 0.006}}$ & $0.669_{{\pm 0.005}}$ & $0.660_{{\pm 0.002}}$ & $0.668_{{\pm 0.004}}$ & $0.666_{{\pm 0.007}}$ \\
 & $\text{ATS}$ & $0.579_{{\pm 0.009}}$ & $0.649_{{\pm 0.002}}$ & $0.600_{{\pm 0.005}}$ & $0.649_{{\pm 0.005}}$ & $0.649_{{\pm 0.006}}$ & $0.615_{{\pm 0.003}}$ & $0.591_{{\pm 0.009}}$ \\
\bottomrule
\end{tabular}}%
\label{tab:full-se-ours-comparison}
\end{table*}
\renewcommand{\arraystretch}{1.0}

\begin{table}[ht!]
\footnotesize
\centering
\setlength{\tabcolsep}{3pt}
\caption{\textbf{AUROC for $\text{SE}_{\text{vanilla}}$ Across Baseline and Calibration Methods for Qwen.} 
Mean and standard error of AUROC ($\uparrow$) scores computed using $\text{SE}_{\text{vanilla}}$, where correctness is defined via an external decoding procedure \citep{kuhn2023semantic}. Results are reported for TriviaQA, NQ, and SQuAD across baseline and calibration methods. 
Bold values indicate the best result within each SC measure for a dataset, while \underline{underlined} bold values denote the overall best result for that dataset.}
\resizebox{\textwidth}{!}{%
\begin{tabular}{llccccccc}
\toprule
 &  & E-SC & L-SC & ML-SC & B-SC & IC-SC & T-SC & G-SC \\
\midrule
\multirow{5}{*}{\rotatebox{90}{\textbf{TriviaQA}}} & $\text{Base}$ & $0.755_{{\pm 0.002}}$ & $0.755_{{\pm 0.002}}$ & $0.755_{{\pm 0.002}}$ & $0.756_{{\pm 0.002}}$ & $0.754_{{\pm 0.002}}$ & $0.756_{{\pm 0.002}}$ & $0.755_{{\pm 0.002}}$ \\
 & $\text{SE}$ & $0.833_{{\pm 0.001}}$ & $0.833_{{\pm 0.001}}$ & $0.833_{{\pm 0.001}}$ & $0.833_{{\pm 0.001}}$ & $0.833_{{\pm 0.001}}$ & $0.833_{{\pm 0.001}}$ & $0.833_{{\pm 0.001}}$ \\
 & $\text{TS}$ & $\mathbf{0.849_{{\pm 0.001}}}$ & $\mathbf{0.844_{{\pm 0.001}}}$ & $\mathbf{0.849_{{\pm 0.002}}}$ & $\mathbf{0.848_{{\pm 0.002}}}$ & $\mathbf{\underline{0.857_{{\pm 0.002}}}}$ & $\mathbf{0.848_{{\pm 0.002}}}$ & $\mathbf{0.848_{{\pm 0.002}}}$ \\
 & $\text{Platt}$ & $0.832_{{\pm 0.002}}$ & $0.832_{{\pm 0.002}}$ & $0.832_{{\pm 0.002}}$ & $0.833_{{\pm 0.002}}$ & $0.832_{{\pm 0.002}}$ & $0.833_{{\pm 0.002}}$ & $0.832_{{\pm 0.002}}$ \\
 & $\text{ATS}$ & $0.834_{{\pm 0.001}}$ & $0.832_{{\pm 0.002}}$ & $0.835_{{\pm 0.001}}$ & $0.836_{{\pm 0.001}}$ & $0.832_{{\pm 0.002}}$ & $0.836_{{\pm 0.001}}$ & $0.834_{{\pm 0.001}}$ \\
\midrule
\multirow{5}{*}{\rotatebox{90}{\textbf{NQ}}} & $\text{Base}$ & $0.690_{{\pm 0.002}}$ & $0.689_{{\pm 0.002}}$ & $0.691_{{\pm 0.002}}$ & $0.692_{{\pm 0.002}}$ & $0.689_{{\pm 0.002}}$ & $0.692_{{\pm 0.002}}$ & $0.691_{{\pm 0.002}}$ \\
 & $\text{SE}$ & $0.749_{{\pm 0.001}}$ & $\mathbf{0.745_{{\pm 0.001}}}$ & $\mathbf{0.749_{{\pm 0.001}}}$ & $0.749_{{\pm 0.001}}$ & $0.745_{{\pm 0.001}}$ & $0.749_{{\pm 0.001}}$ & $\mathbf{0.750_{{\pm 0.001}}}$ \\
 & $\text{TS}$ & $\mathbf{\underline{0.758_{{\pm 0.001}}}}$ & $0.737_{{\pm 0.002}}$ & $0.743_{{\pm 0.002}}$ & $0.747_{{\pm 0.002}}$ & $\mathbf{0.751_{{\pm 0.004}}}$ & $0.747_{{\pm 0.002}}$ & $0.745_{{\pm 0.001}}$ \\
 & $\text{Platt}$ & $0.745_{{\pm 0.001}}$ & $0.743_{{\pm 0.003}}$ & $0.743_{{\pm 0.002}}$ & $0.747_{{\pm 0.003}}$ & $0.744_{{\pm 0.003}}$ & $0.747_{{\pm 0.003}}$ & $0.744_{{\pm 0.002}}$ \\
 & $\text{ATS}$ & $0.740_{{\pm 0.002}}$ & $0.743_{{\pm 0.003}}$ & $0.742_{{\pm 0.002}}$ & $\mathbf{0.750_{{\pm 0.003}}}$ & $0.741_{{\pm 0.002}}$ & $\mathbf{0.750_{{\pm 0.002}}}$ & $0.740_{{\pm 0.002}}$ \\
\midrule
\multirow{5}{*}{\rotatebox{90}{\textbf{SQuAD}}} & $\text{Base}$ & $0.590_{{\pm 0.001}}$ & $0.595_{{\pm 0.001}}$ & $0.595_{{\pm 0.001}}$ & $0.595_{{\pm 0.002}}$ & $0.595_{{\pm 0.002}}$ & $0.595_{{\pm 0.002}}$ & $0.595_{{\pm 0.001}}$ \\
 & $\text{SE}$ & $0.653_{{\pm 0.002}}$ & $0.653_{{\pm 0.001}}$ & $0.653_{{\pm 0.002}}$ & $0.653_{{\pm 0.002}}$ & $0.653_{{\pm 0.002}}$ & $0.653_{{\pm 0.001}}$ & $0.653_{{\pm 0.002}}$ \\
 & $\text{TS}$ & $\mathbf{0.707_{{\pm 0.007}}}$ & $\mathbf{0.717_{{\pm 0.005}}}$ & $\mathbf{0.709_{{\pm 0.007}}}$ & $\mathbf{0.710_{{\pm 0.007}}}$ & $\mathbf{\underline{0.748_{{\pm 0.003}}}}$ & $\mathbf{0.700_{{\pm 0.001}}}$ & $\mathbf{0.707_{{\pm 0.007}}}$ \\
 & $\text{Platt}$ & $0.653_{{\pm 0.002}}$ & $0.652_{{\pm 0.004}}$ & $0.654_{{\pm 0.002}}$ & $0.653_{{\pm 0.001}}$ & $0.649_{{\pm 0.004}}$ & $0.653_{{\pm 0.001}}$ & $0.653_{{\pm 0.002}}$ \\
 & $\text{ATS}$ & $0.575_{{\pm 0.007}}$ & $0.652_{{\pm 0.004}}$ & $0.583_{{\pm 0.008}}$ & $0.621_{{\pm 0.003}}$ & $0.624_{{\pm 0.002}}$ & $0.616_{{\pm 0.004}}$ & $0.578_{{\pm 0.009}}$ \\
\bottomrule
\end{tabular}}%
\label{tab:full-vanilla-se-comparison}
\end{table}

\subsection{Temperatures from Optimised Temperature Scaling}
\Cref{tab:temperature_settings} gives the best final temperature settings after sweeping over hyperparameters based on the Brier score on the validation set, and these are used as the final settings for all results presented in this paper.
\begin{table}[ht!]
\footnotesize
\centering
\caption{\textbf{Optimised Temperature Settings.} Best temperature settings for the temperature scaling (TS) method across models and datasets. Temperatures represent the final optimised after calibration training and hyperparameter selection and are those that are used to give the results shown in \Cref{fig:main results}.}
\resizebox{0.7\textwidth}{!}{%
\begin{tabular}{llccccccc}
\toprule
Dataset & Model & E-SC & L-SC & ML-SC & B-SC & IC-SC & T-SC & G-SC \\
\midrule
\multirow{3}{*}{TriviaQA} & Llama & 1.26 & 1.63 & 1.26 & 1.21 & 1.21 & 1.21 & 1.21 \\
 & Qwen & 1.80 & 2.07 & 1.32 & 1.56 & 1.32 & 1.56 & 1.56 \\
 & Mistral & 1.18 & 1.18 & 1.00 & 1.00 & 1.00 & 1.00 & 1.00 \\
\midrule
\multirow{3}{*}{NQ} & Llama & 2.00 & 2.00 & 1.47 & 1.47 & 1.47 & 1.71 & 1.71 \\
 & Qwen & 2.41 & 2.41 & 2.41 & 2.03 & 2.41 & 2.41 & 2.41 \\
 & Mistral & 1.70 & 1.70 & 1.28 & 1.28 & 1.28 & 1.28 & 1.28 \\
\midrule
\multirow{3}{*}{SQuAD} & Llama & 1.38 & 1.38 & 1.38 & 1.40 & 1.38 & 1.38 & 1.38 \\
 & Qwen & 1.75 & 2.20 & 1.59 & 1.69 & 1.69 & 1.59 & 1.59 \\
 & Mistral & 1.09 & 1.09 & 1.09 & 1.09 & 1.09 & 1.27 & 1.27 \\
\bottomrule
\end{tabular}}%
\label{tab:temperature_settings}
\end{table}

\subsection{Plots of Model Selective Accuracy} 
\begin{figure}[ht!]
    \centering
    \includegraphics[width=1.0\linewidth]{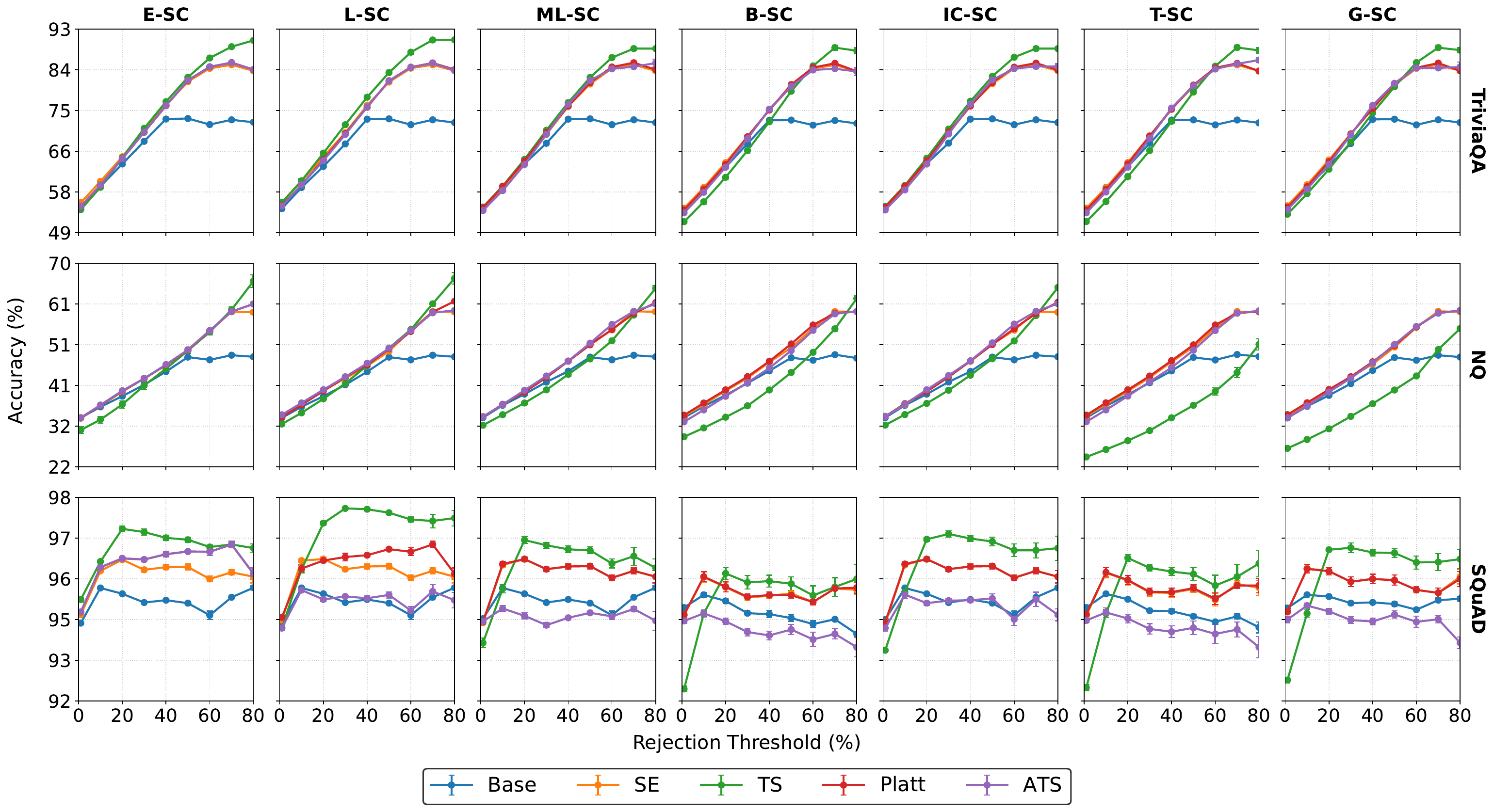}
    \caption{\textbf{Selective Accuracy for Qwen Across Varying Rejection Rates.} Mean selective accuracy ($\uparrow$) curves of the Qwen model across datasets, comparing SC measures across baseline and recalibration. Error bars show one standard error across runs.}
    \label{fig:sel-acc-full-qwen}
\end{figure}

We compare the selective accuracy of recalibration methods and baselines across the different semantic confidence measures introduced in this paper for the Qwen model. The main results are shown in \Cref{fig:select-acc-qwen} for four of the SC measures discussed within this work. For completeness, we include the full results for Qwen across all SC measures of confidence in \Cref{fig:sel-acc-full-qwen}.

\subsection{Model Brier Scores}
As discussed in \Cref{sec:metrics}, the Brier score decomposes into three components, uncertainty, resolution, and reliability, which relate to calibration and discrimination but differ from the literature standard $\widehat{\text{ACE}}$ and AUROC estimators we have used for our main results presentation. To complement the main results on semantic uncertainty quantification shown in \Cref{fig:main results}, we report Brier scores for the Llama model across baseline recalibration methods, and SC measures in \Cref{tab:Llama-brier}.

For TriviaQA and NQ, temperature scaling (TS) consistently improves Brier scores, highlighting its effectiveness on these challenging closed-book tasks where base models are poorly calibrated and discriminative. In contrast, on SQuAD, TS often yields slightly worse Brier scores than the base and baseline models. This is because SQuAD base models are already well-calibrated, resulting in lower Brier scores overall and smaller absolute differences between methods.

These results reinforce our main conclusion: TS is most beneficial for datasets with poor base calibration, such as TriviaQA and NQ. On easier datasets like SQuAD, Brier scores show only minor absolute differences, yet AUROC results in \Cref{fig:main results} reveal that TS can still enhance discriminability even when base calibration is strong.

\begin{table}[h!]
\footnotesize
\centering
\setlength{\tabcolsep}{3pt}
\caption{\textbf{Brier Scores for SC Measures Across Baseline and Recalibration Methods for Llama.} 
Mean and standard error of Brier scores ($\downarrow$) are reported for each SC measure, comparing baseline and recalibration methods for the Llama model.}
\label{tab:Llama-brier}
\resizebox{\textwidth}{!}{%
\begin{tabular}{llccccccc}
\toprule
 &  & E-SC & L-SC & ML-SC & B-SC & IC-SC & T-SC & G-SC \\
\midrule
\multirow{5}{*}{\rotatebox{90}{\textbf{TriviaQA}}} & $\text{Base}$ & $0.197_{{\pm 0.001}}$ & $0.189_{{\pm 0.001}}$ & $0.194_{{\pm 0.001}}$ & $0.174_{{\pm 0.001}}$ & $0.200_{{\pm 0.001}}$ & $0.174_{{\pm 0.001}}$ & $0.190_{{\pm 0.001}}$ \\
 & $\text{SE}$ & $0.147_{{\pm 0.001}}$ & $0.140_{{\pm 0.001}}$ & $0.142_{{\pm 0.001}}$ & $0.147_{{\pm 0.001}}$ & $0.147_{{\pm 0.001}}$ & $0.147_{{\pm 0.001}}$ & $0.145_{{\pm 0.001}}$ \\
 & $\text{TS}$ & $\mathbf{0.140_{{\pm 0.002}}}$ & $\mathbf{0.135_{{\pm 0.002}}}$ & $\mathbf{0.136_{{\pm 0.003}}}$ & $0.150_{{\pm 0.001}}$ & $\mathbf{0.130_{{\pm 0.003}}}$ & $0.151_{{\pm 0.002}}$ & $\mathbf{0.142_{{\pm 0.002}}}$ \\
 & $\text{Platt}$ & $0.147_{{\pm 0.001}}$ & $0.140_{{\pm 0.001}}$ & $0.140_{{\pm 0.001}}$ & $0.143_{{\pm 0.001}}$ & $0.145_{{\pm 0.000}}$ & $\mathbf{0.142_{{\pm 0.001}}}$ & $0.144_{{\pm 0.001}}$ \\
 & $\text{ATS}$ & $0.150_{{\pm 0.001}}$ & $0.140_{{\pm 0.001}}$ & $0.142_{{\pm 0.001}}$ & $\mathbf{0.143_{{\pm 0.001}}}$ & $0.146_{{\pm 0.002}}$ & $0.143_{{\pm 0.001}}$ & $0.147_{{\pm 0.001}}$ \\
\midrule
\multirow{5}{*}{\rotatebox{90}{\textbf{NQ}}} & $\text{Base}$ & $0.358_{{\pm 0.001}}$ & $0.350_{{\pm 0.002}}$ & $0.353_{{\pm 0.001}}$ & $0.309_{{\pm 0.000}}$ & $0.374_{{\pm 0.001}}$ & $0.310_{{\pm 0.000}}$ & $0.345_{{\pm 0.001}}$ \\
 & $\text{SE}$ & $0.240_{{\pm 0.002}}$ & $0.236_{{\pm 0.002}}$ & $0.233_{{\pm 0.002}}$ & $0.222_{{\pm 0.002}}$ & $0.264_{{\pm 0.003}}$ & $0.222_{{\pm 0.002}}$ & $0.231_{{\pm 0.002}}$ \\
 & $\text{TS}$ & $\mathbf{0.200_{{\pm 0.001}}}$ & $\mathbf{0.175_{{\pm 0.002}}}$ & $\mathbf{0.196_{{\pm 0.002}}}$ & $\mathbf{0.207_{{\pm 0.002}}}$ & $\mathbf{0.208_{{\pm 0.000}}}$ & $\mathbf{0.206_{{\pm 0.002}}}$ & $\mathbf{0.199_{{\pm 0.002}}}$ \\
 & $\text{Platt}$ & $0.242_{{\pm 0.001}}$ & $0.239_{{\pm 0.002}}$ & $0.235_{{\pm 0.001}}$ & $0.222_{{\pm 0.001}}$ & $0.266_{{\pm 0.001}}$ & $0.221_{{\pm 0.001}}$ & $0.233_{{\pm 0.001}}$ \\
 & $\text{ATS}$ & $0.235_{{\pm 0.001}}$ & $0.238_{{\pm 0.002}}$ & $0.228_{{\pm 0.002}}$ & $0.222_{{\pm 0.002}}$ & $0.261_{{\pm 0.002}}$ & $0.221_{{\pm 0.002}}$ & $0.228_{{\pm 0.001}}$ \\
\midrule
\multirow{5}{*}{\rotatebox{90}{\textbf{SQuAD}}} & $\text{Base}$ & $\mathbf{0.051_{{\pm 0.000}}}$ & $\mathbf{0.047_{{\pm 0.000}}}$ & $\mathbf{0.051_{{\pm 0.000}}}$ & $\mathbf{0.050_{{\pm 0.001}}}$ & $\mathbf{0.048_{{\pm 0.001}}}$ & $\mathbf{0.050_{{\pm 0.001}}}$ & $\mathbf{0.051_{{\pm 0.000}}}$ \\
 & $\text{SE}$ & $0.056_{{\pm 0.001}}$ & $0.055_{{\pm 0.000}}$ & $0.056_{{\pm 0.001}}$ & $0.060_{{\pm 0.000}}$ & $0.052_{{\pm 0.001}}$ & $0.060_{{\pm 0.000}}$ & $0.055_{{\pm 0.001}}$ \\
 & $\text{TS}$ & $0.067_{{\pm 0.003}}$ & $0.065_{{\pm 0.000}}$ & $0.070_{{\pm 0.001}}$ & $0.077_{{\pm 0.001}}$ & $0.058_{{\pm 0.001}}$ & $0.077_{{\pm 0.001}}$ & $0.068_{{\pm 0.002}}$ \\
 & $\text{Platt}$ & $0.056_{{\pm 0.001}}$ & $0.054_{{\pm 0.001}}$ & $0.056_{{\pm 0.001}}$ & $0.061_{{\pm 0.001}}$ & $0.051_{{\pm 0.001}}$ & $0.060_{{\pm 0.000}}$ & $0.055_{{\pm 0.001}}$ \\
 & $\text{ATS}$ & $0.058_{{\pm 0.001}}$ & $0.055_{{\pm 0.000}}$ & $0.056_{{\pm 0.001}}$ & $0.051_{{\pm 0.001}}$ & $0.049_{{\pm 0.001}}$ & $0.051_{{\pm 0.001}}$ & $0.057_{{\pm 0.001}}$ \\
\bottomrule
\end{tabular}}%
\end{table}

\subsection{Validation of NLI-Based Semantic Clustering} \label{sec: clustering quality}
Following prior work, we employ a DeBERTa-v2-XXLarge natural language inference (NLI) model to cluster semantically equivalent responses. To ensure that token-level likelihood adjustments propagate meaningfully to semantic clusters in our setting, we perform both model-based and manual audits of clustering quality using this methodology, corroborating the findings of \citet{kuhn2023semantic}, who show that NLI-based clustering performs well for the short-form generative QA tasks studied both in their work and ours.

\paragraph{Verification Methodology}
We randomly sampled 100 examples from the Natural Questions (NQ) dataset and generated 10 responses per example using Llama-3.1-8B-Instruct. The resulting semantic clusters were evaluated using a two-stage verification process:
\begin{itemize}
    \item \textbf{LLM Judge:} We prompted Claude~4.5 \citep{anthropic_claude45} to assess whether (i) each cluster contained only semantically equivalent responses and (ii) different clusters corresponded to distinct semantic meanings with no overlap.
    \item \textbf{Human Audit:} We manually reviewed the LLM annotations to verify the correctness of the semantic judgements.
\end{itemize}

\paragraph{Results and Error Analysis}
This audit yielded a clustering accuracy of 94\%. The remaining 6\% of errors were primarily false negatives produced by the NLI model, where responses conveying the same lack of information, often through verbose refusals or meta-level statements (e.g., ``I am not sure, but...''), were incorrectly split into separate clusters.

\subsection{Validation of Lexical vs. NLI Correctness Evaluation}
\label{app:correctness_validation}

In our main evaluation, we employ a hybrid strategy to quantify uncertainty: we use a DeBERTa-based NLI model to cluster model generations (where variability is high) but rely on standard lexical metrics to determine the correctness of the final response against ground-truth answers. We adopt this methodology to remain strictly comparable with prior work in semantic uncertainty quantification \citep{kuhn2023semantic, farquhar2024detecting} and to avoid the prohibitive computational cost of performing NLI pairwise comparisons for every correctness check.

\paragraph{Ablation Study: Lexical vs. NLI Correctness.}
To validate the robustness of this approach, we conducted an ablation study comparing our standard lexical evaluation against a purely NLI-based correctness check. Specifically, we used the Llama-3.1-8B-Instruct model on the TriviaQA dataset to generate responses and independently assessed their correctness using both the standard lexical criteria (see \Cref{app:correctness_validation}) and the DeBERTa-based NLI model used for clustering. In the NLI setting, a response is deemed correct if it entails, and is entailed by, a ground-truth answer within the context of the question, following the same bidirectional entailment logic used for clustering responses.

\paragraph{Results.}
We observed a 91\% agreement rate between the standard lexical methodology and the NLI-based evaluation when determining response correctness relative to the ground truth. This high level of consistency corroborates that the literature standard approach aligns well with the NLI approach whilst being computationally much cheaper, highlighting its practical utility and robustness for practitioners. 

\subsection{Model Calibration Results Are Robust to the Choice of Metric} \label{sec: robustness of choice of calibration metric}
As noted in \Cref{sec:metrics}, bin-based metrics such as ECE and ACE can be sensitive to the choice of binning scheme. This has motivated the development of alternative, bin-free calibration metrics, such as MCB-CORP, which is non-parametric and comes with strong statistical guarantees \citep{dimitriadis2021stable}.

To evaluate the robustness of our findings, we compare ACE, ECE and MCB-CORP for the Qwen model in \Cref{tab:ace vs mcb-corp}. Consistent with \Cref{fig:main results}, we observe that TS consistently improves semantic calibration on the more challenging closed-book datasets (TriviaQA and NQ), while its effect on the open-book dataset SQuAD is negligible, reflecting the already strong calibration of base models here.

We further quantify robustness in \Cref{tab:combined_volatility} by reporting the mean absolute rank change (volatility) when replacing ACE with MCB-CORP. Volatility is minimal for TriviaQA and NQ and slightly higher for SQuAD, where methods perform similarly. Overall, these results confirm that our main conclusions are independent of the specific calibration metric chosen.

\begin{table}[h!]
\footnotesize
\centering
\setlength{\tabcolsep}{3pt}
\caption{\textbf{Calibration Metric Comparison of SC Measures Across Methods for Qwen.} Mean and standard error of calibration error for SC measures under $\ace$ (bin-based) ($\downarrow$) and MCB-CORP (bin-free) ($\downarrow$) across baseline and recalibration methods on TriviaQA, NQ, and SQuAD. \textbf{Bold} entries indicate the best-performing method for each SC measure within a dataset, allowing direct comparison between bin-based and bin-free calibration metrics.}
\resizebox{\textwidth}{!}{%
\begin{tabular}{ccccccccc|ccccccc}
\toprule
 &  & \multicolumn{7}{c}{$\widehat{\textbf{ACE}}$} & \multicolumn{7}{c}{\textbf{CORP-MCB}} \\
 &  & E-SC & L-SC & ML-SC & B-SC & IC-SC & T-SC & G-SC & E-SC & L-SC & ML-SC & B-SC & IC-SC & T-SC & G-SC \\
\midrule
\multirow{5}{*}{\rotatebox{90}{\textbf{TriviaQA}}} & $\text{Base}$ & $0.308_{{\pm 0.001}}$ & $0.299_{{\pm 0.001}}$ & $0.304_{{\pm 0.000}}$ & $0.240_{{\pm 0.000}}$ & $0.320_{{\pm 0.001}}$ & $0.240_{{\pm 0.000}}$ & $0.293_{{\pm 0.001}}$ & $0.105_{{\pm 0.001}}$ & $0.096_{{\pm 0.001}}$ & $0.101_{{\pm 0.001}}$ & $0.062_{{\pm 0.000}}$ & $0.114_{{\pm 0.001}}$ & $0.062_{{\pm 0.000}}$ & $0.093_{{\pm 0.000}}$ \\
 & $\text{SE}$ & $0.165_{{\pm 0.002}}$ & $0.159_{{\pm 0.001}}$ & $0.158_{{\pm 0.001}}$ & $0.097_{{\pm 0.002}}$ & $0.194_{{\pm 0.001}}$ & $0.099_{{\pm 0.002}}$ & $0.145_{{\pm 0.002}}$ & $0.035_{{\pm 0.001}}$ & $0.028_{{\pm 0.001}}$ & $0.031_{{\pm 0.000}}$ & $0.015_{{\pm 0.000}}$ & $0.045_{{\pm 0.001}}$ & $0.015_{{\pm 0.000}}$ & $0.027_{{\pm 0.001}}$ \\
 & $\text{TS}$ & $\mathbf{0.080_{{\pm 0.001}}}$ & $\mathbf{0.066_{{\pm 0.005}}}$ & $\mathbf{0.069_{{\pm 0.002}}}$ & $\mathbf{0.082_{{\pm 0.003}}}$ & $\mathbf{0.085_{{\pm 0.004}}}$ & $\mathbf{0.081_{{\pm 0.003}}}$ & $\mathbf{0.067_{{\pm 0.002}}}$ & $\mathbf{0.010_{{\pm 0.000}}}$ & $\mathbf{0.006_{{\pm 0.001}}}$ & $\mathbf{0.009_{{\pm 0.000}}}$ & $\mathbf{0.011_{{\pm 0.001}}}$ & $\mathbf{0.011_{{\pm 0.001}}}$ & $\mathbf{0.010_{{\pm 0.001}}}$ & $\mathbf{0.008_{{\pm 0.000}}}$ \\
 & $\text{Platt}$ & $0.170_{{\pm 0.001}}$ & $0.168_{{\pm 0.002}}$ & $0.163_{{\pm 0.002}}$ & $0.100_{{\pm 0.002}}$ & $0.203_{{\pm 0.001}}$ & $0.101_{{\pm 0.003}}$ & $0.150_{{\pm 0.001}}$ & $0.035_{{\pm 0.000}}$ & $0.032_{{\pm 0.001}}$ & $0.032_{{\pm 0.001}}$ & $0.015_{{\pm 0.000}}$ & $0.050_{{\pm 0.001}}$ & $0.015_{{\pm 0.000}}$ & $0.028_{{\pm 0.000}}$ \\
 & $\text{ATS}$ & $0.204_{{\pm 0.003}}$ & $0.168_{{\pm 0.002}}$ & $0.178_{{\pm 0.002}}$ & $0.132_{{\pm 0.003}}$ & $0.219_{{\pm 0.001}}$ & $0.133_{{\pm 0.002}}$ & $0.183_{{\pm 0.003}}$ & $0.050_{{\pm 0.001}}$ & $0.032_{{\pm 0.001}}$ & $0.037_{{\pm 0.001}}$ & $0.020_{{\pm 0.001}}$ & $0.057_{{\pm 0.001}}$ & $0.021_{{\pm 0.001}}$ & $0.040_{{\pm 0.001}}$ \\
\midrule
\multirow{5}{*}{\rotatebox{90}{\textbf{NQ}}} & $\text{Base}$ & $0.487_{{\pm 0.001}}$ & $0.473_{{\pm 0.001}}$ & $0.480_{{\pm 0.001}}$ & $0.399_{{\pm 0.001}}$ & $0.496_{{\pm 0.001}}$ & $0.399_{{\pm 0.001}}$ & $0.468_{{\pm 0.001}}$ & $0.253_{{\pm 0.001}}$ & $0.237_{{\pm 0.002}}$ & $0.245_{{\pm 0.002}}$ & $0.181_{{\pm 0.001}}$ & $0.258_{{\pm 0.002}}$ & $0.181_{{\pm 0.001}}$ & $0.235_{{\pm 0.001}}$ \\
 & $\text{SE}$ & $0.319_{{\pm 0.002}}$ & $0.318_{{\pm 0.002}}$ & $0.311_{{\pm 0.002}}$ & $0.228_{{\pm 0.002}}$ & $0.354_{{\pm 0.002}}$ & $0.228_{{\pm 0.002}}$ & $0.294_{{\pm 0.002}}$ & $0.121_{{\pm 0.001}}$ & $0.118_{{\pm 0.001}}$ & $0.116_{{\pm 0.001}}$ & $0.072_{{\pm 0.001}}$ & $0.142_{{\pm 0.001}}$ & $0.072_{{\pm 0.001}}$ & $0.106_{{\pm 0.001}}$ \\
 & $\text{TS}$ & $\mathbf{0.155_{{\pm 0.002}}}$ & $\mathbf{0.084_{{\pm 0.002}}}$ & $\mathbf{0.085_{{\pm 0.003}}}$ & $\mathbf{0.085_{{\pm 0.001}}}$ & $\mathbf{0.192_{{\pm 0.002}}}$ & $\mathbf{0.083_{{\pm 0.002}}}$ & $\mathbf{0.155_{{\pm 0.002}}}$ & $\mathbf{0.034_{{\pm 0.001}}}$ & $\mathbf{0.013_{{\pm 0.001}}}$ & $\mathbf{0.018_{{\pm 0.001}}}$ & $\mathbf{0.012_{{\pm 0.000}}}$ & $\mathbf{0.043_{{\pm 0.001}}}$ & $\mathbf{0.011_{{\pm 0.000}}}$ & $\mathbf{0.039_{{\pm 0.003}}}$ \\
 & $\text{Platt}$ & $0.314_{{\pm 0.001}}$ & $0.318_{{\pm 0.001}}$ & $0.307_{{\pm 0.001}}$ & $0.231_{{\pm 0.002}}$ & $0.357_{{\pm 0.000}}$ & $0.232_{{\pm 0.002}}$ & $0.289_{{\pm 0.001}}$ & $0.117_{{\pm 0.001}}$ & $0.119_{{\pm 0.001}}$ & $0.113_{{\pm 0.001}}$ & $0.074_{{\pm 0.001}}$ & $0.143_{{\pm 0.001}}$ & $0.074_{{\pm 0.001}}$ & $0.102_{{\pm 0.001}}$ \\
 & $\text{ATS}$ & $0.315_{{\pm 0.004}}$ & $0.318_{{\pm 0.001}}$ & $0.280_{{\pm 0.002}}$ & $0.236_{{\pm 0.004}}$ & $0.353_{{\pm 0.002}}$ & $0.237_{{\pm 0.003}}$ & $0.295_{{\pm 0.004}}$ & $0.120_{{\pm 0.002}}$ & $0.118_{{\pm 0.000}}$ & $0.097_{{\pm 0.001}}$ & $0.074_{{\pm 0.002}}$ & $0.139_{{\pm 0.001}}$ & $0.074_{{\pm 0.002}}$ & $0.108_{{\pm 0.002}}$ \\
\midrule
\multirow{5}{*}{\rotatebox{90}{\textbf{SQuAD}}} & $\text{Base}$ & $0.051_{{\pm 0.001}}$ & $0.047_{{\pm 0.000}}$ & $0.051_{{\pm 0.001}}$ & $0.048_{{\pm 0.000}}$ & $0.048_{{\pm 0.000}}$ & $0.048_{{\pm 0.000}}$ & $0.051_{{\pm 0.001}}$ & $\mathbf{0.003_{{\pm 0.000}}}$ & $\mathbf{0.004_{{\pm 0.000}}}$ & $\mathbf{0.003_{{\pm 0.000}}}$ & $0.004_{{\pm 0.000}}$ & $0.005_{{\pm 0.000}}$ & $0.004_{{\pm 0.000}}$ & $\mathbf{0.003_{{\pm 0.000}}}$ \\
 & $\text{SE}$ & $\mathbf{0.050_{{\pm 0.001}}}$ & $0.048_{{\pm 0.001}}$ & $0.044_{{\pm 0.001}}$ & $0.046_{{\pm 0.000}}$ & $0.048_{{\pm 0.002}}$ & $\mathbf{0.046_{{\pm 0.000}}}$ & $0.049_{{\pm 0.001}}$ & $0.004_{{\pm 0.000}}$ & $0.006_{{\pm 0.000}}$ & $0.004_{{\pm 0.000}}$ & $0.004_{{\pm 0.000}}$ & $\mathbf{0.005_{{\pm 0.000}}}$ & $0.004_{{\pm 0.000}}$ & $0.004_{{\pm 0.000}}$ \\
 & $\text{TS}$ & $0.058_{{\pm 0.001}}$ & $0.060_{{\pm 0.001}}$ & $0.053_{{\pm 0.002}}$ & $0.051_{{\pm 0.001}}$ & $0.049_{{\pm 0.003}}$ & $0.052_{{\pm 0.002}}$ & $0.055_{{\pm 0.001}}$ & $0.008_{{\pm 0.000}}$ & $0.011_{{\pm 0.000}}$ & $0.007_{{\pm 0.001}}$ & $0.007_{{\pm 0.000}}$ & $0.007_{{\pm 0.000}}$ & $0.007_{{\pm 0.001}}$ & $0.008_{{\pm 0.000}}$ \\
 & $\text{Platt}$ & $0.050_{{\pm 0.001}}$ & $\mathbf{0.047_{{\pm 0.000}}}$ & $\mathbf{0.044_{{\pm 0.001}}}$ & $\mathbf{0.046_{{\pm 0.000}}}$ & $\mathbf{0.047_{{\pm 0.002}}}$ & $0.046_{{\pm 0.000}}$ & $\mathbf{0.048_{{\pm 0.001}}}$ & $0.004_{{\pm 0.000}}$ & $0.005_{{\pm 0.000}}$ & $0.004_{{\pm 0.000}}$ & $\mathbf{0.004_{{\pm 0.000}}}$ & $0.005_{{\pm 0.001}}$ & $0.004_{{\pm 0.000}}$ & $0.004_{{\pm 0.000}}$ \\
 & $\text{ATS}$ & $0.056_{{\pm 0.001}}$ & $\mathbf{0.047_{{\pm 0.000}}}$ & $0.055_{{\pm 0.001}}$ & $0.053_{{\pm 0.001}}$ & $0.053_{{\pm 0.001}}$ & $0.053_{{\pm 0.001}}$ & $0.056_{{\pm 0.001}}$ & $0.004_{{\pm 0.000}}$ & $0.005_{{\pm 0.000}}$ & $0.004_{{\pm 0.000}}$ & $0.005_{{\pm 0.000}}$ & $0.005_{{\pm 0.000}}$ & $\mathbf{0.004_{{\pm 0.000}}}$ & $0.004_{{\pm 0.000}}$ \\
\bottomrule
\end{tabular}}%
\label{tab:ace vs mcb-corp}
\end{table}

\begin{table}[h!]
\footnotesize
\centering
\caption{\textbf{$\ace$ vs $\ece$ Comparison of SC Measures Across Methods for Qwen.} Mean and standard error of calibration error for SC measures under $\ace$ (bin-based) ($\downarrow$) and $\ece$ ($\downarrow$) across baseline and recalibration methods on TriviaQA, NQ, and SQuAD. \textbf{Bold} entries indicate the best-performing method for each SC measure within a dataset.}
\setlength{\tabcolsep}{3pt}
\resizebox{\textwidth}{!}{%
\begin{tabular}{llcccccccccccccc}
\toprule
 & Method & \multicolumn{7}{c}{\textbf{ACE}} & \multicolumn{7}{c}{\textbf{ECE}} \\
 &  & E-SC & L-SC & ML-SC & B-SC & IC-SC & T-SC & G-SC & E-SC & L-SC & ML-SC & B-SC & IC-SC & T-SC & G-SC \\
\midrule
\multirow{5}{*}{\rotatebox{90}{\textbf{TriviaQA}}} & $\text{Base}$ & $0.308_{{\pm 0.001}}$ & $0.299_{{\pm 0.001}}$ & $0.304_{{\pm 0.000}}$ & $0.240_{{\pm 0.000}}$ & $0.320_{{\pm 0.001}}$ & $0.240_{{\pm 0.000}}$ & $0.293_{{\pm 0.001}}$ & $0.308_{{\pm 0.001}}$ & $0.299_{{\pm 0.001}}$ & $0.304_{{\pm 0.000}}$ & $0.240_{{\pm 0.000}}$ & $0.321_{{\pm 0.001}}$ & $0.240_{{\pm 0.000}}$ & $0.293_{{\pm 0.001}}$ \\
 & $\text{SE}$ & $0.165_{{\pm 0.002}}$ & $0.159_{{\pm 0.001}}$ & $0.158_{{\pm 0.001}}$ & $0.097_{{\pm 0.002}}$ & $0.194_{{\pm 0.001}}$ & $0.099_{{\pm 0.002}}$ & $0.145_{{\pm 0.002}}$ & $0.165_{{\pm 0.002}}$ & $0.158_{{\pm 0.001}}$ & $0.158_{{\pm 0.001}}$ & $0.100_{{\pm 0.002}}$ & $0.194_{{\pm 0.001}}$ & $0.101_{{\pm 0.002}}$ & $0.145_{{\pm 0.002}}$ \\
 & $\text{TS}$ & $\mathbf{0.080_{{\pm 0.001}}}$ & $\mathbf{0.066_{{\pm 0.005}}}$ & $\mathbf{0.069_{{\pm 0.002}}}$ & $\mathbf{0.082_{{\pm 0.003}}}$ & $\mathbf{0.085_{{\pm 0.004}}}$ & $\mathbf{0.081_{{\pm 0.003}}}$ & $\mathbf{0.067_{{\pm 0.002}}}$ & $\mathbf{0.080_{{\pm 0.001}}}$ & $\mathbf{0.066_{{\pm 0.004}}}$ & $\mathbf{0.069_{{\pm 0.002}}}$ & $\mathbf{0.086_{{\pm 0.004}}}$ & $\mathbf{0.087_{{\pm 0.004}}}$ & $\mathbf{0.085_{{\pm 0.005}}}$ & $\mathbf{0.069_{{\pm 0.002}}}$ \\
 & $\text{Platt}$ & $0.170_{{\pm 0.001}}$ & $0.168_{{\pm 0.002}}$ & $0.163_{{\pm 0.002}}$ & $0.100_{{\pm 0.002}}$ & $0.203_{{\pm 0.001}}$ & $0.101_{{\pm 0.003}}$ & $0.150_{{\pm 0.001}}$ & $0.170_{{\pm 0.001}}$ & $0.168_{{\pm 0.002}}$ & $0.163_{{\pm 0.002}}$ & $0.101_{{\pm 0.002}}$ & $0.203_{{\pm 0.001}}$ & $0.102_{{\pm 0.002}}$ & $0.150_{{\pm 0.001}}$ \\
 & $\text{ATS}$ & $0.204_{{\pm 0.003}}$ & $0.168_{{\pm 0.002}}$ & $0.178_{{\pm 0.002}}$ & $0.132_{{\pm 0.003}}$ & $0.219_{{\pm 0.001}}$ & $0.133_{{\pm 0.002}}$ & $0.183_{{\pm 0.003}}$ & $0.204_{{\pm 0.003}}$ & $0.168_{{\pm 0.002}}$ & $0.178_{{\pm 0.002}}$ & $0.132_{{\pm 0.003}}$ & $0.219_{{\pm 0.001}}$ & $0.133_{{\pm 0.002}}$ & $0.183_{{\pm 0.003}}$ \\
\midrule
\multirow{5}{*}{\rotatebox{90}{\textbf{NQ}}} & $\text{Base}$ & $0.487_{{\pm 0.001}}$ & $0.473_{{\pm 0.001}}$ & $0.480_{{\pm 0.001}}$ & $0.399_{{\pm 0.001}}$ & $0.496_{{\pm 0.001}}$ & $0.399_{{\pm 0.001}}$ & $0.468_{{\pm 0.001}}$ & $0.486_{{\pm 0.001}}$ & $0.472_{{\pm 0.002}}$ & $0.479_{{\pm 0.001}}$ & $0.399_{{\pm 0.001}}$ & $0.496_{{\pm 0.001}}$ & $0.399_{{\pm 0.001}}$ & $0.468_{{\pm 0.001}}$ \\
 & $\text{SE}$ & $0.319_{{\pm 0.002}}$ & $0.318_{{\pm 0.002}}$ & $0.311_{{\pm 0.002}}$ & $0.228_{{\pm 0.002}}$ & $0.354_{{\pm 0.002}}$ & $0.228_{{\pm 0.002}}$ & $0.294_{{\pm 0.002}}$ & $0.319_{{\pm 0.002}}$ & $0.318_{{\pm 0.002}}$ & $0.311_{{\pm 0.002}}$ & $0.228_{{\pm 0.002}}$ & $0.353_{{\pm 0.002}}$ & $0.228_{{\pm 0.002}}$ & $0.294_{{\pm 0.002}}$ \\
 & $\text{TS}$ & $\mathbf{0.155_{{\pm 0.002}}}$ & $\mathbf{0.084_{{\pm 0.002}}}$ & $\mathbf{0.085_{{\pm 0.003}}}$ & $\mathbf{0.085_{{\pm 0.001}}}$ & $\mathbf{0.192_{{\pm 0.002}}}$ & $\mathbf{0.083_{{\pm 0.002}}}$ & $\mathbf{0.155_{{\pm 0.002}}}$ & $\mathbf{0.155_{{\pm 0.002}}}$ & $\mathbf{0.081_{{\pm 0.002}}}$ & $\mathbf{0.085_{{\pm 0.003}}}$ & $\mathbf{0.082_{{\pm 0.003}}}$ & $\mathbf{0.192_{{\pm 0.002}}}$ & $\mathbf{0.081_{{\pm 0.003}}}$ & $\mathbf{0.155_{{\pm 0.002}}}$ \\
 & $\text{Platt}$ & $0.314_{{\pm 0.001}}$ & $0.318_{{\pm 0.001}}$ & $0.307_{{\pm 0.001}}$ & $0.231_{{\pm 0.002}}$ & $0.357_{{\pm 0.000}}$ & $0.232_{{\pm 0.002}}$ & $0.289_{{\pm 0.001}}$ & $0.313_{{\pm 0.001}}$ & $0.318_{{\pm 0.001}}$ & $0.306_{{\pm 0.001}}$ & $0.230_{{\pm 0.002}}$ & $0.356_{{\pm 0.000}}$ & $0.232_{{\pm 0.002}}$ & $0.288_{{\pm 0.001}}$ \\
 & $\text{ATS}$ & $0.315_{{\pm 0.004}}$ & $0.318_{{\pm 0.001}}$ & $0.280_{{\pm 0.002}}$ & $0.236_{{\pm 0.004}}$ & $0.353_{{\pm 0.002}}$ & $0.237_{{\pm 0.003}}$ & $0.295_{{\pm 0.004}}$ & $0.315_{{\pm 0.004}}$ & $0.317_{{\pm 0.001}}$ & $0.280_{{\pm 0.002}}$ & $0.235_{{\pm 0.004}}$ & $0.352_{{\pm 0.002}}$ & $0.236_{{\pm 0.003}}$ & $0.295_{{\pm 0.004}}$ \\
\midrule
\multirow{5}{*}{\rotatebox{90}{\textbf{SQuAD}}} & $\text{Base}$ & $0.051_{{\pm 0.001}}$ & $0.047_{{\pm 0.000}}$ & $0.051_{{\pm 0.001}}$ & $0.048_{{\pm 0.000}}$ & $0.048_{{\pm 0.000}}$ & $0.048_{{\pm 0.000}}$ & $0.051_{{\pm 0.001}}$ & $\mathbf{0.053_{{\pm 0.001}}}$ & $0.051_{{\pm 0.001}}$ & $0.053_{{\pm 0.000}}$ & $0.054_{{\pm 0.000}}$ & $0.056_{{\pm 0.000}}$ & $0.054_{{\pm 0.000}}$ & $0.053_{{\pm 0.001}}$ \\
 & $\text{SE}$ & $\mathbf{0.050_{{\pm 0.001}}}$ & $0.048_{{\pm 0.001}}$ & $0.044_{{\pm 0.001}}$ & $0.046_{{\pm 0.000}}$ & $0.048_{{\pm 0.002}}$ & $\mathbf{0.046_{{\pm 0.000}}}$ & $0.049_{{\pm 0.001}}$ & $0.053_{{\pm 0.001}}$ & $0.051_{{\pm 0.001}}$ & $\mathbf{0.050_{{\pm 0.002}}}$ & $\mathbf{0.049_{{\pm 0.001}}}$ & $0.050_{{\pm 0.001}}$ & $\mathbf{0.049_{{\pm 0.001}}}$ & $0.053_{{\pm 0.001}}$ \\
 & $\text{TS}$ & $0.058_{{\pm 0.001}}$ & $0.060_{{\pm 0.001}}$ & $0.053_{{\pm 0.002}}$ & $0.051_{{\pm 0.001}}$ & $0.049_{{\pm 0.003}}$ & $0.052_{{\pm 0.002}}$ & $0.055_{{\pm 0.001}}$ & $0.064_{{\pm 0.001}}$ & $0.060_{{\pm 0.001}}$ & $0.056_{{\pm 0.002}}$ & $0.054_{{\pm 0.000}}$ & $\mathbf{0.049_{{\pm 0.002}}}$ & $0.053_{{\pm 0.002}}$ & $0.062_{{\pm 0.001}}$ \\
 & $\text{Platt}$ & $0.050_{{\pm 0.001}}$ & $\mathbf{0.047_{{\pm 0.000}}}$ & $\mathbf{0.044_{{\pm 0.001}}}$ & $\mathbf{0.046_{{\pm 0.000}}}$ & $\mathbf{0.047_{{\pm 0.002}}}$ & $0.046_{{\pm 0.000}}$ & $\mathbf{0.048_{{\pm 0.001}}}$ & $0.053_{{\pm 0.001}}$ & $\mathbf{0.049_{{\pm 0.001}}}$ & $0.050_{{\pm 0.002}}$ & $0.049_{{\pm 0.001}}$ & $0.050_{{\pm 0.002}}$ & $0.050_{{\pm 0.002}}$ & $\mathbf{0.052_{{\pm 0.001}}}$ \\
 & $\text{ATS}$ & $0.056_{{\pm 0.001}}$ & $\mathbf{0.047_{{\pm 0.000}}}$ & $0.055_{{\pm 0.001}}$ & $0.053_{{\pm 0.001}}$ & $0.053_{{\pm 0.001}}$ & $0.053_{{\pm 0.001}}$ & $0.056_{{\pm 0.001}}$ & $0.057_{{\pm 0.001}}$ & $\mathbf{0.049_{{\pm 0.001}}}$ & $0.057_{{\pm 0.001}}$ & $0.058_{{\pm 0.001}}$ & $0.057_{{\pm 0.001}}$ & $0.057_{{\pm 0.001}}$ & $0.056_{{\pm 0.001}}$ \\
\bottomrule
\end{tabular}}%
\label{tab:best_calibration_trivia_qa}
\end{table}

\begin{table}[h!]
\centering
\caption{\textbf{Ranking Volatility of Calibration Metrics Across Methods for Qwen.} Mean absolute rank change (volatility) of methods on TriviaQA, NQ, and SQuAD when switching from ACE (bin-based) to MCB-CORP (bin-free). Higher values indicate greater sensitivity of method rankings to the choice of calibration metric. \textbf{Bold} entries denote the lowest mean volatility per dataset, corresponding to the most stable ranking of methods.}
\begin{tabular}{cccc}
\toprule
 & TriviaQA & NQ & SQuAD  \\
\midrule
Base & 0.00 & 0.00 & 1.43 \\
SE & 0.29 & 0.14 & 1.29  \\
TS & 0.00 & 0.00 & 0.71 \\
Platt & 0.29 & 0.00 & 1.29  \\
ATS & 0.00 & 0.14 & 2.14  \\
\midrule
Mean & \textbf{0.11} & \textbf{0.06} & \textbf{1.37} \\
\bottomrule
\end{tabular}

\label{tab:combined_volatility}
\end{table}

\section{{Extended Analysis: Inductive Bias and Overfitting in Semantic Calibration}} 
\label{sec:inductive_bias_discussion}

Our results demonstrate that scalar TS consistently outperforms more expressive methods, such as ATS and Platt Scaling, in the context of semantic UQ. In this section, we expand on the theoretical and practical reasons for this finding, focusing on two key mechanisms: the preservation of semantic hierarchies and the prevention of overfitting to semantically irrelevant tokens.

\paragraph{Rank Preservation (TS vs Platt).}
A critical advantage of scalar TS is its strict preservation of the base model's likelihood-based token rankings at any given token position. TS applies a strictly monotonic transformation, guaranteeing that if the model assigns a higher logit to one token over another, this preference is preserved after temperature scaling. In contrast, more expressive methods like Platt Scaling introduce shift parameters or token-specific scaling. While theoretically more flexible, these affine transformations can distort the relative ranking of tokens. In the semantic setting, this is particularly dangerous: an affine transformation might inadvertently suppress a \textit{semantically crucial} token while upweighting a generic, high-frequency token. By design, TS is constrained to respect the model's original rank-ordering, making it inherently robust to such semantic distortions.

\paragraph{Overparameterisation and Filler Tokens (TS vs. ATS).}
The comparison with ATS highlights a critical failure mode driven by model capacity. ATS introduces transformer heads with millions of parameters, creating a high susceptibility to overparameterisation given the relatively small size of standard QA calibration sets. We argue that ATS exploits this capacity to overfit to \textit{semantically irrelevant} features. Since generic filler tokens such as stop words and punctuation are far more abundant than semantically loaded terms, ATS can drastically reduce the token-level NLL by aggressively optimising temperatures for these frequent but semantically irrelevant tokens. Empirically, we find that ATS frequently achieves near-zero training loss despite its poor generalisation to semantic UQ, indicating that this overfitting mechanism is indeed likely at play. Indeed, this is further supported from our observations that TS achieves a worse loss during training, yet achieves much better downstream semantic UQ performance. 

In contrast, we argue that Scalar TS provides a superior inductive bias for semantic calibration. By enforcing a single global constraint across the entire generation, TS acts as a powerful regulariser against the overfitting behaviour observed in ATS. Specifically, the method lacks the capacity to minimise loss by merely fitting frequent, non-semantic filler tokens. Instead, TS forces the calibration to reflect the uncertainty of the sequence as a holistic unit, thereby better capturing the overall meaning conveyed. In this context, TS represents a robust Occam-style model selection \citep{mackay2003information}, where the simplest sufficient constraint yields superior semantic generalisation.

\section{{Comparisons of Semantic Measures of Confidence}}

\begin{figure}[ht!]
    \centering
    \includegraphics[width=1.0\linewidth]{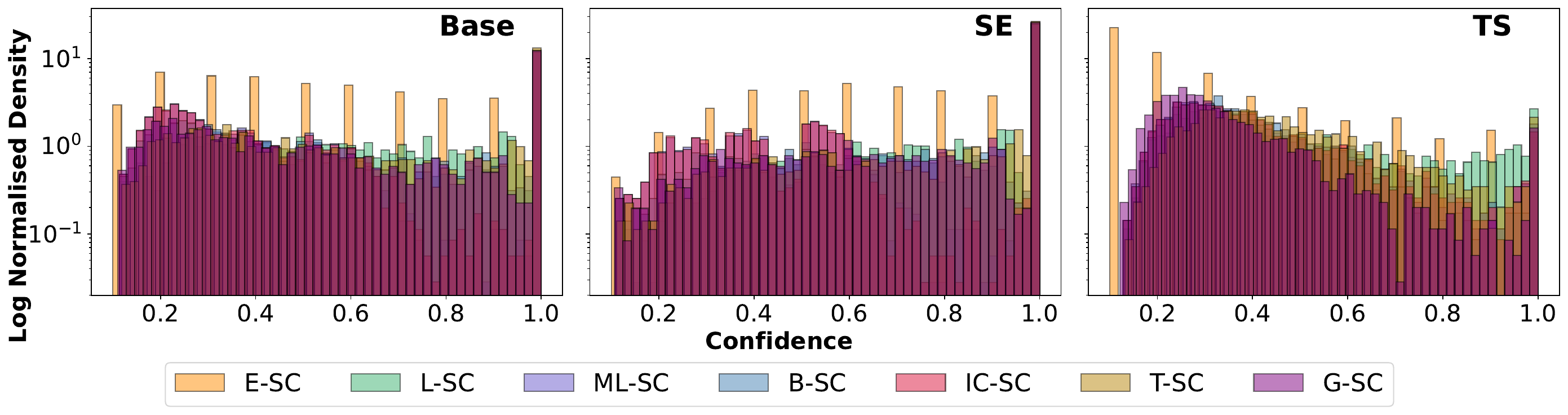}
    \caption{\textbf{SC Distribution Plots.} Distributions of semantic confidence measures for the Base, SE, and TS methods on the NQ dataset with the Llama model.}
    \label{fig:distribution-plots}
\end{figure}

\paragraph{Semantic Confidence Measure Distribution Plots.} To highlight differences between SC measures, and to illustrate the effect of the temperature parameter on their distributions, we plot the semantic measure distributions for the Base, SE, and TS methods of the Llama model on the NQ dataset in \Cref{fig:distribution-plots}.

\textbf{Correlation Between Semantic Measures Across Temperature Settings.} To complement the distribution plots in \Cref{fig:distribution-plots}, we report pairwise Pearson correlations between the semantic confidence (SC) measures under the Base, SE, and TS settings.

\textbf{Correlation Between Semantic Measures Across Temperature Settings.} To complement the distribution plots show in \Cref{fig:distribution-plots}, we report pairwise Pearson correlations between the semantic confidence (SC) measures under the Base, SE, and TS settings.

\clearpage

\begin{figure}[t!]
    \centering
    \includegraphics[width=1.0\linewidth]{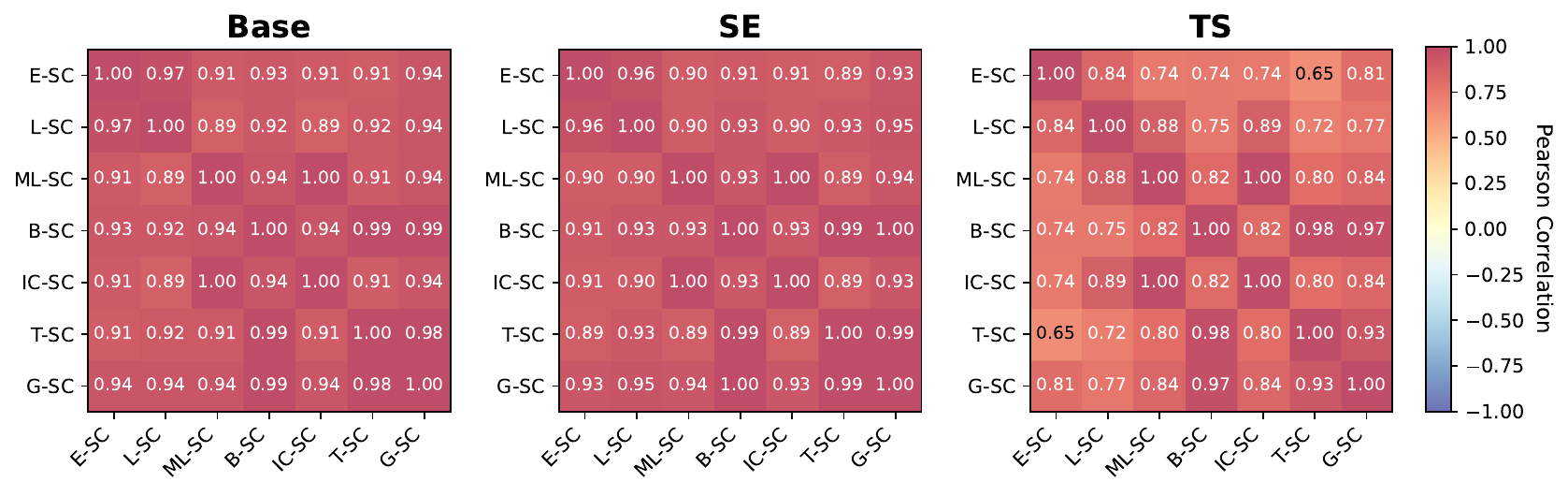}
    \caption{\textbf{Correlation matrices of semantic confidence measures.} Pairwise Pearson correlations between SC measures for the Base, SE, and TS methods on the NQ dataset with the Llama model. While correlations are very high under Base and SE settings, indicating redundancy between measures, optimised Temperature Scaling (TS) substantially lowers correlations, revealing greater diversity in how measures capture semantic uncertainty.}
    \label{fig:sc-measure-correlation}
\end{figure}

The results in \Cref{fig:sc-measure-correlation} show that at Base and SE temperatures, semantic confidence measures are almost perfectly correlated (coefficients $>$ 0.9), behaving as near reparameterisations of one another. Under optimised Temperature Scaling (TS), correlations drop substantially (down to ~0.65), indicating that TS increases the diversity of the measures and highlights their distinct behaviour. Crucially, this reduced alignment is consistent with the differences observed in downstream calibration and discrimination metrics, confirming that the measures are not redundant but capture complementary aspects of semantic uncertainty.

\end{appendices}

%% file: refs.bib
@inproceedings{naeini2015obtaining,
  title={Obtaining well calibrated probabilities using bayesian binning},
  author={Naeini, Mahdi Pakdaman and Cooper, Gregory and Hauskrecht, Milos},
  booktitle={Proceedings of the AAAI conference on artificial intelligence},
  volume={29 (1)},
  year={2015}
}

@article{xie2024calibrating,
  title={Calibrating language models with adaptive temperature scaling},
  author={Xie, Johnathan and Chen, Annie S and Lee, Yoonho and Mitchell, Eric and Finn, Chelsea},
  journal={arXiv preprint arXiv:2409.19817},
  year={2024}
}

@inproceedings{williams-etal-2018-broad,
    title = "A Broad-Coverage Challenge Corpus for Sentence Understanding through Inference",
    author = "Williams, Adina  and
      Nangia, Nikita  and
      Bowman, Samuel",
    editor = "Walker, Marilyn  and
      Ji, Heng  and
      Stent, Amanda",
    booktitle = "Proceedings of the 2018 Conference of the North {A}merican Chapter of the Association for Computational Linguistics: Human Language Technologies, Volume 1 (Long Papers)",
    month = jun,
    year = "2018",
    address = "New Orleans, Louisiana",
    publisher = "Association for Computational Linguistics",
    url = "https://aclanthology.org/N18-1101/",
    doi = "10.18653/v1/N18-1101",
    pages = "1112--1122"
}

@article{lin2023generating,
  title={Generating with confidence: Uncertainty quantification for black-box large language models},
  author={Lin, Zhen and Trivedi, Shubhendu and Sun, Jimeng},
  journal={arXiv preprint arXiv:2305.19187},
  year={2023}
}

@article{yang2023bayesian,
  title={Bayesian low-rank adaptation for large language models},
  author={Yang, Adam X and Robeyns, Maxime and Wang, Xi and Aitchison, Laurence},
  journal={arXiv preprint arXiv:2308.13111},
  year={2023}
}

@inproceedings{blundell2015weight,
  title={Weight uncertainty in neural network},
  author={Blundell, Charles and Cornebise, Julien and Kavukcuoglu, Koray and Wierstra, Daan},
  booktitle={International conference on machine learning},
  pages={1613--1622},
  year={2015},
  organization={PMLR}
}

@article{lakshminarayanan2017simple,
  title={Simple and scalable predictive uncertainty estimation using deep ensembles},
  author={Lakshminarayanan, Balaji and Pritzel, Alexander and Blundell, Charles},
  journal={Advances in neural information processing systems},
  volume={30},
  year={2017}
}

@article{liu2020simple,
  title={Simple and principled uncertainty estimation with deterministic deep learning via distance awareness},
  author={Liu, Jeremiah and Lin, Zi and Padhy, Shreyas and Tran, Dustin and Bedrax Weiss, Tania and Lakshminarayanan, Balaji},
  journal={Advances in neural information processing systems},
  volume={33},
  pages={7498--7512},
  year={2020}
}

@inproceedings{mukhoti2023deep,
  title={Deep deterministic uncertainty: A new simple baseline},
  author={Mukhoti, Jishnu and Kirsch, Andreas and van Amersfoort, Joost and Torr, Philip HS and Gal, Yarin},
  booktitle={Proceedings of the IEEE/CVF Conference on Computer Vision and Pattern Recognition},
  pages={24384--24394},
  year={2023}
}

@article{dimitriadis2021stable,
  title={Stable reliability diagrams for probabilistic classifiers},
  author={Dimitriadis, Timo and Gneiting, Tilmann and Jordan, Alexander I},
  journal={Proceedings of the National Academy of Sciences},
  volume={118},
  number={8},
  pages={e2016191118},
  year={2021},
  publisher={National Academy of Sciences}
}

@article{de2010isotone,
  title={Isotone optimization in R: pool-adjacent-violators algorithm (PAVA) and active set methods},
  author={De Leeuw, Jan and Hornik, Kurt and Mair, Patrick},
  journal={Journal of statistical software},
  volume={32},
  pages={1--24},
  year={2010}
}

@article{nikitin2024kernel,
  title={Kernel Language Entropy: Fine-grained Uncertainty Quantification for LLMs from Semantic Similarities},
  author={Nikitin, Alexander and Kossen, Jannik and Gal, Yarin and Marttinen, Pekka},
  journal={arXiv preprint arXiv:2405.20003},
  year={2024}
}

@book{cantoni2004statistical,
  title={Statistical Learning Theory and Stochastic Optimization: Ecole D'Ete de Probabilites de Saint-Flour XXXI-2001},
  author={Cantoni, Olivier and Picard, Jean},
  year={2004},
  publisher={Springer}
}

@inproceedings{malininuncertainty,
  title={Uncertainty Estimation in Autoregressive Structured Prediction},
  author={Malinin, Andrey and Gales, Mark},
  booktitle={International Conference on Learning Representations},
    year = {2021}
}

@inproceedings{aichberger2025improving,
  title={Improving uncertainty estimation through semantically diverse language generation},
  author={Aichberger, Lukas and Schweighofer, Kajetan and Ielanskyi, Mykyta and Hochreiter, Sepp},
  booktitle={The Thirteenth International Conference on Learning Representations},
  year={2025}
}

@inproceedings{niculescu2005predicting,
  title={Predicting good probabilities with supervised learning},
  author={Niculescu-Mizil, Alexandru and Caruana, Rich},
  booktitle={Proceedings of the 22nd international conference on Machine learning},
  pages={625--632},
  year={2005}
}

@inproceedings{wenzel2020good,
  title={How Good is the Bayes Posterior in Deep Neural Networks Really?},
  author={Wenzel, Florian and Roth, Kevin and Veeling, Bastiaan and Swiatkowski, Jakub and Tran, Linh and Mandt, Stephan and Snoek, Jasper and Salimans, Tim and Jenatton, Rodolphe and Nowozin, Sebastian},
  booktitle={International Conference on Machine Learning},
  pages={10248--10259},
  year={2020},
  organization={PMLR}
}

@article{wei2024long,
  title={Long-form factuality in large language models},
  author={Wei, Jerry and Yang, Chengrun and Song, Xinying and Lu, Yifeng and Hu, Nathan and Huang, Jie and Tran, Dustin and Peng, Daiyi and Liu, Ruibo and Huang, Da and others},
  journal={Advances in Neural Information Processing Systems},
  volume={37},
  pages={80756--80827},
  year={2024}
}

@article{platt1999probabilistic,
  title={Probabilistic outputs for support vector machines and comparisons to regularized likelihood methods},
  author={Platt, John and others},
  journal={Advances in large margin classifiers},
  volume={10},
  number={3},
  pages={61--74},
  year={1999},
  publisher={Cambridge, MA}
}

@inproceedings{nixon2019measuring,
  title={Measuring Calibration in Deep Learning.},
  author={Nixon, Jeremy and Dusenberry, Michael W and Zhang, Linchuan and Jerfel, Ghassen and Tran, Dustin},
  booktitle={CVPR workshops},
  volume={2},
  year={2019}
}

@techreport{radford2018improving,
  title={Improving Language Understanding by Generative Pre-Training},
  author={Radford, Alec and Narasimhan, Karthik and Salimans, Tim and Sutskever, Ilya},
  year={2018},
  institution={OpenAI},
  url={https://openai.com/research/language-unsupervised}
}

@misc{bachmann2021rapidfuzz,
  author       = {Maximilian Bachmann},
  title        = {RapidFuzz: A fast fuzzy string matching library in C++ and Python},
  year         = {2021},
  publisher    = {Zenodo},
  doi          = {10.5281/zenodo.5141757},
  url          = {https://github.com/maxbachmann/RapidFuzz}
}

@incollection{flach2016classifier,
  title={Classifier calibration},
  author={Flach, Peter A},
  booktitle={Encyclopedia of machine learning and data mining},
  year={2016},
  publisher={Springer US}
}

@inproceedings{murray2018correcting,
  title={Correcting Length Bias in Neural Machine Translation},
  author={Murray, Kenton and Chiang, David},
  booktitle={Proceedings of the Third Conference on Machine Translation: Research Papers},
  pages={212--223},
  year={2018}
}

@misc{mistral2024ministraux,
  author       = {MistralAI},
  title        = {Introducing Ministrel: Our New Lightweight Model},
  year         = {2024},
  month        = {October},
  url          = {https://mistral.ai/news/ministraux/},
  note         = {Accessed: 2025-01-17}
}

@misc{qwen2024qwen,
  author       = {Qwen Team},
  title        = {Qwen2.5: Advancing Open-Source Language Models},
  year         = {2024},
  month        = {September},
  url          = {https://qwenlm.github.io/blog/qwen2.5/},
  note         = {Accessed: 2025-01-17}
}

@article{dubey2024llama,
  title={The llama 3 herd of models},
  author={Dubey, Abhimanyu and Jauhri, Abhinav and Pandey, Abhinav and Kadian, Abhishek and Al-Dahle, Ahmad and Letman, Aiesha and Mathur, Akhil and Schelten, Alan and Yang, Amy and Fan, Angela and others},
  journal={arXiv preprint arXiv:2407.21783},
  year={2024}
}

@article{gneiting2007strictly,
  title={Strictly proper scoring rules, prediction, and estimation},
  author={Gneiting, Tilmann and Raftery, Adrian E},
  journal={Journal of the American statistical Association},
  volume={102},
  number={477},
  pages={359--378},
  year={2007},
  publisher={Taylor \& Francis}
}

@article{vaswani2017attention,
  title={Attention is all you need},
  author={Vaswani, Ashish and Shazeer, Noam and Parmar, Niki and Uszkoreit, Jakob and Jones, Llion and Gomez, Aidan N and Kaiser, Lukasz and Polosukhin, Illia},
  journal={Advances in Neural Information Processing Systems},
  year={2017}
}

@article{farquhar2024detecting,
  title={Detecting hallucinations in large language models using semantic entropy},
  author={Farquhar, Sebastian and Kossen, Jannik and Kuhn, Lorenz and Gal, Yarin},
  journal={Nature},
  volume={630},
  number={8017},
  pages={625--630},
  year={2024},
  publisher={Nature Publishing Group UK London}
}

@inproceedings{guo2017calibration,
  title={On calibration of modern neural networks},
  author={Guo, Chuan and Pleiss, Geoff and Sun, Yu and Weinberger, Kilian Q},
  booktitle={International conference on machine learning},
  pages={1321--1330},
  year={2017},
  organization={PMLR}
}

@article{kadavath2022language,
  title={Language models (mostly) know what they know},
  author={Kadavath, Saurav and Conerly, Tom and Askell, Amanda and Henighan, Tom and Drain, Dawn and Perez, Ethan and Schiefer, Nicholas and Hatfield-Dodds, Zac and DasSarma, Nova and Tran-Johnson, Eli and others},
  journal={arXiv preprint arXiv:2207.05221},
  year={2022}
}

@article{santilli2025revisiting,
  title={Revisiting uncertainty quantification evaluation in language models: Spurious interactions with response length bias results},
  author={Santilli, Andrea and Golinski, Adam and Kirchhof, Michael and Danieli, Federico and Blaas, Arno and Xiong, Miao and Zappella, Luca and Williamson, Sinead},
  journal={arXiv preprint arXiv:2504.13677},
  year={2025}
}

@article{achiam2023gpt,
  title={Gpt-4 technical report},
  author={Achiam, Josh and Adler, Steven and Agarwal, Sandhini and Ahmad, Lama and Akkaya, Ilge and Aleman, Florencia Leoni and Almeida, Diogo and Altenschmidt, Janko and Altman, Sam and Anadkat, Shyamal and others},
  journal={arXiv preprint arXiv:2303.08774},
  year={2023}
}

@article{murphy1973new,
  title={A new vector partition of the probability score},
  author={Murphy, Allan H},
  journal={Journal of Applied Meteorology and Climatology},
  volume={12},
  number={4},
  pages={595--600},
  year={1973}
}

@article{bradley1997use,
  title={The use of the area under the ROC curve in the evaluation of machine learning algorithms},
  author={Bradley, Andrew P},
  journal={Pattern recognition},
  volume={30},
  number={7},
  pages={1145--1159},
  year={1997},
  publisher={Elsevier}
}

@article{kuhn2023semantic,
  title={Semantic uncertainty: Linguistic invariances for uncertainty estimation in natural language generation},
  author={Kuhn, Lorenz and Gal, Yarin and Farquhar, Sebastian},
  journal={International Conference on Learning Representations},
  year={2023}
}

@inproceedings{
he2021deberta,
title={DEBERTA: DECODING-ENHANCED BERT WITH DISENTANGLED ATTENTION},
author={Pengcheng He and Xiaodong Liu and Jianfeng Gao and Weizhu Chen},
booktitle={International Conference on Learning Representations},
year={2021},
url={https://openreview.net/forum?id=XPZIaotutsD}
}

@misc{anthropic_claude45,
  title        = {Claude 4.5 Model Family Announcement},
  author       = {{Anthropic}},
  howpublished = {\url{https://platform.claude.com/docs/en/about-claude/models/whats-new-claude-4-5}},
  note         = {Accessed: 2026-01},
  year         = {2025}
}

@book{mackay2003information,
  title={Information Theory, Inference, and Learning Algorithms},
  author={MacKay, David J. C.},
  year={2003},
  publisher={Cambridge University Press}
}

@article{huang2020tutorial,
  title={A tutorial on calibration measurements and calibration models for clinical prediction models},
  author={Huang, Yingxiang and Li, Wentao and Macheret, Fima and Gabriel, Rodney A and Ohno-Machado, Lucila},
  journal={Journal of the American Medical Informatics Association},
  volume={27},
  number={4},
  pages={621--633},
  year={2020},
  publisher={Oxford University Press}
}

@inproceedings{desai2020calibration,
  title={Calibration of Pre-trained Transformers},
  author={Desai, Shrey and Durrett, Greg},
  booktitle={Proceedings of the 2020 Conference on Empirical Methods in Natural Language Processing (EMNLP)},
  pages={295--302},
  year={2020}
}

@article{joshi2017triviaqa,
  title={Triviaqa: A large scale distantly supervised challenge dataset for reading comprehension},
  author={Joshi, Mandar and Choi, Eunsol and Weld, Daniel S and Zettlemoyer, Luke},
  journal={arXiv preprint arXiv:1705.03551},
  year={2017}
}

@article{kwiatkowski2019natural,
  title={Natural Questions: A Benchmark for Question Answering Research},
  author={Kwiatkowski, Tom and Palomaki, Jennimaria and Redfield, Olivia and Collins, Michael and Parikh, Ankur and Alberti, Chris and Epstein, Danielle and Polosukhin, Illia and Devlin, Jacob and Lee, Kenton and others},
  journal={Transactions of the Association for Computational Linguistics},
  volume={7},
  pages={453--466},
  year={2019},
  publisher={MIT Press}
}

@article{rajpurkar2016squad,
  title={Squad: 100,000+ questions for machine comprehension of text},
  author={Rajpurkar, P},
  journal={arXiv preprint arXiv:1606.05250},
  year={2016}
}

@article{hu2021lora,
  title={Lora: Low-rank adaptation of large language models},
  author={Hu, Edward J and Shen, Yelong and Wallis, Phillip and Allen-Zhu, Zeyuan and Li, Yuanzhi and Wang, Shean and Wang, Lu and Chen, Weizhu},
  journal={arXiv preprint arXiv:2106.09685},
  year={2021}
}

@article{loshchilov2017decoupled,
  title={Decoupled weight decay regularization},
  author={Loshchilov, Ilya and Hutter, Frank},
  journal={arXiv preprint arXiv:1711.05101},
  year={2017}
}
